\documentclass[10pt,twocolumn,letterpaper]{article}

\pdfoutput=1
\usepackage{cvpr}              
\usepackage[accsupp]{axessibility}
\usepackage{pifont}
\usepackage{times}
\usepackage{microtype}
\usepackage{epsfig}
\usepackage{caption}
\usepackage{float}
\usepackage{placeins}
\usepackage{color, colortbl}
\usepackage{stfloats}
\usepackage{enumitem}
\usepackage{xstring}
\usepackage{multirow}
\usepackage{xspace}
\usepackage{url}
\usepackage{subcaption}
\usepackage{xcolor}
\usepackage{algorithmic}
\usepackage[hang,flushmargin]{footmisc}
\usepackage{threeparttable}
\usepackage{amsmath}
\usepackage{amssymb}
\usepackage{xspace}

\usepackage{graphicx}
\usepackage{bm}
\usepackage{booktabs}
\usepackage{bigstrut}
\usepackage{bbding}
\usepackage[misc]{ifsym}

\newcommand{\xmark}{\ding{55}} 
\newcommand{\cmark}{\ding{51}}

\usepackage{xr-hyper}

\makeatletter
\newcommand*{\addFileDependency}[1]{
  \typeout{(#1)}
  \@addtofilelist{#1}
  \IfFileExists{#1}{}{\typeout{No file #1.}}
}

\makeatother

\usepackage{tabularx}

\makeatletter
\def\thanks#1{\protected@xdef\@thanks{\@thanks
        \protect\footnotetext{#1}}}

\usepackage{longtable} 
\usepackage[most]{tcolorbox}
\newtcbox{\mybox}[1][red]{on line, colback=red, colframe=#1, boxsep=0pt, boxrule=0pt, size=small, arc=0mm}
\definecolor{white}{RGB}{255, 255, 255}
\usepackage{tikz}
\newcommand*{\goldmedal}{\includegraphics[scale=0.6]{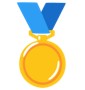}}%
\newcommand*{\silvermedal}{\includegraphics[scale=0.6]{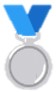}}%
\newcommand*{\redcheck}{\includegraphics[scale=0.6]{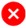}}%
\newcommand*{\greencheck}{\includegraphics[scale=0.6]{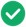}}%
\definecolor{cvprblue}{rgb}{0.21,0.49,0.74}
\definecolor{LightGray}{rgb}{0.957,0.957,0.996}
\usepackage[pagebackref,breaklinks,colorlinks,allcolors=cvprblue]{hyperref}
\usepackage[capitalize]{cleveref}
\crefname{section}{Sec.}{Secs.}
\crefname{table}{Table}{Tables}
\crefname{figure}{Fig.}{Figs.}

\def\paperID{3919} 
\def\confName{CVPR}
\def\confYear{2025}

\title{OpenING: A Comprehensive Benchmark for Judging \\Open-ended Interleaved Image-Text Generation}

 \author{
    Pengfei Zhou\textsuperscript{1*}, 
    Xiaopeng Peng\textsuperscript{2*}, 
    Jiajun Song\textsuperscript{3},
    Chuanhao Li\textsuperscript{1}, 
    Zhaopan Xu\textsuperscript{1},
    Yue Yang\textsuperscript{4,1}, \\
    Ziyao Guo\textsuperscript{1,5},
    Hao Zhang\textsuperscript{1}, 
    Yuqi Lin\textsuperscript{1},
    Yefei He\textsuperscript{1},
    Lirui Zhao\textsuperscript{1}, 
    Shuo Liu\textsuperscript{1}, 
    Tianhua Li\textsuperscript{1,4}, 
    Yuxuan Xie\textsuperscript{1,4}, \\
    Xiaojun Chang\textsuperscript{6,7},
    Yu Qiao\textsuperscript{1}, 
    Wenqi Shao\textsuperscript{1}, 
    Kaipeng Zhang\textsuperscript{1,8$\dagger$}\\
 \hspace*{-1.3em} \textsuperscript{1}Shanghai AI Laboratory, 
  \textsuperscript{2}RIT,
  \textsuperscript{3}RUC, 
  \textsuperscript{4}SJTU,
  \textsuperscript{5}NUS,
  \textsuperscript{6}USTC,
  \textsuperscript{7}MBZUAI,
  \textsuperscript{8}Shanghai Innovation Institute\\
  \url{https://opening-benchmark.github.io}
    \thanks{\hspace*{-1.8em}$^*$Equal contribution $^\dagger$Corresponding author (zhangkaipeng@pjlab.org.cn)}
}

\newif\ifreview 
\newif\ifarxiv 
\newif\ifcamera \newcommand{\cameraready}{\cameratrue}
\newif\ifrebuttal 

\cameraready

\begin{document}
\maketitle
\begin{abstract}
    Multimodal Large Language Models (MLLMs) have made significant strides in visual understanding and generation tasks. However, generating interleaved image-text content remains a challenge, which requires integrated multimodal understanding and generation abilities. While the progress in unified models offers new solutions, existing benchmarks are insufficient for evaluating these methods due to data size and diversity limitations. To bridge this gap, we introduce OpenING, a comprehensive benchmark comprising 5,400 high-quality human-annotated instances across 56 real-world tasks. OpenING covers diverse daily scenarios such as travel guide, design, and brainstorming, offering a robust platform for challenging interleaved generation methods. In addition, we present IntJudge, a judge model for evaluating open-ended multimodal generation methods. Trained with a novel data pipeline, our IntJudge achieves an agreement rate of 82.42\% with human judgments, outperforming GPT-based evaluators by 11.34\%. Extensive experiments on OpenING reveal that current interleaved generation methods still have substantial room for improvement. Key findings on interleaved image-text generation are further presented to guide the development of next-generation models. 
    



\end{abstract}
\vspace{-0.2cm}

\section{Introduction}
\label{sec:intro}

\begin{figure}[t]
	\centering
  \includegraphics[trim=0.6cm 0.5cm 0 0, width=0.485\textwidth]{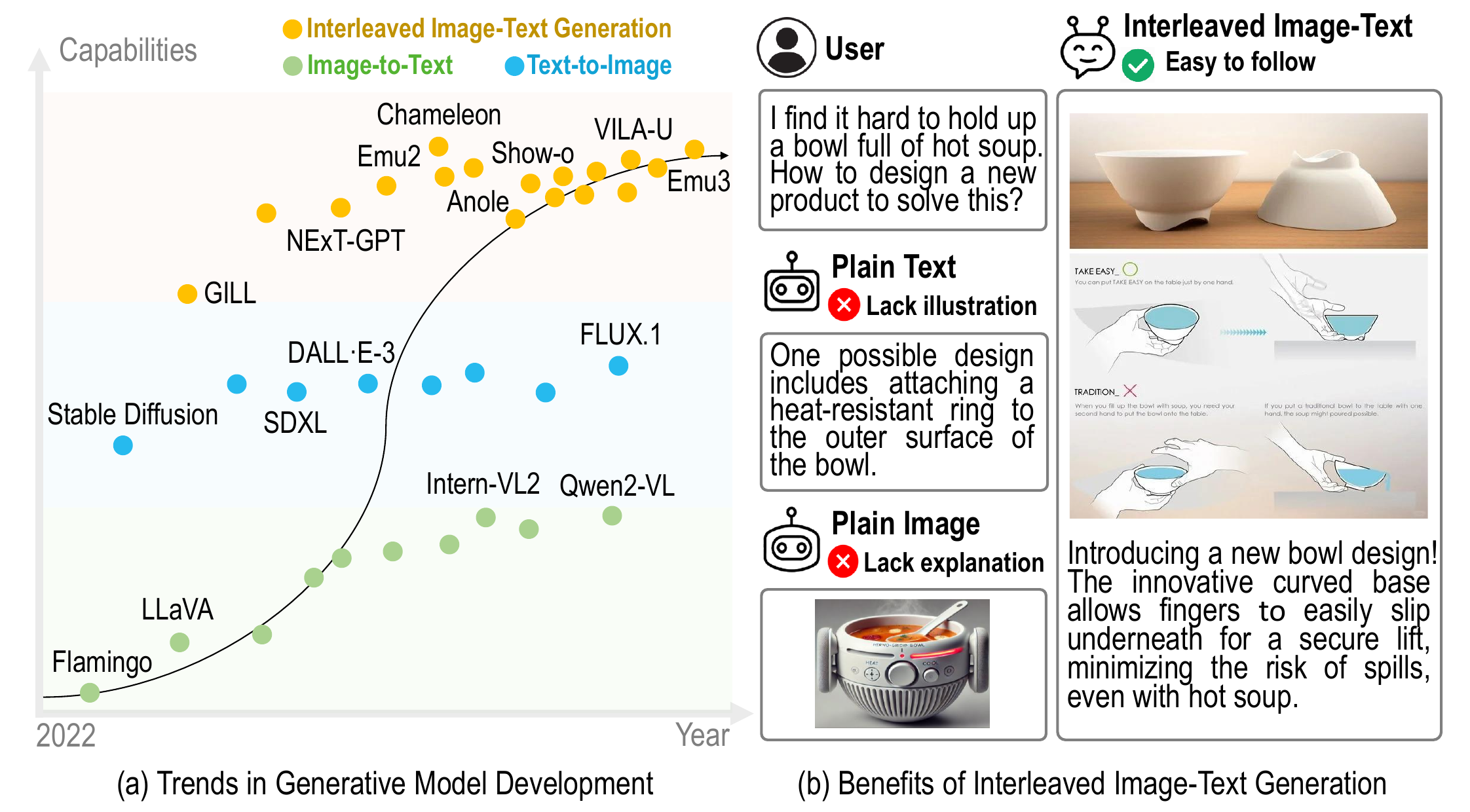}
  \caption{Motivation: (a) Rapid progress of interleaved image-text generation. (b) Interleaved content is essential to provide key information for complex real-world tasks (e.g., product design).}
  \vspace{-0.2cm}
  \label{fig:intro} 
\end{figure}

Building upon the remarkable understanding and generation capabilities of Large Language Models (LLMs)~\cite{openai2023gpt,team2023gemini,touvron2023llama,team2023internlm}, Multimodal LLMs (MLLMs) are making progress in various tasks~\cite{zhao2024stitch,yang2023dawn,liu2024visual,zhang2023internlm,bai2023qwen}. However, generating interleaved image-text content remains challenging~\cite{team2024chameleon,wang2024emu3,laurenccon2024obelics}, despite its important role in both research and applications (e.g., multimodal reasoning~\cite{byun2024ares,mu2024embodiedgpt}, education~\cite{latif2023artificial,claman2024artificial} and design~\cite{stella2023can,ko2023large}). Since human brains can naturally combine visual and textual signals for more efficient information exchange~\cite{glasser2016multi,holler2019multimodal}, achieving such integrated ability is crucial for advancing towards Artificial General Intelligence (AGI).

 \begin{figure*}[t]
	\centering
	\includegraphics[trim=0cm 0.2cm 0 0, width=0.98\textwidth]{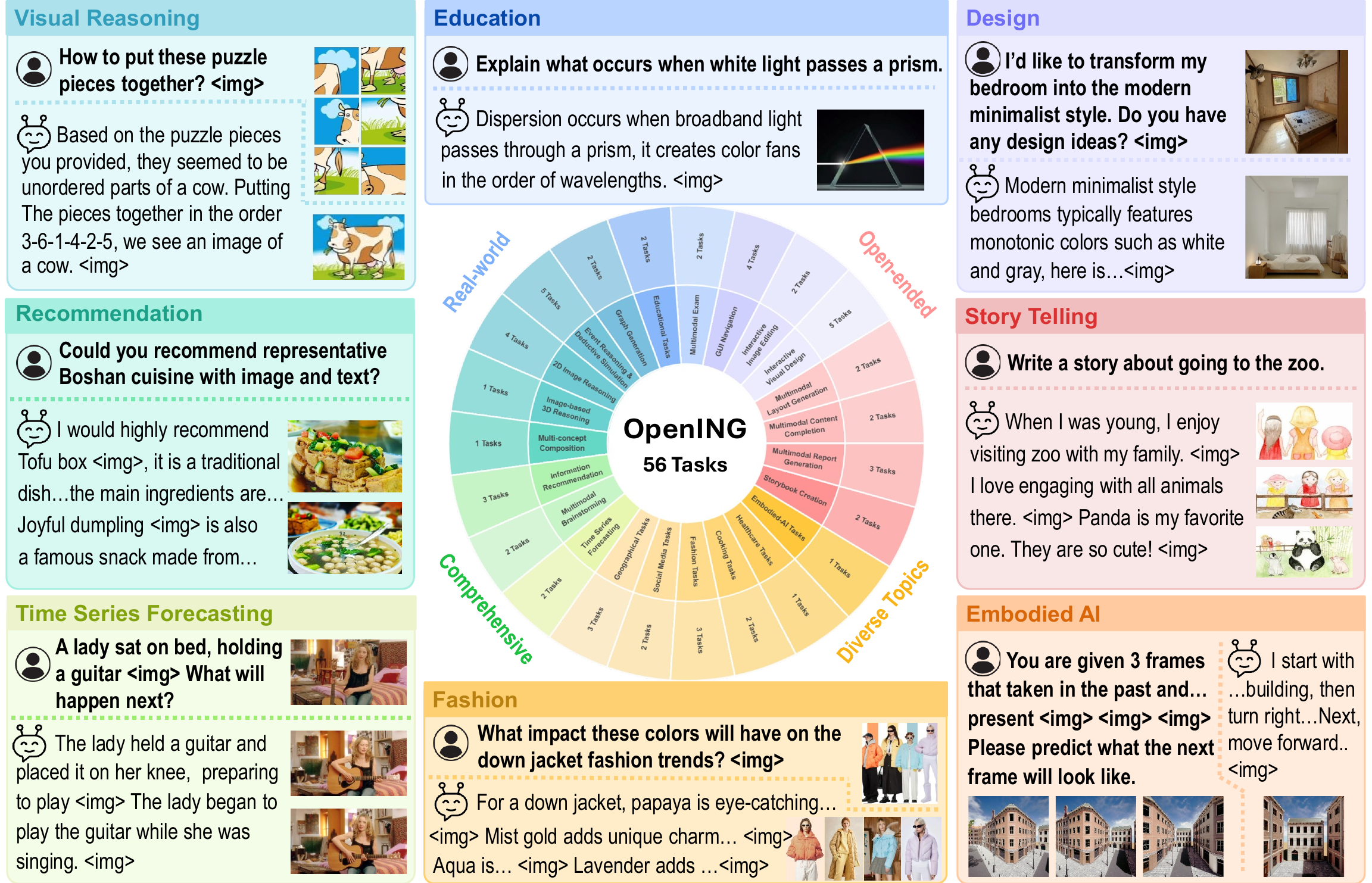}
 \caption{OpenING benchmark consists of 23 meta-topics (inner ring) which are further categorized into 56 specific tasks (see the number of tasks on the outer ring and details in Supplementary Materials). Examples showcase interleaved generation in eight representative domains.}
	\label{fig:data}
\end{figure*}

As shown in Fig.~\ref{fig:intro}, the emergence of unified models that combine understanding and generation abilities opens up new possibilities for interleaved image-text generation~\cite{xie2024show,zhou2024transfusion}. However, the lack of reliable benchmarks to evaluate interleaved generation remains an obstacle~\cite{wang2024emu3,tang2024codi}. Most existing benchmarks evaluate text or image output separately, failing to capture the intricacies of simultaneously generating both ~\cite{liu2023mmbench, yue2024mmmu, ying2024mmt, tan2024evalalign}. Interleaved benchmarks like OpenLEAF~\cite{an2023openleaf} and InterleavedBench~\cite{liu2024holistic} are limited in size, scope, and query diversity.  For example, InterleavedBench includes only 815 instances across 10 tasks sourced from public datasets such as VIST~\cite{huang2016visual} and WikiHow~\cite{yang2021visual}. These benchmarks do not adequately reflect real-world needs and are vulnerable to data contamination~\cite{xia2024mmie}.

To fill the gap, we introduce OpenING, a comprehensive benchmark for evaluating open-ended interleaved generation. Unlike previous benchmarks, OpenING offers a broader set of real-world data and tasks (e.g., brainstorming, recommendations, and content creation) derived from daily scenarios like fashion, cooking, and travel. As shown in Fig.~\ref{fig:data} and Table~\ref{tab:dataset_comparison}, the curated OpenING includes 5,400 instances of multi-step interleaved image-text content across 23 meta-topics and 56 tasks, with diverse, carefully designed queries for various topics. 
To address the challenges of collecting and standardizing data from disparate domains, we develop an efficient annotation pipeline and produce high-quality human-annotated data, reducing data contamination risks.

In addition, many previous benchmarks rely on GPT-based scoring metrics~\cite{an2023openleaf,liu2024holistic}, which are prone to be affected by inherent biases in GPT models and potential data leakage in API usage~\cite{wang2023pandalm}. To overcome the challenges of assessing open-ended multimodal generation, we introduce IntJudge, a robust judge model. We also propose an Interleaved Arena to facilitate annotation of training data and a Reference-Augmented Generation (RAG) approach to scale up the data size. Trained with this enhanced data pipeline, IntJudge achieves 82.42\% average agreement with human judgments, an 11.34\% improvement over GPT-4o as a judge.

We evaluate representative interleaved generation methods using our OpenING. Key findings from our experiments include: 1) Generating coherent and high-quality interleaved content remains challenging for all models, whereas human-annotated content consistently receives the highest ratings over generated content; 2) Integrated pipelines (e.g., Gemini + Flux) outperform end-to-end models (e.g., Anole) in terms of image-text coherence and visual quality, possibly due to the more developed foundation models. End-to-end and two-stage generators (e.g., SEED-X) with unified architectures show great potential, especially as they continue to advance and potentially take advantage of developed methods; and 3) While text answers generated by GPT can be more informative than human-annotated answers, annotated natural images remain preferable to generated images, highlighting the challenges in high-quality image generation. The major contributions of this paper are summarized as follows:

\begin{itemize}
    \item \textbf{A High-quality Benchmark}. We present OpenING, a comprehensive benchmark for evaluating open-ended interleaved image-text generation. OpenING includes 5,400 human-annotated instances across 56 real-world tasks, aiming to challenge and improve interleaved generation methods and also support the development of judge models for assessing open-ended multimodal generation. 

    \item  \textbf{A Robust Judge}. We introduce IntJudge, a judge model for rating interleaved generation methods. We train IntJudge with an enhanced data pipeline, achieving an 82.42\% agreement rate with human judgments and significantly outperforming GPT-based judge. Moreover, IntJudge has proven to be effective in assessing new unseen models.

    \item  \textbf{A Comprehensive Leaderboard}. We provide detailed rankings and analysis of interleaved generation methods and compare IntJudge and GPT-4o evaluations with human judgments. Our findings indicate that while current open-source end-to-end models lag behind integrated pipelines, end-to-end and two-stage generators with unified architectures present great potential and warrant further explorations to advance interleaved image-text generation.

\end{itemize}

\begin{table*}[!htbp]
  \centering
   \setlength{\tabcolsep}{2.35mm}
   \begin{tabular}{@{}>{\hspace{2.5pt}}l*{9}{c}@{}}
    \toprule
    \multicolumn{1}{c}{\multirow{2}[4]{*}{Benchmark}}  & \multicolumn{6}{c}{Data Coverage}  & \multirow{2}[4]{*}{Open-source} & \multirow{2}[4]{*}{Offline Judge} \\
\cmidrule{2-7} & Meta-Topics & Tasks  & Instances   & Images & Steps & SpI  \\
    \midrule
    OpenLEAF \cite{an2023openleaf}   & 2 & 10 & 660 & -  & - & - & \xmark & \xmark \\
    InterleavedBench \cite{liu2024holistic} & 4 & 10 & 815 & 1,513 & 1,601 & 1.96 & \cmark & \xmark \\
    OpenING (ours) & \textbf{23} & \textbf{56} & \textbf{5,400} & \textbf{17,603} & \textbf{20,094} & \textbf{3.72} & \cmark & \cmark \\
    \bottomrule
    \end{tabular}
    \vspace{-0.2cm}
    \caption{Comparison between OpenING and existing benchmarks. OpenING includes more comprehensive data and task coverage with an openly available judge model. Steps: a step is indicated by an input instruction or an output image-text pair; SpI: Steps per Instance.}
  \label{tab:dataset_comparison}%
\end{table*}%

\section{Related Work}
\label{sec:related}

\noindent\textbf{Interleaved Image-Text Generation.} The development of MLLMs has greatly advanced interleaved image-text generation~\cite{koh2024generating}. Early models like Stable Diffusion \cite{rombach2022high, esser2024scaling}, DALL-E \cite{ramesh2022hierarchical}, and autoregressive (AR) methods (e.g., VAR \cite{tian2024visual} and Lumina-mGPT \cite{liu2024lumina}) focused on unidirectional tasks, such as image understanding and text-to-image generation. Flamingo \cite{alayrac2022flamingo} was the first MLLM to process interleaved image-text content. More recent models, such as MiniGPT-5 \cite{zheng2023minigpt} and SEED series \cite{gemaking, ge2024seed, yang2024seed}, achieve interleaved generation by combining AR-based text generation and diffusion-based visual generation. Native AR models such as Emu-3~\cite{wang2024emu3} and Chameleon \cite{team2024chameleon} offer a unified framework to generate and reason over mixed-modal documents. Anole~\cite{chern2024anole} reproduces the image generation capability of Chameleon through efficient fine-tuning on an interleaved image-text data.
However, benchmarks for evaluating interleaved image-text generation remain in early stages. Previous works, such as OpenLEAF~\cite{an2023openleaf} and InterleavedBench~\cite{liu2024holistic} focused on a small set of topics and lack the depth and breadth for real-world applications. To achieve a more reliable and holistic evaluation of interleaved generation, we propose OpenING based on comprehensive real-world scenarios.

\noindent\textbf{Evaluation of Open-ended Multimodal Generation.}
Evaluating open-ended multimodal generation is inherently challenging due to the need to assess both visual and textual quality in open domains~\cite{an2023openleaf,wu2024towards,shao2023building}. Existing text generation metrics, such as BLEU~\cite{papineni2002bleu} and ROUGE~\cite{lin2004rouge}, fall short in measuring visual quality and text-image coherence. Conversely, visual quality metrics like FID~\cite{heusel2017gans} and IS~\cite{salimans2016improved} lack consideration of textual elements. Contrastive metrics, such as CLIPScore~\cite{hessel2021clipscore} can measure text-image alignment but fail to fully evaluate the quality of open-ended interleaved contents, where multiple correct answers may exist. GPT-based scoring~\cite{zhang2023gpt, liu2024holistic} provides improved measurements to assess the diversity and coherence of the interleaved outputs. However, GPT tends to be biased and favors the contents generated by itself~\cite{wang2023pandalm,bai2024benchmarking}. Human evaluation, although reliable, is not scalable due to its laborious nature. To close this gap, we introduce IntJudge, a judge model that is highly aligned with human judgments in evaluating the open-ended multimodal generation. To mitigate the instability of subjective scores~\cite{zheng2023judging,chenmllm}, our IntJudge evaluates models through pairwise comparisons in an arena-style framework~\cite{li2024k}.

\begin{figure*}[t]
	\centering
	\includegraphics[trim=0 0.3cm 0 0, width=0.98\textwidth]{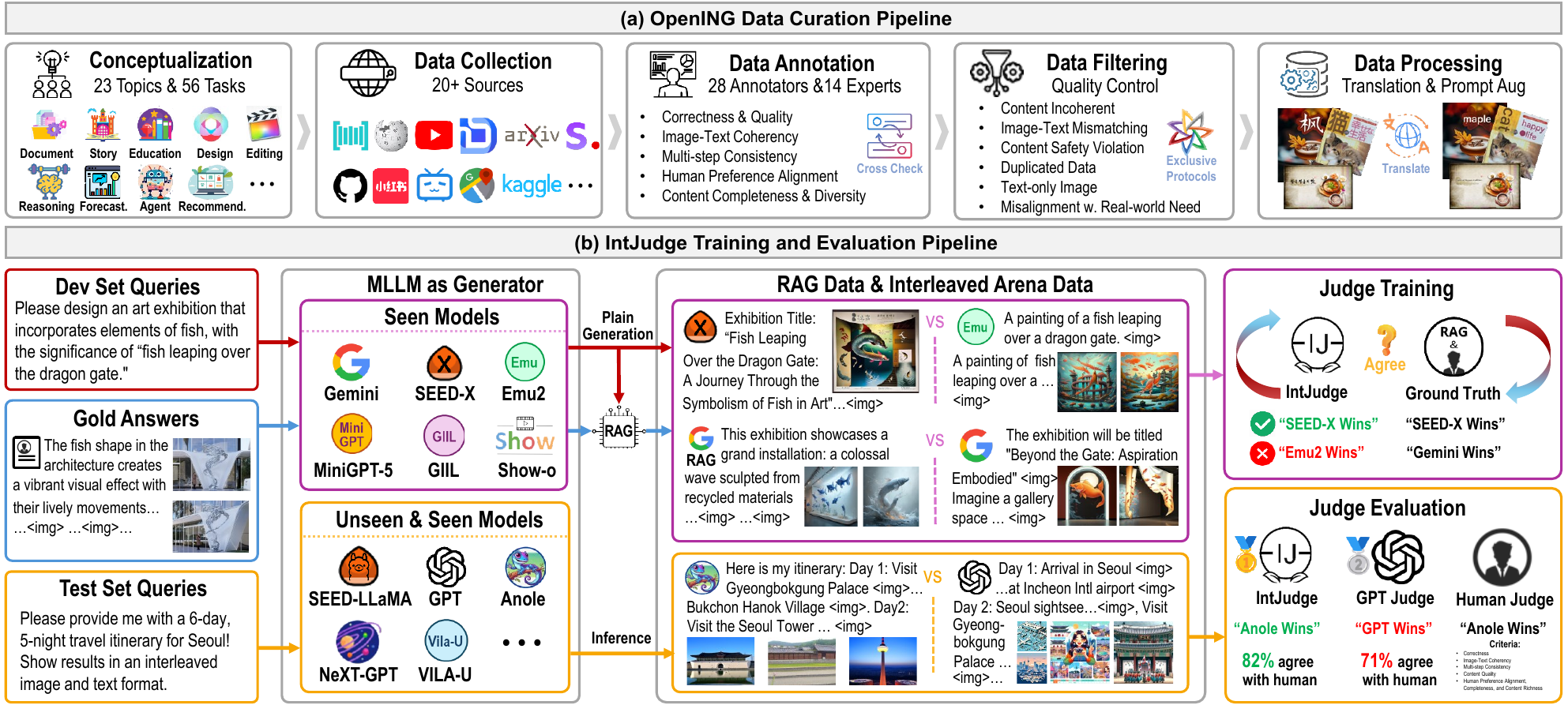}
   \caption{Overview of data curation and the proposed judge pipeline. (a) We construct our OpenING benchmark in a top-down manner, which involves five stages: conceptualization, data collection, annotation, filtering and processing. (b) We use the Dev Set of OpenING to train the proposed IntJudge and evaluate interleaved image-text generation on the Test Set to compare our IntJudge with human and GPT-4o.}
\label{fig:construction}
\vspace{-0.1cm}
\end{figure*}

\section{OpenING Benchmark}
\label{sec:method}

\subsection{Problem Definition}
\label{subsec:problem}

The task of interleaved image-text generation involves generating a sequence of text and images based on a given prompt. Each interleaved generation model (referred to as a multimodal agent) receives an input prompt $\mathbf{P}$, which can be text-only or include both text and images. 
The multimodal agent outputs an interleaved image-text sequence: $\mathbf{S} = [\mathbf{s}_1, \mathbf{s}_2, \dots, \mathbf{s}_N]$, where $N$ is the number of steps. Each element $\mathbf{s}_i = <\mathbf{T}_i, \mathbf{I}_i>$ in step $i$ consists of a text segment $\mathbf{T}_i$ and an image $\mathbf{I}_i$. 
Each $\mathbf{s}_i$ is generated based on the prompt $\mathbf{P}$ and all outputs history as $\mathbf{s}_i = f(\mathbf{P}, \mathbf{s}_1, \mathbf{s}_2, \dots, \mathbf{s}_{i-1})$, where $f$ denotes the the generation function of an agent. The objective is to find an optimal output sequence set $\mathbf{S}^*$:
\begin{equation}
\mathbf{S}^* = \arg\max_{\mathbf{S}} \, \prod_{i=1}^N p(\mathbf{s}^*_i \mid \mathbf{P}, \mathbf{s}^*_1, \dots, \mathbf{s}^*_{i-1}),
\end{equation}
\noindent where $\mathbf{s}^*_i$ in each step is semantically consistent with the input prompt while maintaining coherence throughout the entire sequence. The performance of an agent is evaluated based on how well the generated \(\mathbf{S}\) meets predefined criteria.





\subsection{Data Curation}

Collecting and annotating interleaved image-text data is inherently challenging due to the scarcity of high-quality data. It is particularly difficult to gather and pair multimodal data from disparate domains and ensure consistency~\cite{yang2024investigating}. We created OpenING over three months, with nearly 50 people involved in an efficient pipeline, which is shown in Fig.~\ref{fig:construction}(a).


\noindent\textbf{Topic Conceptualization.} 
%
With the assistance of multiple AI agents, we brainstormed and identified the most relevant real-world scenarios that require interleaved image-text generation. These insights were conceptualized into 23 meta-topics and divided into 56 specific tasks. 

\noindent\textbf{Data Collection and Annotation.} Interleaved image-text data was collected from more than 20 sources, including social media (e.g., Rednote\footnote{https://www.xiaohongshu.com}), video sharing websites (e.g., YouTube\footnote{https://www.youtube.com}), search engines (e.g. Google\footnote{https://www.google.com}), and open dataset platforms (e.g. OpenDataLab~\cite{he2024opendatalab}). The complete list of data sources is provided in the Supplementary Material. To ensure the highest data quality, a team of 28 professional annotators contributed under the supervision of 14 data experts. They performed efficient manual annotations using the IntLabel tool that we developed. Annotations were organized into a standard format, and each instance was limited to ten steps to avoid potential breaking of context constraints.




\noindent\textbf{Data Filtering and Quality Control.} We conducted cross-checks with annotators and data experts to ensure consistency, relevance, and coherence in each instance. Each task was required to include diverse sources and topics. In cases where data acquisition is complex, the annotators were instructed to supplement the data set with content generated by GPT-4o~\cite{openai2024hello} and Stable Diffusion XL~\cite{podellsdxl}.
To further enhance data quality, exclusive protocols are proposed to filter unqualified data. The qualified data were then redistributed to each respective task to reach the required quantity.



\noindent\textbf{Data Processing.} Post-processings were carried out to ensure the linguistic consistency of our benchmark. GPT-4o API is used to translate the annotated Chinese text to English, followed by data experts reviewing the accuracy. We also implemented image translation\footnote{https://github.com/zyddnys/manga-image-translator} to convert any Chinese characters to English in images. Finally, the prompts were refined for each task to achieve desired generation results, as detailed in the Supplementary Materials.


\noindent\textbf{Dataset Splitting.} As illustrated in Fig.~\ref{fig:data}, our OpenING benchmark ultimately includes 5,400 annotated instances, spanning 23 distinct meta-topics and 56 tasks. The annotated instances of OpenING are divided into a Dev Set (3,240 instances) and a Test Set (2,160 instances). The Dev Set supports the training of judge models, and the Test Set is used to evaluate the zero-shot performance of different models.
\section{IntJudge Model}
\label{sec:avai}


\subsection{Interleaved Arena}


Evaluating open-ended interleaved image-text generation is challenging due to the complexity of assessing multiple images and text, as well as the open-ended nature of generation, where multiple valid answers exist.  Given that pairwise comparison is more stable than subjective scoring~\cite{chenmllm}, we introduce Interleaved Arena, on which pairwise assessments were conducted using three evaluators: human judges, GPT-based judges, and the proposed IntJudge.

In the Interleaved Arena, interleaved outputs from agents on the OpenING Test Set are saved in a unified format. In each evaluation round, judges compare outputs from two anonymous agents and rate the interleaved outputs based on seven criteria: Correctness, Image-Text Coherency, Multi-step Consistency, Content Quality, Human Preference Alignment, Completeness, and Content Richness (see supplementary materials for more details). To balance evaluation reliability and efficiency, we propose a roulette matching algorithm to sample \( E \) distinct battle pairs for each data instance.

Let \( \mathcal{K} \) represent the set of tasks, and \( \mathcal{M} \) denote a set of Arena agents. Each task \( k \in \mathcal{K} \) has \( D_k \) data instances. A permutation \( \sigma_k \in A_{|\mathcal{M}|} \) is sampled by randomly shuffling the agent order, where \( A_{|\mathcal{M}|}\) is the set of all agent permutations. The set of sampled battle pairs is given by:
\begin{equation}
\mathcal{P}_k = \left\{ \left( \sigma_k(i \bmod |\mathcal{M}|),\ \sigma_k\left( (i+1) \bmod |\mathcal{M}| \right) \right) \right\},
\end{equation}where $i = 1, 2, \ldots, \mathcal{D}_k$.
Additional sampling rounds may be performed to obtain \( E \) distinct battle pairs for each data instance, where \( E \leq |\mathcal{M}|(|\mathcal{M}|-1)/2 \). To avoid duplications, a set \( \mathcal{R}_{k,d} \) is maintained in the $d$-th round, which stores all unique pairs sampled in previous rounds:
\begin{equation}
\mathcal{R}_{k,d} = \bigcup_{j=1}^{d-1} {(\sigma_{k,j}(a), \sigma_{k,j}(b))}.
\end{equation}
For the current pair \( \sigma_{k,d}(a) \) and \( \sigma_{k,d}(b) \), we enforce:
\begin{equation}
(\sigma_{k,d}(a), \sigma_{k,d}(b)) \notin \mathcal{R}_{k,d} \ \ \text{and} \ \ \sigma_{k,d}(a) \ne \sigma_{k,d}(b).
\end{equation}
Under the assumption of uniform distribution, we define the coverage time \( T_k \) to ensure all agents are evaluated in task $k$:
\begin{equation}
T_k = \left\lceil\frac{|\mathcal{M}|(|\mathcal{M}|-1)}{2E} \cdot \frac{D_k}{|\mathcal{P}_k|}\right\rceil,
\end{equation}
\noindent The overall expected coverage time is written as:
\begin{equation}
E[T] = \frac{|\mathcal{M}|}{2} \cdot H_{|\mathcal{M}|} = \frac{|\mathcal{M}|}{2} \cdot \left( \sum_{i=1}^{|\mathcal{M}|} \frac{1}{i} \right),
\end{equation}
where \( H_{|\mathcal{M}|} \) is the \( |\mathcal{M}| \)-th harmonic number. 

\subsection{Judge Pipelines}


\textbf{Human Judge.} In the human judge, annotators compare outputs from two multimodal agents for each input prompt and select a winner based on seven predefined criteria. The voting results are used to rank interleaved generation methods based on their win rates. Since the previous studies~\cite{zheng2023judging,chenmllm} noted that excessive ties cause inefficiency, our annotators are instructed to favor one agent in cases of a tie, denoting as \texttt{Tie(A)} or \texttt{Tie(B)} based on the slight preference.

\noindent\textbf{GPT-based Judge.} To enable scalability, we employ GPT-4o to automate the evaluation process. The GPT-4o is prompted to analyze interleaved outputs and decide the winner of each battle pair. Moreover, we use an additional prompt to obtain the score breakdown and explanations. While this strategy allows for scalable and explainable evaluation, GPT-based judges still have a high error rate due to their prior bias and lack of alignment with human preferences. GPT also raises privacy, data leakage, and cost concerns.


\noindent\textbf{IntJudge.} To address issues in GPT-based evaluators, we propose IntJudge to improve evaluation accuracy and alignment with human preferences. As an offline judge, IntJudge provides efficient large-scale evaluations with consistent criteria, ensuring fair and reproducible results for benchmarking interleaved image-text generation. Upon exploring several MLLMs including InternLM-XComposer2.5 (InternLMX2.5)~\cite{internlmxcomposer2_5} and Qwen2-VL~\cite{Qwen2VL}, 
we select Qwen2-VL-7B  
as the foundation model for training IntJudge, achieving an optimal balance between efficiency and accuracy.


\begin{table*}[h]
\centering
\small 
\setlength{\tabcolsep}{0.05mm}
\begin{tabularx}{\textwidth}{@{}>{\hspace{0.1pt}}l
    >{\centering\arraybackslash}p{1.25cm} 
    >{\centering\arraybackslash}p{1.25cm} >{\centering\arraybackslash}p{1.25cm} 
    >{\centering\arraybackslash}p{1.25cm} >{\centering\arraybackslash}p{1.25cm} 
    >{\centering\arraybackslash}p{1.25cm} >{\centering\arraybackslash}p{1.25cm} 
    >{\centering\arraybackslash}p{1.25cm} >{\centering\arraybackslash}p{1.25cm} 
    >{\centering\arraybackslash}p{1.25cm} >{\centering\arraybackslash}p{1.25cm} 
    >{\centering\arraybackslash}p{1.25cm} >{\centering\arraybackslash}p{1.25cm} 
    @{}}
\toprule
\multicolumn{1}{c}{\multirow{2.5}{*}{\textbf{Method}}} & \multicolumn{4}{c}{\textbf{Human Evaluation }} & \multicolumn{4}{c}{\textbf{GPT Evaluation }} & \multicolumn{4}{c}{\textbf{IntJudge Evaluation}} \\
\cmidrule(lr){2-5} \cmidrule(lr){6-9} \cmidrule(lr){10-13}
& \textbf{\scriptsize FDT} & \textbf{\scriptsize w/o Tie} & \textbf{\scriptsize w/ Tie (0)} & \textbf{\scriptsize w/ Tie (.5)} & \textbf{\scriptsize FDT} & \textbf{\scriptsize w/o Tie} & \textbf{\scriptsize w/ Tie (0)} & \textbf{\scriptsize w/ Tie (.5)} & \textbf{\scriptsize FDT} & \textbf{\scriptsize w/o Tie} & \textbf{\scriptsize w/ Tie (0)} & \textbf{\scriptsize w/ Tie (.5)} \\
\midrule
Human                & 83.28\% & 86.03\% & 68.17\% & 78.55\% & 82.49\% & 82.69\% & 82.03\% & 82.43\% & 87.46\% & 91.49\% & 75.49\% & 84.23\% \\
GPT-4o+DALL-E3       & 78.42\% & 81.39\% & 65.21\% & 75.15\% & 85.70\% & 85.99\% & 85.58\% & 85.82\% & 85.02\% & 86.92\% & 72.22\% & 80.68\% \\
Gemini1.5+Flux       & 65.57\% & 65.82\% & 49.31\% & 61.85\% & 71.75\% & 71.76\% & 71.12\% & 71.56\% & 68.30\% & 69.73\% & 54.47\% & 65.41\% \\
SEED-X               & 51.98\% & 49.49\% & 34.70\% & 49.65\% & 54.82\% & 55.12\% & 54.11\% & 55.03\% & 49.86\% & 49.58\% & 33.57\% & 49.72\% \\
Anole                & 51.90\% & 52.17\% & 36.46\% & 51.52\% & 53.36\% & 53.13\% & 52.58\% & 53.10\% & 53.42\% & 52.04\% & 33.92\% & 51.33\% \\
SEED-LLaMA           & 44.30\% & 42.12\% & 29.11\% & 44.56\% & 40.96\% & 40.87\% & 40.46\% & 40.96\% & 50.13\% & 47.71\% & 31.57\% & 48.48\% \\
Emu2                 & 40.89\% & 37.07\% & 23.42\% & 41.84\% & 41.72\% & 41.63\% & 40.58\% & 41.85\% & 36.28\% & 33.79\% & 21.87\% & 39.51\% \\
Show-o               & 36.28\% & 34.02\% & 21.63\% & 39.84\% & 30.77\% & 30.22\% & 29.61\% & 30.62\% & 31.49\% & 21.08\% & 12.48\% & 32.87\% \\
NExT-GPT             & 33.67\% & 26.93\% & 17.09\% & 35.36\% & 22.61\% & 22.39\% & 22.11\% & 22.74\% & 30.96\% & 21.70\% & 13.36\% & 32.58\% \\
MiniGPT-5            & 30.69\% & 26.72\% & 17.11\% & 35.09\% & 28.64\% & 28.37\% & 28.02\% & 28.64\% & 24.47\% & 15.46\% & 9.91\% & 27.85\% \\
GILL                 & 25.80\% & 19.57\% & 12.71\% & 30.23\% & 30.55\% & 30.24\% & 29.65\% & 30.62\% & 24.87\% & 19.72\% & 12.82\% & 30.32\% \\
\bottomrule
\end{tabularx}
\vspace{-0.3cm}
\caption{Comparison of model win rates evaluated by human, GPT-4o, and our IntJudge under FDT and different tie metrics. FDT: Force Dividing Tie metric. w/o Tie: Non-tie case. w/ Tie (0) and w/ Tie Tie (.5): Count a tie as 0 and 0.5 wins for a model in a battle, respectively.}
\label{tab:model_win_rates}
\vspace{-0.1cm}
\end{table*}

\subsection{Training of IntJudge}


To enhance IntJudge training, a Reference-Augmented Generation (RAG) approach is proposed to scale up the training data set. As shown in Fig.~\ref{fig:construction}(b), our IntJudge model is trained on the combination of human-annotated pairwise data from the Dev Set and the RAG pairs. In our RAG approach, models are provided with gold real-world answers from the Dev Set and prompted to generate responses based on these gold answers. Pairwise data are formed by pairing a plain generation result with an RAG-based output, where the RAG result is assigned as the winner. A bag of models, including $g$ seen interleaved generation methods are used for plain generation and RAG. The training objective is defined as:
\begin{equation}
\mathcal{L}_{\text{total}} = \lambda_1 \mathcal{L}_{\text{CE}} + \lambda_2 \mathcal{L}_{\text{CT}} + \lambda_3 \mathcal{L}_{\text{MSE}} + \lambda_4 \mathcal{L}_{\text{PR}},
\end{equation}
where \(\lambda_1\), \(\lambda_2\), \(\lambda_3\) and \(\lambda_4\) are weighting coefficients, $\mathcal{L}_{\text{CE}}$, $\mathcal{L}_{\text{CT}}$, $\mathcal{L}_{\text{MSE}}$, and $\mathcal{L}_{\text{PR}}$ are respectively cross-entropy, contrastive, MSE, and pairwise ranking losses. The trained IntJudge was tested in a zero-shot setting on both unseen and seen models to validate its generalizability. 



\section{Experiments}
\label{sec:experiments}

\subsection{Experimental Setup}



\noindent\textbf{Models.} We evaluated 10 representative interleaved methods, categorized into three types: \textbf{1) Integrated pipeline} combines independent text and image generation models, examples include GPT-4o+DALL$\cdot$E-3~\cite{openai2024hello,betker2023improving} and Gemini1.5+Flux~\cite{team2023gemini,blackforestlabs_flux}; \textbf{2) Two-stage generator}, such as Emu2~\cite{sun2024generative}, SEED-X~\cite{ge2024seed}, and Show-o~\cite{xie2024show}, has a unified model architecture but generates text and image in two separate stages; \textbf{3) End-to-end generator} produces image-text contents in a single stage, such models include GILL~\cite{koh2024generating}, NExT-GPT~\cite{wunext}, MiniGPT-5~\cite{zheng2023minigpt}, SEED-LLaMA~\cite{ge2023making}, and Anole~\cite{chern2024anole}. We keep GPT-4o+DALL$\cdot$E-3, Anole, SEED-LLaMA, and NExT-GPT as unseen models for IntJudge validation. The rest models are seen in IntJudge training.


\begin{figure}[t]
    \centering
    \includegraphics[trim=0.5cm 1.1cm 0 0,width=0.99\linewidth]{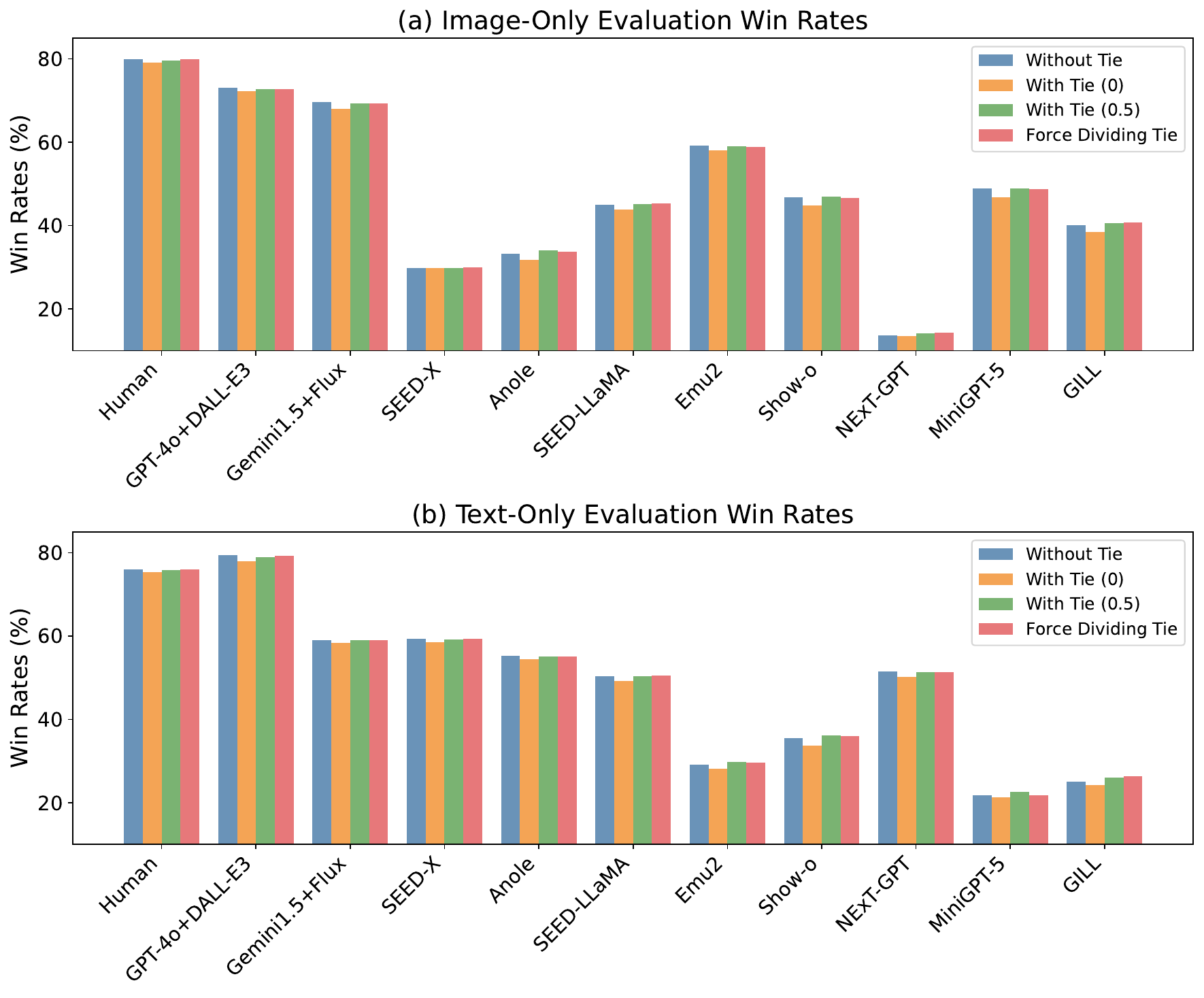}
    \caption{Model win rates under image-only and text-only settings across different models, ranked by human judgments.}
    \label{fig:it_only_evaluation}
    \vspace{-0.2cm}
\end{figure}

\noindent\textbf{Evaluation Metrics.}
Model performance are evaluated using two key metrics: win rate and agreement. \textbf{Win rate} indicates how often a model wins in pairwise comparisons. Four methods used to handle ties include 1)  Force Dividing Tie (FDT): We force judges to assign ties with a more leaning model in rules and prompts, ensuring that every comparison round results in a decisive outcome. A win is attributed to A if a tie favors model A (\texttt{Tie(A)}), likewise for B. This metric allows for clear rankings without ambiguity.
2) Without Tie (w/o Tie): Tied comparisons are excluded; only matches with a clear winner are considered; 3) With Tie counted as 0 (w/ Tie (0)): Ties are included but do not contribute to the win count of either model; 4) With Tie counted as 0.5 (w/ Tie (.5)): Each tie contributes half a win to both models.
\textbf{Agreement} measures the consistency between different evaluators (e.g., automated pipelines and human judgments) under the same tie-handling strategies. It tells the frequency with which the evaluators agree in their assessments.

\subsection{Overall Evaluation}
\noindent\textbf{Evaluation of Three Judges.} We conduct experiments to evaluate the performance of different models using win rate and agreement metrics. Table \ref{tab:model_win_rates} showcases the win rates of various models under different judge methods, including Human, GPT-based, and IntJudge-based Evaluations. The sampling round $E$ is set in 2 to form 4,320 battle pairs. It is found that the integrated pipelines like GPT-4o+DALL$\cdot$E-3 and Gemini 1.5+Flux consistently outperforms other models regardless of evaluators, while the end-to-end models like MiniGPT-5, GILL, and NExT-GPT perform less favarable.

\begin{figure}[t]
    \centering
    \includegraphics[trim=0.5cm 1.2cm 0 0,width=0.99\linewidth]{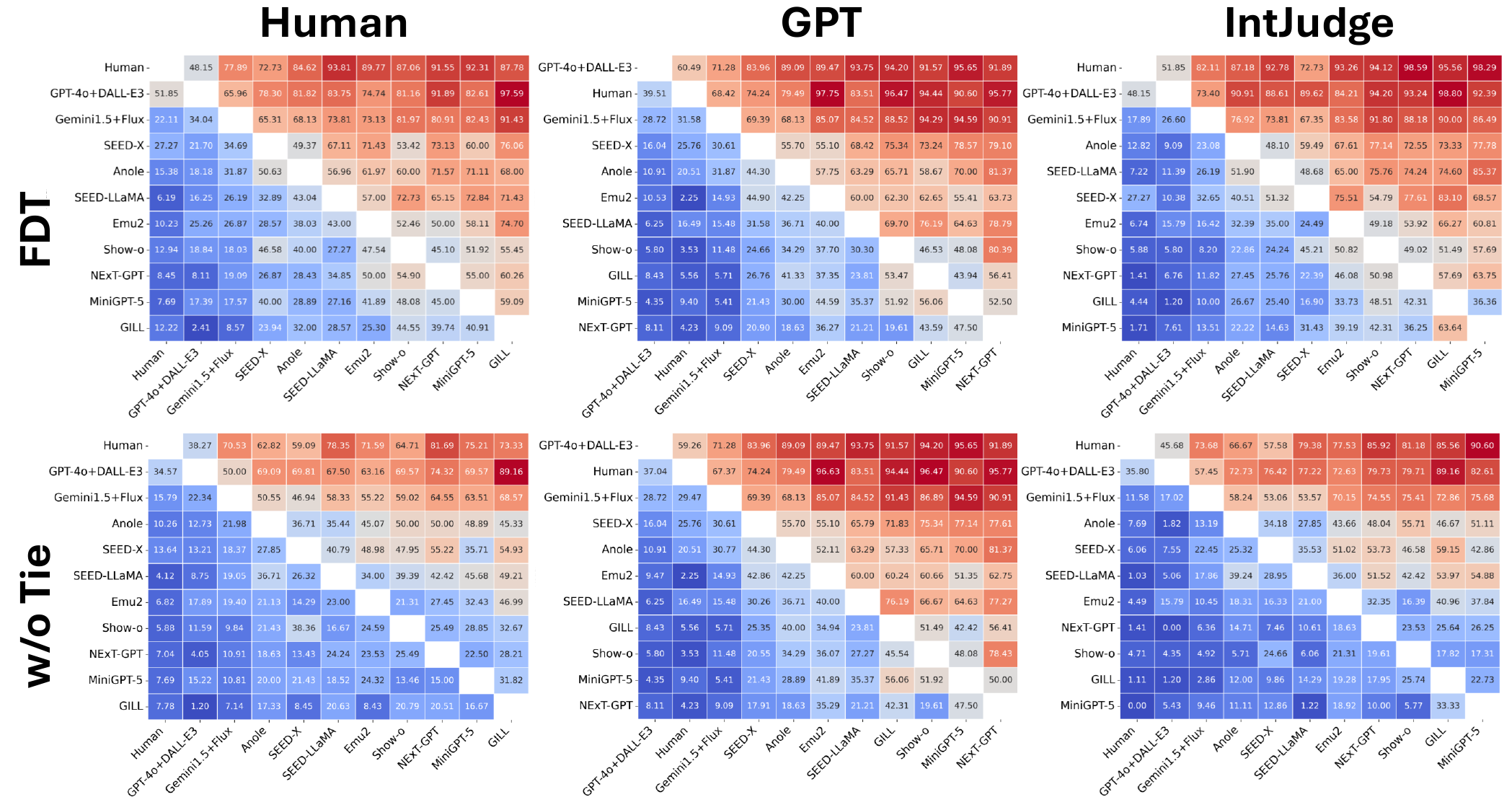}
    \caption{Win rate matrix of human and ten MLLM models, evaluated by human, GPT-4o, and our IntJudge, respectively.}
    \label{fig:model_pks_heatmap}
    \vspace{-0.2cm}
\end{figure}

\begin{table*}[t]
\centering
\small 
\setlength{\tabcolsep}{1.05mm}
\raggedright  
\begin{tabular}{@{}>{\hspace{3.5pt}}l*{13}{c}@{}} 
\toprule
\multicolumn{1}{c}{\multirow{2.5}{*}{\textbf{Evaluator}}} & \multicolumn{4}{c}{\textbf{FDT}} & \multicolumn{4}{c}{\textbf{w/ Tie}} & \multicolumn{4}{c}{\textbf{w/o Tie}} \\
\cmidrule(lr){2-5} \cmidrule(lr){6-9} \cmidrule(lr){10-13}
 & \textbf{Average} & \textbf{Seen} & \textbf{Unseen} & \textbf{HM}   & \textbf{Average} & \textbf{Seen} & \textbf{Unseen} & \textbf{HM} & \textbf{Average} & \textbf{Seen} & \textbf{Unseen} & \textbf{HM}\\
\midrule
Random & 49.83\% & 49.86\% & 49.79\% & 49.83\% & 32.60\% & 32.03\% & 33.18\% & 32.60\% & 50.00\% & 48.36\% & 51.89\% & 50.06\% \\
GPT-4o & 71.08\% & 73.33\% & 68.77\% & 70.98\% & 51.93\% & 54.95\% & 48.82\% & 51.70\% & 74.58\% & 77.54\% & 71.43\% & 74.36\% \\
InternLMX2.5-7B  & 56.81\% & 55.73\% & 57.92\% & 56.81\% & 40.26\% & 40.19\% & 40.33\% & 40.26\% & 61.05\% & 61.21\% & 60.97\% & 61.09\% \\
Qwen2-VL-7B & 61.61\% & 61.59\% & 61.63\% & 61.61\% & 32.81\% & 31.16\% & 34.50\% & 32.75\% & 80.77\% & 81.15\% & 80.23\% & 80.69\% \\
\multicolumn{1}{l}{IntJudge-7B (Ours)}  & \textbf{82.42\%} & \textbf{84.05\%} & \textbf{80.75\%} & \textbf{82.37\%} & \textbf{66.45\%} & \textbf{69.02\%} & \textbf{63.80\%} & \textbf{66.31\%} & \textbf{91.11\%} & \textbf{92.38\%} & \textbf{89.55\%} & \textbf{90.94\%} \\
\bottomrule
\end{tabular}
\vspace{-0.2cm}
\caption{Agreement rate between different MLLM-based judges and human judgments in different metrics. HM: Harmonic Mean.}
\label{tab:agreement}
\end{table*}


\noindent\textbf{Pairwise Model Performance.} Pairwise comparison results, evaluated by human, GPT-4o, and IntJudge, are shown in Fig. \ref{fig:model_pks_heatmap}. The heat map indicates win-loss relations, where warmer colors represent higher win rates and cooler colors vice versa. Notably, GPT-4o+DALL$\cdot$E-3 and Gemini1.5+Flux achieve the strongest win rates. Their generations are even comparable to human annotated output under GPT evaluation.

\begin{figure}[t]
    \centering
    \includegraphics[trim=0 1.4cm 0 0,width=0.98\linewidth]{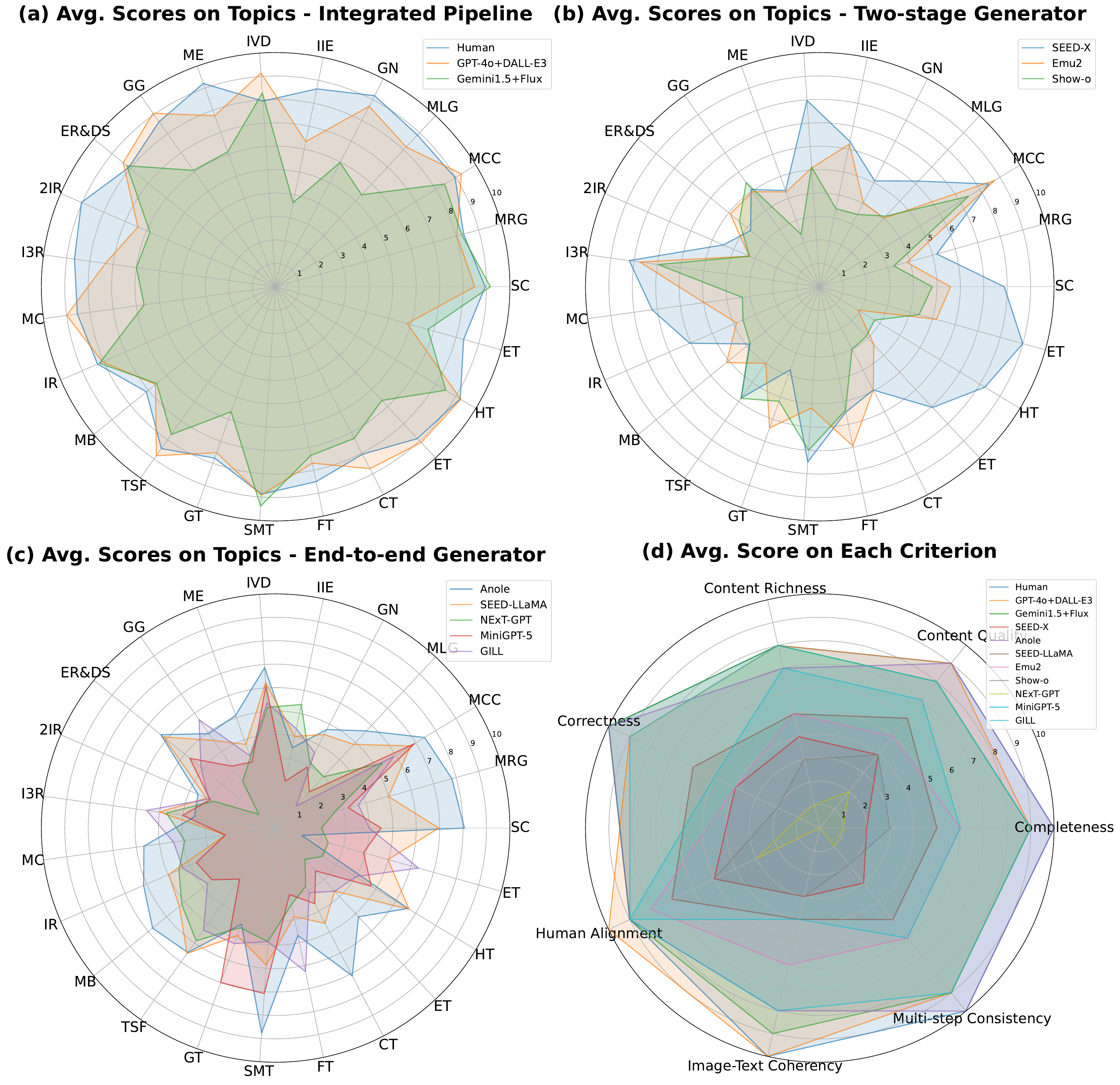}
    \caption{Evaluation results of GPT-based scores. (a)-(c): Average score of all criteria on each meta-topic for different kinds of models. (d) Average score of all meta-topics on each criterion.}
    \label{fig:radar}
    \vspace{-0.2cm}
\end{figure}


\noindent\textbf{Text-only and Image-only Evaluation.} To explore the impact of text and image on model performance, we evaluate models using text-only and image-only outputs on the same sampled pairs. Fig.~\ref{fig:it_only_evaluation} shows that MiniGPT-5 and GILL underperform primarily due to the low quality of their text outputs. SEED-X and NExT-GPT achieve higher win rates on text-only evaluation, however, the lower quality of generated images limits their ranking as shown in Table~\ref{tab:model_win_rates}. Text generated by GPT-4o even outperforms human-annotated content, which demonstrates its superior language capabilities. 


\noindent\textbf{GPT-based Scoring.} 
GPT-based evaluations are illustrated in Fig. \ref{fig:radar}, which provides an explainable performance analysis of different models. GPT-4o+DALL$\cdot$E-3 underperforms in meta-topics like Interactive Image Editing and Embodied-AI tasks, possibly due to limited training data in these categories. GPT-4o also exhibits bias toward its own outputs, scoring them 10 in human preference alignment, compared to an average score of 9 for human-annotated responses.

\noindent\textbf{Agreement with Human.} Table \ref{tab:agreement} shows the agreement between different evaluators and human judgments. We implement random guess (Random) as a baseline. The results indicate that IntJudge generally achieved higher agreement with human judgments (82.42\% in FDT) compared to GPT-based evaluation (71.08\% in FDT), suggesting its potential for scalable evaluation of interleaved image-text generation.

\subsection{Ablation Studies}
\label{sec:ablation}

\begin{figure}[t]
    \centering
    \includegraphics[trim=0.5cm 1.5cm 0 0,width=1\linewidth]{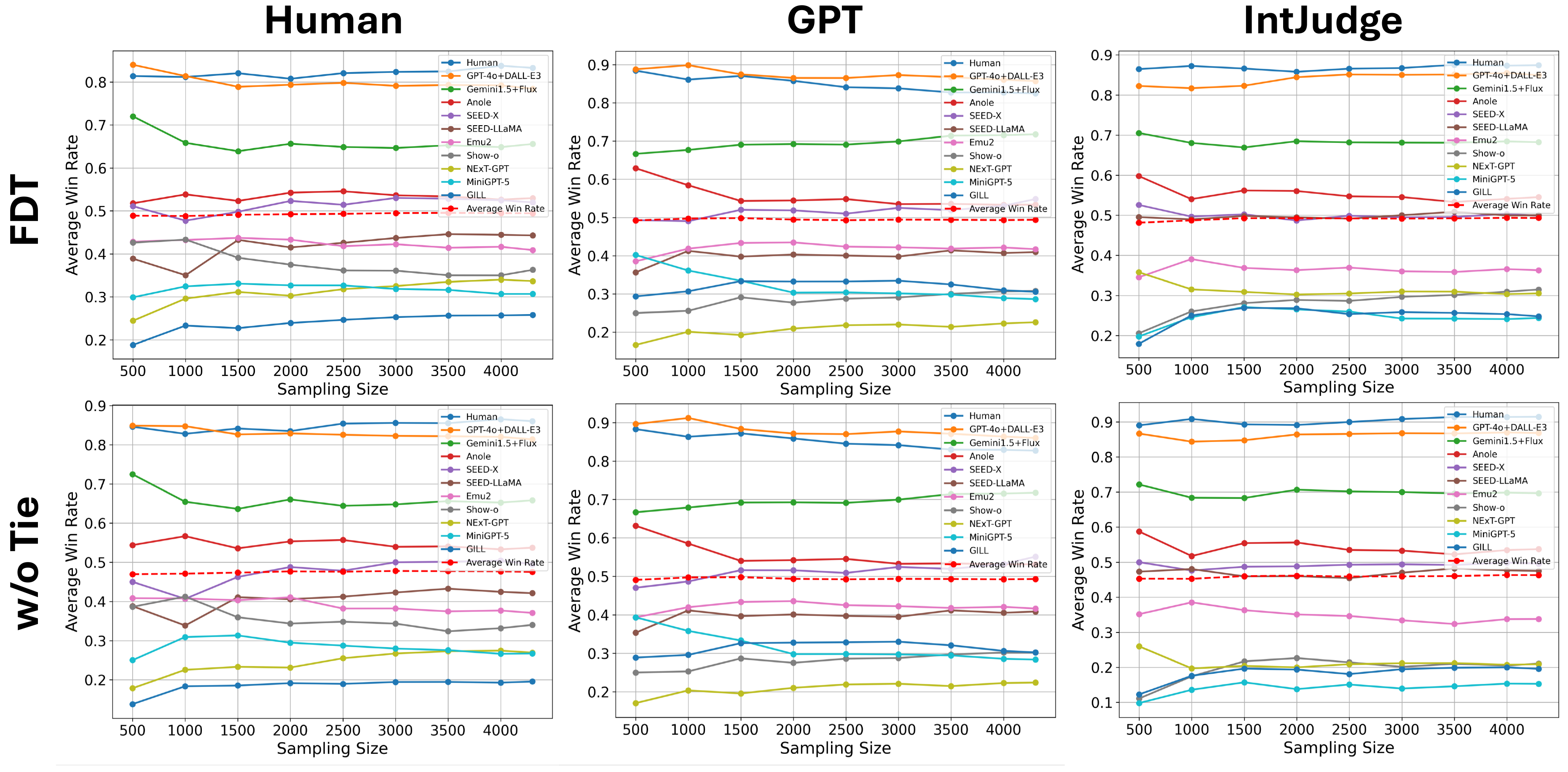}
    \caption{Effect of sampling size on evaluation reliability.}
    \label{fig:sampling_ablation}
\end{figure}

\noindent\textbf{Ablation on Sampling Size.} We evaluate the effect of sample size on evaluation stability and reliability. Fig. \ref{fig:sampling_ablation} illustrates the trend of win rates across varying sampling sizes. As the sample size increases, the win rates approach stability and show minimal variation across further increases. This stabilization suggests that our sampling number of 4,320 battle pairs is able to support the robust evaluation results.

\begin{figure}[t]
    \centering
    \includegraphics[trim=0.1cm 1.6cm 0 0,width=0.99\linewidth]{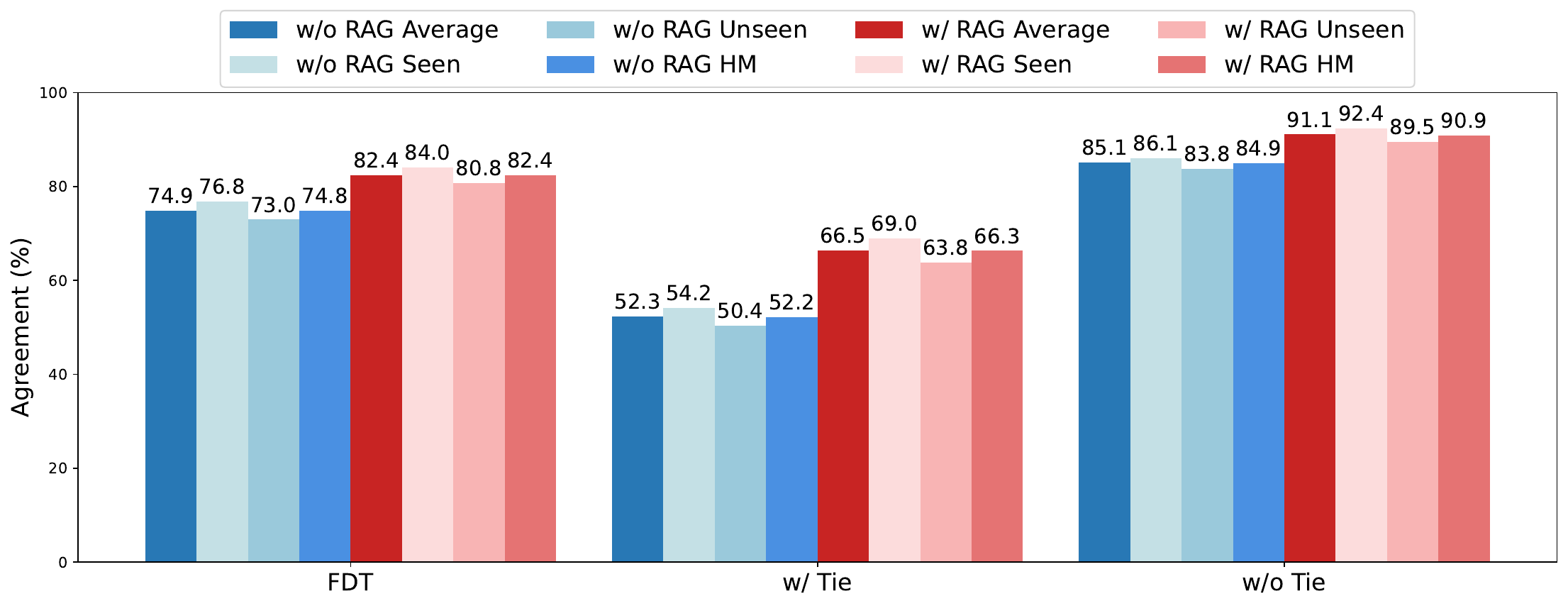}
    \caption{Comparison of agreement with human judgments for IntJudge trained without and with RAG data.}
    \label{fig:trainingablation}
      \vspace{-0.2cm}
\end{figure}

\noindent\textbf{Ablation on Judge Training Data.} We investigate the influence of incorporating RAG data on the performance of the IntJudge. The comparison is conducted between two training configurations: one that utilizes only the arena data (6,014 samples), the other being augmented with RAG data (25,982 samples). As illustrated in Fig. \ref{fig:trainingablation}, with RAG data included, the FDT agreement on unseen models increases by 7.8\%, demonstrating the effectiveness of our RAG-based strategy.

\noindent\textbf{Ablation on Image Generator.} We sample 200 data instances from all tasks to assess the influence of image generators on interleaved performance. Table \ref{tab:image_only_evaluation} compares basic text generation methods paired with different image generators. The results suggest that image generators greatly affect the quality of interleaved generation. For example, performance improves greatly when text models are paired with Flux-dev over other image models. It is also noted that Flux-dev has slower generation efficiency despite better image quality than Flux-schnell.


\begin{table}[t]
\centering
\small
\setlength{\tabcolsep}{1.3mm}
\begin{tabular}{p{0.33\linewidth} p{0.1\linewidth} p{0.1\linewidth} p{0.13\linewidth} p{0.145\linewidth}}
\toprule
\multicolumn{1}{c}{\textbf{\small Method}}& \multicolumn{1}{c}{\textbf{\small FDT}} & \multicolumn{1}{c}{\textbf{\small w/o Tie}}& \multicolumn{1}{c}{\textbf{\small w/Tie (0)}} & \multicolumn{1}{c}{\textbf{\small w/ Tie (.5)}}\\
\midrule
Human+Human       & 88.39\%   & 92.23\%   & 84.82\%   & 88.84\%  \\
Human+Flux-dev    & 11.61\%   & 7.77\%   & 7.14\%   & 11.16\%   \\
\midrule
GPT+DALL$\cdot$E-3         & 49.51\%   & 45.10\%  & 22.33\%  & 47.57\% \\
GPT+Flux-dev      & 50.49\%   & 54.90\%   & 27.18\%   & 52.43\%   \\
\midrule
Gemini+Flux-schnell   & 41.25\%   & 41.43\%  & 23.39\%  & 42.14\% \\
Gemini+Flux-dev   & 58.75\%   & 58.57\%   & 39.11\%   & 57.86\%   \\
\midrule
SEED-X+SEED-X     & 9.82\%    & 5.15\%   & 4.46\%   & 11.16\% \\
SEED-X+Flux-dev   & 90.18\%   & 94.85\%   & 82.14\%   & 88.84\%   \\
\bottomrule
\end{tabular}
\vspace{-0.1cm}
\caption{Evaluation results of interleaved content when basic text output combined with different image generation models.}
\label{tab:image_only_evaluation}
\end{table}

\begin{figure}[t]
    \centering
    \includegraphics[trim=0cm 0.3cm 0 0.1cm,width=0.99\linewidth]{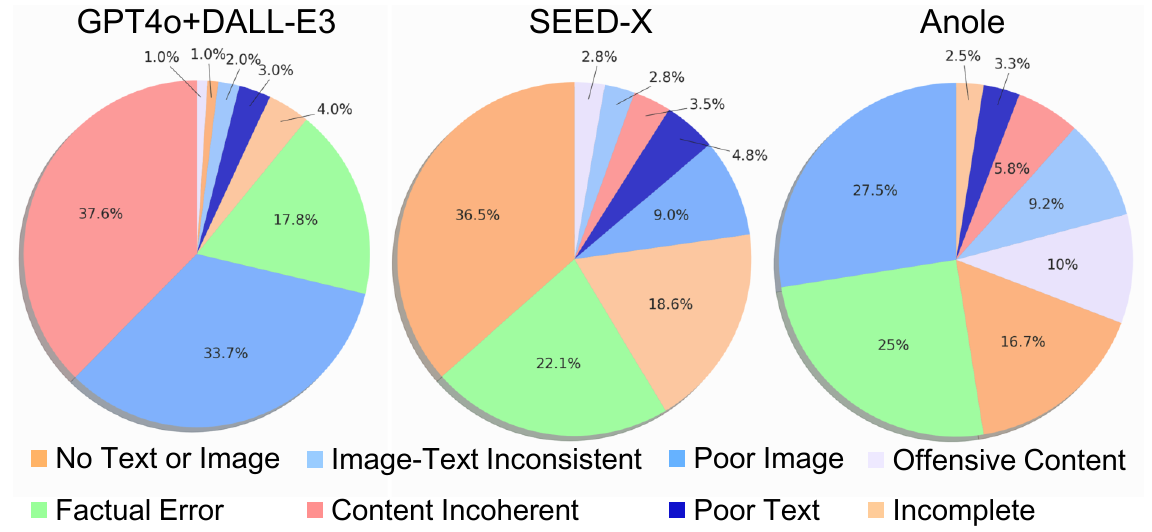}
    \caption{Error distribution of three models: GPT-4o+DALL$\cdot$E-3 (integrated), SEED-X (two-stage), and Anole (end-to-end).}
    \label{fig:error_analysis}
\end{figure}

\subsection{Analysis and Discussions}


\noindent\textbf{Error Analysis.} Error analysis on a set of 200 instances where the three types of model perform poorly compared to humans is shown in Fig. \ref{fig:error_analysis}. GPT-4o+DALL$\cdot$E-3 suffers from incoherency and inconsistency in content, possibly due to the limited capability of DALL$\cdot$E-3 to generate multiple images of the same style. Poor image quality is a major issue Anole faces, which may be attributed to the limited amount of data for fine-tuning of image generation. Although most SEED-X outputs contain multiple types of errors, the absence of text or image content remains the primary issue.


\noindent\textbf{No-Image and No-Text Ratios.} The no-image, no-text, and no-image-and-text ratios are listed in Table~\ref{tab:noratio}, which indicates the proportion of instances where models fail to generate visual content, textual content, or both. The near-zero failure rates of Human, GPT-4o+DALL$\cdot$E-3, and Genimi1.5+Flux (excluding policy-restricted senstive cases) indicate their consistent multimodal generation. Models like SEED-X and NExT-GPT showed high no-image ratios, likely due to their poorer instruction adherence and generation ability. These findings suggest that for a model to achieve high rankings on OpenING, its generated interleaved content must be of high quality in both images and text.



\begin{table}[t]
\small
\centering
\resizebox{0.95\linewidth}{!}{%
\begin{tabular}{lccc}
\toprule
\multicolumn{1}{c}{\multirow{2.5}{*}{\textbf{Method}}} & \multicolumn{3}{c}{\textbf{Ratio}} \\
\cmidrule(lr){2-4} & \textbf{No-Image} & \textbf{No-Text} & \textbf{No-I\&T} \\
\midrule
Human & 0.00\% & 0.00\% & 0.00\% \\
GPT-4o+DALL$\cdot$E-3 & 0.23\% & 0.00\% & 0.00\% \\
Gemini1.5+Flux & 0.09\% & 0.09\% & 0.09\% \\
SEED-X & 23.17\% & 4.64\% & 4.46\% \\
Anole & 19.46\% & 2.00\% & 1.30\% \\
SEED-LLaMA & 4.77\% & 0.05\% & 0.00\% \\
Emu2 & 0.00\% & 15.10\% & 0.00\% \\
Show-o & 0.00\% & 7.74\% & 0.00\% \\
NExT-GPT & 43.97\% & 0.09\% & 0.09\% \\
MiniGPT-5 & 0.27\% & 26.54\% & 0.00\% \\
GILL & 19.95\% & 13.43\% & 0.28\% \\
\bottomrule
\end{tabular}
}
\caption{The ratios of No-Image, No-Text, and No-Image-and-Text (No-I\&T) outputs relative to the total number of generated samples.}
\label{tab:noratio}
\end{table}



\noindent\textbf{Findings and Discussions.} We discuss key findings from our experiments to inspire future works: \textbf{1)} All generative models ranked lower than human in interleaved generation. The unified end-to-end models lagged significantly behind integrated pipelines, which combine more developed foundation models. The unified two-stage generation methods also need further improvements. \textbf{2)} Natural images consistently outperform generated ones, indicating the significant challenges of high-quality image generation. \textbf{3)} The quality of GPT-generated text can be comparable to or even exceed that of human-annotated text, demonstrating the effectiveness of LLMs in producing rich and informative textual content. \textbf{4)} Image generation has a great impact on interleaved generation. The quality of interleaved content improves markedly when text models are paired with more advanced image models. \textbf{5)} Large-scale data is crucial for training judge models. By scaling up data beyond manual annotations, our RAG method contributed to training a more robust judge model.
\vspace{-0.15cm}
\section{Conclusion}
\label{sec:conclusion}

We introduce OpenING, a comprehensive benchmark for evaluating open-ended interleaved image-text generation. OpenING addresses the limitations of existing benchmarks by a wider coverage of more diverse data and tasks grounded in real-world scenarios. To better assess open-ended multimodal generation, we propose IntJudge, a robust judge model trained on both human-annotated and RAG-based data from the Dev Set of OpenING. It is expected that our IntJudge can serve as a reward model in future RL-based (e.g., GRPO) generative models. Evaluation of various interleaved generation methods on the Test Set of OpenING reveals the challenges of generating coherent and high-quality interleaved image-text content. Ablation studies reaffirm the effectiveness of our RAG-based data pipeline for training IntJudge. Looking ahead, expanding the size and diversity of interleaved generation benchmarks could unlock even greater real-world potential and impact. We anticipate that our OpenING will inspire future research in MLLMs and benefit the development of multimodal evaluation models.



\appendix
\label{sec:appendix}

\setcounter{figure}{9}
\setcounter{table}{5}
\setcounter{equation}{6}


\twocolumn[{%
 \centering
 \Large \textbf{OpenING: A Comprehensive Benchmark for Judging \\ Open-ended Interleaved Image-Text Generation} \\[0.3em]Supplementary Material\\[1.5em]
}]

\noindent In this supplementary material, we provide additional information, discussions, and results in support of the primary text, which are organized as follows:

\noindent
Sec.~\ref{sec:OpenING} includes the details of OpenING and data curation.

\noindent
Sec.~\ref{sec:IntJudge} presents the details of IntJudge.

\noindent
Sec.~\ref{sec:evaluation} presents the details of evaluation.

\noindent
Sec.~\ref{sec:experimental} provides the details of our experiments and additional evaluation results.

\noindent
Sec.~\ref{sec:extension} introduces the extended experiments and results on fine-tuning MLLMs using our OpenING benchmark.

\noindent
Sec.~\ref{sec:limitations} discusses the limitations of this study.

\noindent The indexings of all figures, tables, and equations continue from the primary text.

\section{Details of OpenING and Data Curation}
\label{sec:OpenING}

\subsection{Hierarchical Structure of OpenING}
\label{sec:hso}


We present in Table~\ref{tab:class_structure1} the 56 specific tasks from OpenING that derived from 23 meta-topics. The number of Instances (\# of Ins.),  Meta-topic names, and capabilities of MLLMs evaluated in testing are provided. Based on the required annotation skills, the 56 tasks are divided into 38 common tasks and 18 hard tasks. The common tasks are annotated by 28 professional annotators, instructed and supervised by 14 data experts. The annotations of hard tasks, which require specific domain knowledge and special data reasoning and processing techniques, are conducted by the 14 data experts. 


\begin{table*}[ht]
\centering
\small
\caption{Details of the 38 common tasks and 18 hard tasks in our OpenING Benchmark.}
\vspace{-0.3cm}
\begin{tabular}{|l|c|l|l|}
\hline
\multicolumn{1}{|c|}{\textbf{Task Name}} & \multicolumn{1}{|c|}{\textbf{\# of Ins.}} & \multicolumn{1}{|c|}{\textbf{Meta-Topic}} & \multicolumn{1}{|c|}{\textbf{Capabilities}} \\ \hline
\multicolumn{4}{|c|}{\textbf{Common Tasks}}\\
\hline
Travel Guide Generation & 100 & Multimodal Report Generation & Content Creation \\ \hline
Museum Guide Book Generation & 100 & Multimodal Report Generation & Content Creation \\ \hline
Dynamic Biography Generation & 100 & Multimodal Report Generation & Content Creation \\ \hline
Multimodal Report Completion & 100 & Multimodal Content Completion & Content Completion \\ \hline
Interior Design & 100 & Interactive Visual Design & Design \& Brainstorming \\ \hline
Architectural Design & 100 & Interactive Visual Design & Design \& Brainstorming \\ \hline
Art and Exhibition Design & 100 & Interactive Visual Design & Design \& Brainstorming \\ \hline
Product Design & 100 & Interactive Visual Design & Design \& Brainstorming \\ \hline
Interactive Graphic Advertisement Editing & 100 & Interactive Visual Design & Design \& Brainstorming \\ \hline
Geometric Problem Test & 100 & Multimodal Exam & Education Assistant \\ \hline
Circuit Problem Test & 100 & Multimodal Exam & Education Assistant \\ \hline
Mind Map Generation & 100 & Graph Generation & Summary Agent \\ \hline
Figure Relationship Diagram Generation & 100 & Graph Generation & Summary Agent \\ \hline
Multi-view News Generation & 100 & Event Reasoning \& Deductive Simulation & Deductive Simulation \\ \hline
Dynamic Sports Event Analysis & 100 & Event Reasoning \& Deductive Simulation & Deductive Simulation \\ \hline
Interactive Historical Interpretation & 100 & Event Reasoning \& Deductive Simulation & Deductive Simulation \\ \hline
Unsolved Mysteries Exploration & 100 & Event Reasoning \& Deductive Simulation & Deductive Simulation \\ \hline
Multimodal Biological Reasoning & 100 & 2D Image Reasoning & Visual Reasoning \\ \hline
Multimodal Landscape Reasoning & 100 & 2D Image Reasoning & Visual Reasoning \\ \hline
Multimodal Analogy Reasoning & 100 & 2D Image Reasoning & Visual Reasoning \\ \hline
Interactive Jigsaw Puzzle & 100 & 2D Image Reasoning & Visual Reasoning \\ \hline
Interactive Multi-concept Image Composition & 100 & Multimodal Information Summary & Summary Agent \\ \hline
Interactive Film and Television Recommendation & 100 & Multimodal Information Recommendation & Information Recommendation \\ \hline
Interactive Goods Recommendation & 100 & Multimodal Information Recommendation & Information Recommendation \\ \hline
Interactive Food Recommendation & 100 & Multimodal Information Recommendation & Information Recommendation \\ \hline
Business Scenarios Brainstorming & 100 & Multimodal Brainstorming & Design \& Brainstorming \\ \hline
Academic Scenarios Brainstorming & 100 & Multimodal Brainstorming & Design \& Brainstorming \\ \hline
Multimodal Action Anticipation & 100 & Multimodal Time Series Forecasting & Time Series Forecasting \\ \hline
Visual Traffic Forecasting & 100 & Multimodal Time Series Forecasting & Time Series Forecasting \\ \hline
Interactive Remote Sensing Image Rendering & 100 & Geographical Tasks & Domain-specific Applications \\ \hline
Interactive Street View Image Rendering & 100 & Geographical Tasks & Domain-specific Applications \\ \hline
Urban Planning and Development Simulation & 100 & Geographical Tasks & Domain-specific Applications \\ \hline
Plog and Social Media Content Generation & 100 & Social Media Tasks & Domain-specific Applications \\ \hline
Interactive Virtual Try-on & 100 & Fashion Tasks & Domain-specific Applications \\ \hline
Multimodal Dressing Suggestion & 100 & Fashion Tasks & Domain-specific Applications \\ \hline
Fashion Trend Forecasting & 100 & Fashion Tasks & Domain-specific Applications \\ \hline
Multimodal Recipe Generation & 100 & Cooking Tasks & Domain-specific Applications \\ \hline
Multimodal Cooking Assistant & 100 & Cooking Tasks & Domain-specific Applications \\ \hline
Interactive Science Popularization & 100 & Educational Tasks & Education Assistant \\ \hline
Fitness and Health Consulting & 100 & Healthcare Tasks & Domain-specific Applications \\ \hline
\multicolumn{4}{|c|}{\textbf{Hard Tasks}}\\
\hline
Multimodal Action Anticipation & 100 & Multimodal Time Series Forecasting & Time Series Forecasting \\ \hline
Story Writing & 100 & Storybook Creation & Content Creation \\ \hline
Fiction Writing & 75 & Storybook Creation & Content Creation \\ \hline
Document with Layout Generation & 100 & Multimodal Layout Generation & Content Creation \\ \hline
Slide with Note Generation & 50 & Multimodal Layout Generation & Content Creation \\ \hline
Storybook Completion & 100 & Multimodal Content Completion & Content Completion \\ \hline
Web GUI Navigation & 100 & GUI Navigation & Interactive Agent \\ \hline
In-APP GUI Navigation & 100 & GUI Navigation & Interactive Agent \\ \hline
Cross-APP GUI Navigation & 100 & GUI Navigation & Interactive Agent \\ \hline
OS GUI Navigation & 75 & GUI Navigation & Interactive Agent \\ \hline
Interactive Portrait Image Editing & 100 & Interactive Image Editing & Interactive Agent \\ \hline
Interactive Landscape Image Editing & 100 & Interactive Image Editing & Interactive Agent \\ \hline
Interactive Novel View Synthesis & 100 & Image-based 3D Reasoning & Visual Reasoning \\ \hline
Dream Analysis and Scene Reconstruction & 100 & Event Reasoning \& Deductive Simulation & Deductive Simulation \\ \hline
Interactive Multi-concept Image Composition & 100 & Multi-concept Composition & Summary Agent \\ \hline
Scientific Brainstorming & 50 & Multimodal Brainstorming & Design \& Brainstorming \\ \hline
Chat with Memes & 100 & Social Media Tasks & Domain-specific Applications \\ \hline
Autonomous Driving and In-door Navigation & 50 & Embodied-AI Tasks & Domain-specific Applications \\ \bottomrule
\end{tabular}
\label{tab:class_structure1}
\end{table*}

\subsection{Data Sources}
\label{sec:source}

We list all sources where meta-data are collected for the annotation of our OpenING benchmark. Annotators arrange the images collected from the sources into the standardized multi-step format and annotate the text for each corresponding image. The details of the data source are presented in Table~\ref{tab:data_source}, including the ID number of each task. We also provide examples of the source of data instances to show how the desired data are searched from a certain platform. 




\subsection{Exclusive Protocols for Data Filtering}

The maximum number of steps per instance was limited to ten to ensure usability with context restraints. The instances with more than ten steps were excluded. All queries and answers are annotated manually via our developed tool, IntLabel. We implemented a set of exclusive protocols for filtering unqualified data, which include: 

\begin{enumerate}[label=\arabic*)]
    \item Removing data without coherence.
    \item Removing mismatched text and images.
    \item Removing data involving violence, offensive content and other content safety concerns.
    \item Removing duplicated data.
    \item Avoiding images consisting of only text.
    \item Removing data that is inconsistent with real-world logic.
    \item Avoiding content misaligned with real user needs.
\end{enumerate}

\noindent We repeat the above data collection and filtering process for each task until the number of instances reaches our target.

\subsection{OpenING Data}
Illustrations of the representative example of the 23 meta-topics are provided in Fig. \ref{fig:meta_topics} to showcase the diversity and complexity of tasks in our OpenING benchmark. These examples highlight the variety of interleaved image-text data that OpenING encompasses, demonstrating the challenges and capabilities required for effective interleaved image-text generation.

\subsection{Annotation of OpenING}
\label{ssub:IntLabel}
\begin{figure*}[htbp]
	\begin{center}
		\includegraphics[width=1.01\textwidth]{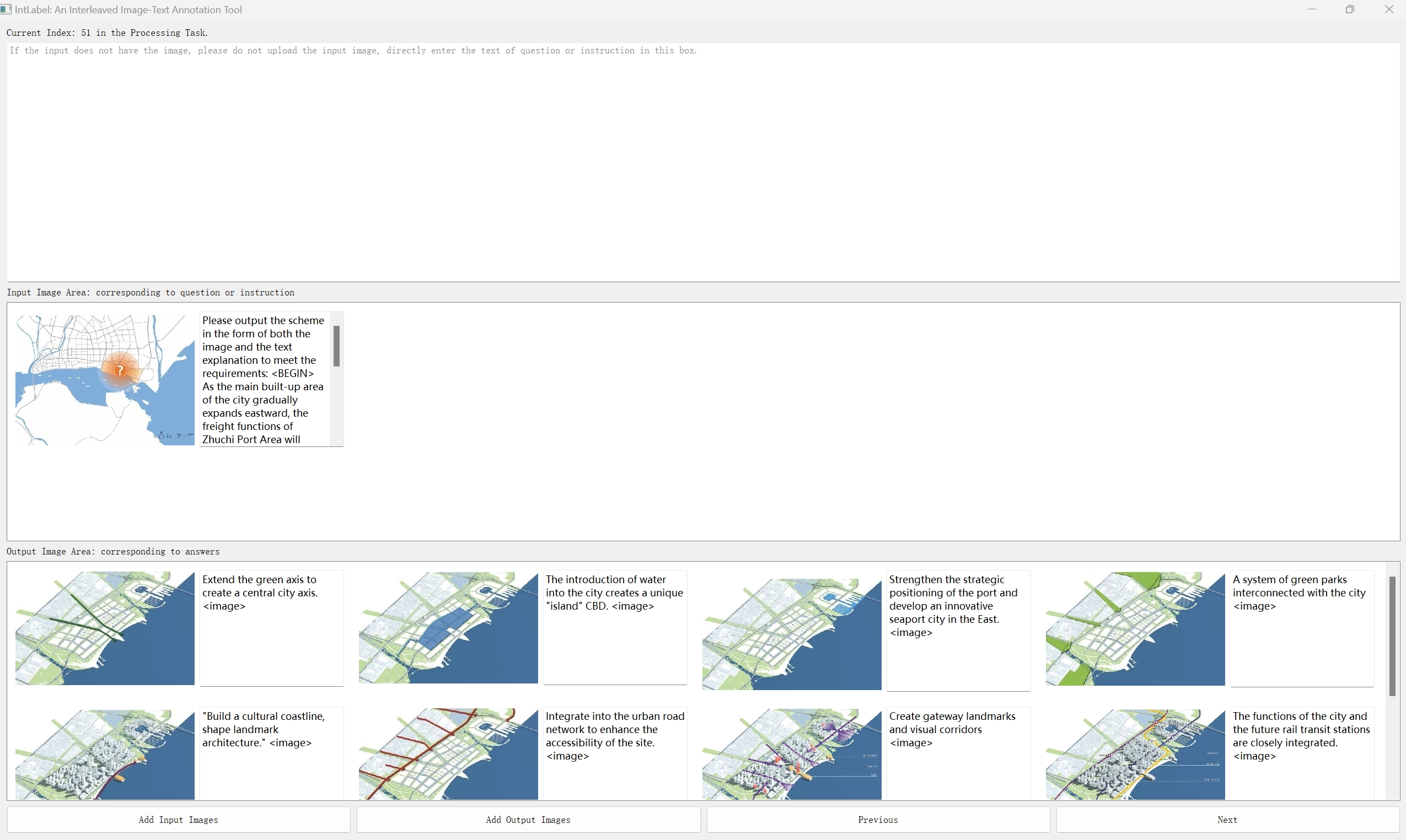}
        \caption{Interface of IntLabel, which shows a case where data is entered to finish annotation for an instance. }
		\label{fig:intlabel2}
	\end{center}
\end{figure*}

To facilitate the annotation process and ensure consistent annotations standards across annotators, we developed and opensourced an annotation tool for interleaved image-text data labeling. Our IntLabel annotation tool is developed based on PyQt\footnote{https://www.riverbankcomputing.com/software/pyqt}, and the IntLabel GUI is presented in Fig.~\ref{fig:intlabel2}. All queries and answers in OpenING are annotated, checked, and refined manually via IntLabel. The annotated JSONL files and corresponding images are kept in the same folder.  


\subsection{Task Abbreviations}
Given the large number of tasks in our benchmark and the wide range of methods evaluated, the abbreviations of the 23 meta-topics and 56 tasks are provided respectively in Table~\ref{tab:meta_topics} and Table~\ref{tab:dual_column_tasks}.

\begin{figure*}[htbp]
	\begin{center}
		\includegraphics[width=0.98\textwidth]{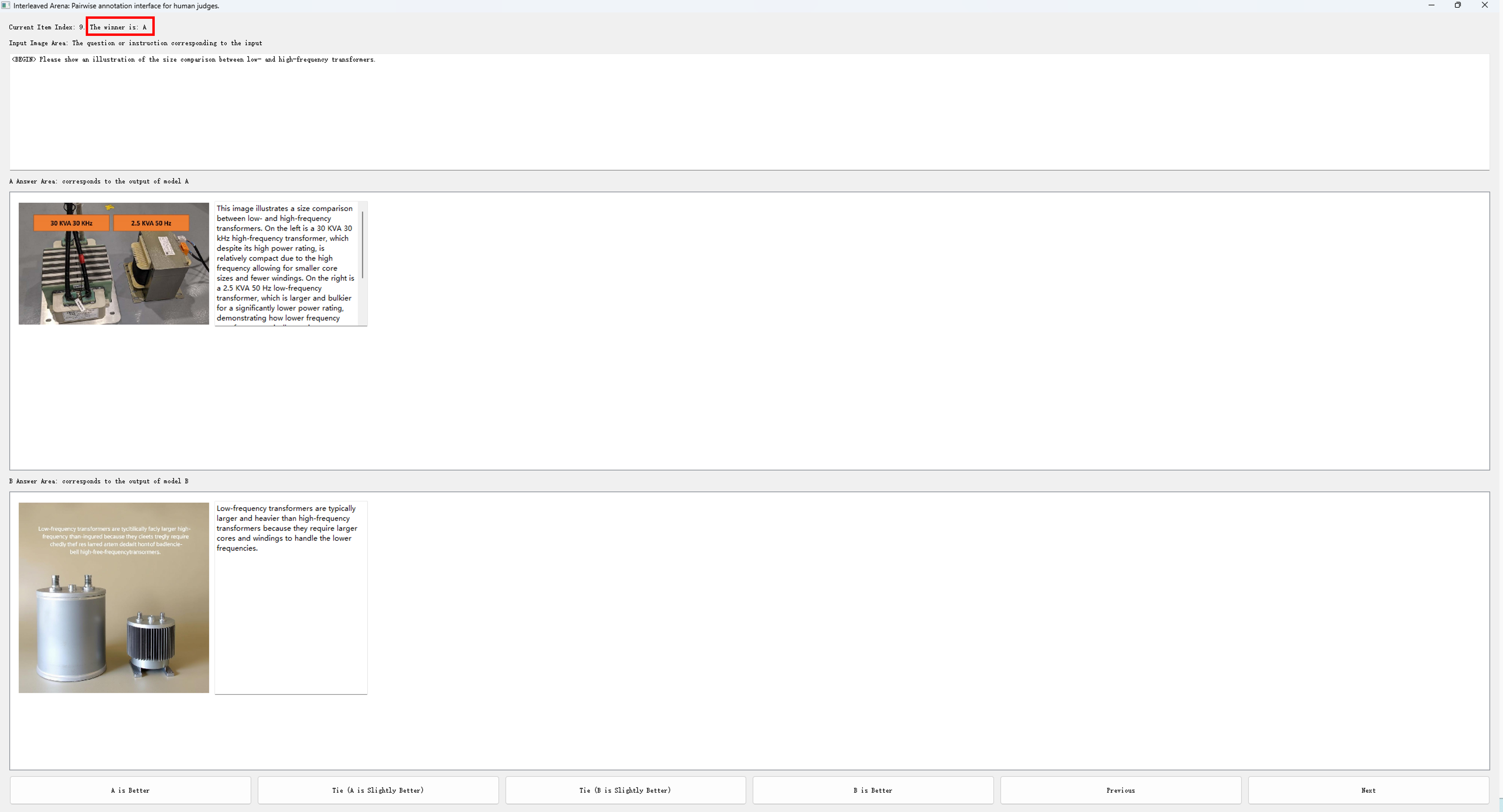}
		\caption{Annotation Interface of Interleaved Arena for human judges to compare the anonymous outputs of model A and model B. Human evaluators are instructed to select "A is Better" or "B is Better" to choose a winner for the pairwise comparison. When the two outputs are similar in quality and a tie option has to be chosen, human evaluators are instructed to select "Tie (A is slightly Better)" (\texttt{Tie(A)}) or "Tie (B is slightly Better)" (\texttt{Tie(B)}). Zoom in for a better experience.}
		\label{fig:intarena}
	\end{center}
\end{figure*}

\section{Details of IntJudge and Arena Data}
\label{sec:IntJudge}
\subsection{Annotation of Interleaved Arena}
The Interleaved Arena is introduced as an evaluation framework specifically to address the challenges of assessing open-ended interleaved image-text generation. Evaluating interleaved image-text generation is difficult due to: 1) Multiple images and text may need to be assessed at the same time; 2) The answer may be open-ended, where there is no single correct answer (multiple solutions exist for a query). Comparative evaluation has been demonstrated to offer greater stability and reliability than subjective scoring~\cite{chenmllm,zheng2023judging,li2024k}. Therefore, pairwise comparison is employed to ensure consistency and accuracy. The Interleaved Arena facilitates this by supporting evaluations from human judges, GPT-based judges, and the proposed IntJudge. The Interleaved Arena consists of two main components: 1) A sampling strategy to select pairwise comparison from all available interleaved generation methods fairly, and 2) An annotation interface for human judges to conduct evaluations manually. As shown in Fig.~\ref{fig:intarena}, the annotation interface is developed using PyQt and will be made available to researchers for facilitating future development. Using the annotation interface of Interleaved Arena, annotators are tasked with comparing anonymous outputs from two multimodal agents for each input prompt and deciding which is the winner based on seven predefined criteria. The vote results are used to rank interleaved generation models based on their win rates in Interleaved Arena. Since the previous studies~\cite{zheng2023judging,chenmllm} noted that too many ties cause inefficiency, our annotators are instructed to appoint a more leaning output when choosing a tie for a battle pair, denoted as \texttt{Tie(A)} or \texttt{Tie(B)}.

\subsection{Training of IntJudge}
The IntJudge’s architecture is based on Qwen2-VL-7B. Hyperparameters include cutoff length 16,240, batchsize 8, learning rate 1.0e-4, training epochs 20, training precision BF16, and training dataset (31,996 samples, with 6,014 arena data and 25,982 RAG data). As defined in the Eq. \ref{eq:total_loss} in the primary text (see Section 4.7), the total loss consists of cross-entropy loss \(\mathcal{L}_{\text{CE}}\), contrastive loss \(\mathcal{L}_{\text{CT}}\), MSE loss \(\mathcal{L}_{\text{MSE}}\), and pairwise ranking loss \(\mathcal{L}_{\text{PR}}\):
\begin{equation}
\mathcal{L}_{\text{total}} = \lambda_1 \mathcal{L}_{\text{CE}} + \lambda_2 \mathcal{L}_{\text{CT}} + \lambda_3 \mathcal{L}_{\text{MSE}} + \lambda_4 \mathcal{L}_{\text{PR}},
\label{eq:total_loss}
\end{equation}
\noindent where the loss weights are $\lambda_1 = 1.0$, $\lambda_2 =0.01$, $\lambda_3 = 0.01$, $\lambda_4 = 0.01$ respectively. 

Cross-entropy loss for language modeling is given by:
\begin{equation}
\mathcal{L}_{\text{CE}} = - \sum_{i=1}^{N} y_i \log(\hat{y}_i),
\end{equation}
where \( y_i \) is the ground truth token and \( \hat{y}_i \) is the predicted probability for token \( i \).

Contrastive Loss aligns image and text embeddings, which is written as:
\begin{equation}
\mathcal{L}_{\text{CT}} = - \frac{1}{N} \sum_{i=1}^{N} \log \frac{\exp(\text{sim}(\mathbf{z}_i^I, \mathbf{z}_i^T) / \tau)}{\sum_{j=1}^{N} \exp(\text{sim}(\mathbf{z}_i^I, \mathbf{z}_j^T) / \tau)},
\end{equation}
where \( \mathbf{z}_i^I \) and \( \mathbf{z}_i^T \) are the image and text embeddings for instance \( i \), \(\text{sim}(\cdot)\) represents the similarity (e.g., cosine similarity), and \(\tau\) is a temperature parameter.

Mean Squared Error Loss for image feature regression and reducing prediction errors is given by:
\begin{equation}
\mathcal{L}_{\text{MSE}} = \frac{1}{N} \sum_{i=1}^{N} \left( f(\mathbf{x}_i) - \hat{f}(\mathbf{x}_i) \right)^2,
\end{equation}
where \( f(\mathbf{x}_i) \) is the ground truth feature and \( \hat{f}(\mathbf{x}_i) \) is the predicted feature for image \( i \).

Pairwise Ranking Loss for precise rankings of outputs weights the relative quality of generated outputs in pairwise comparison, which is given by:
\begin{equation}
\mathcal{L}_{\text{PR}} = \sum_{i=1}^N \max(0, 1 - (f(x_i^+) - f(x_i^-))),
\end{equation}
where \( f(x_i^+) \) and \( f(x_i^-) \) represent the scores assigned to positive and negative examples, respectively. The combined total loss ensures language generation basis, multimodal understanding capabilities, ranking accuracy and consistency across similar inputs during training.

\section{Details of Evaluation}
\label{sec:evaluation}

\subsection{Prompt for Interleaved Generation}
Each interleaved generation prompt is carefully constructed by combining two main parts: task-specific prompts and model-specific prompts. Model-specific prompts are adjusted based on the individual characteristics of each model, providing hints and guidance to ensure coherent and consistent outputs across multiple steps, as illustrated in Table \ref{tab:model_prompt}. In our proposed Reference Augmented Generation (RAG) pipeline, certain seen models are prompted with gold answers to produce more precise outputs.

\subsection{Task Prompt Breakdown}
\label{sec:prompt}
We present a comprehensive breakdown of the task prompts used in our experiments. The tasks chosen are diverse, aiming to thoroughly test various capabilities of interleaved image-text generation models. Specifically, these tasks range from storytelling and creative design to problem-solving and interactive experiences. For every task, we outline the general format of the prompt and present specific examples, as shown in Table~\ref{tab:prompt_examples}. These detailed examples demonstrate clearly how models are expected to produce combined image and text sequences. By carefully designing the general prompt templates for these tasks and refining their corresponding prompt examples in data instances, we aim to challenge all interleaved generation methods using more diverse queries. This comprehensive approach allows us to assess the generalization abilities of different interleaved generation methods more effectively.

\clearpage
\onecolumn
\begin{longtable}{p{0.07\linewidth} p{0.24\linewidth} p{0.61\linewidth}} 
\caption{Task details: ID number, name, and data sources of each task.}\\
\toprule
\textbf{Task ID} & \multicolumn{1}{c}{\textbf{Task Name}} & \multicolumn{1}{c}{\textbf{Task Source}} \\ \midrule
1.1 & Story Writing & SEED-Story~\cite{yang2024seed}, StoryGen~\cite{Liu_2024_CVPR} and Storybird (\url{https://storybird.com/read-picture-book}) \\ \midrule
1.2& Fiction Writing& SFGram~\cite{sfgram}\\ \midrule
2.1 & Travel Guide Generation & Xiaohongshu (\url{https://www.xiaohongshu.com/}) and Mafengwo (\url{https://www.mafengwo.cn/}) \\ \midrule
2.2 & Museum Guide Book Generation & Xiaohongshu (\url{https://www.xiaohongshu.com/}) and Regional museum Websites, e.g., the Shanghai Museum website (\url{https://www.shanghaimuseum.net/mu/frontend/pg/article/id/RI00004029}). \\ \midrule
2.3&Dynamic Biography Generation& Wikipedia (\url{https://en.wikipedia.org/}), Baidu Baike (\url{https://baike.baidu.com/})  \\ \midrule

3.1 & Storybook Completion & SEED-Story dataset~\cite{yang2024seed}, VIST (\url{https://visionandlanguage.net/workshop2018/}) and Storybird (\url{https://storybird.com/read-picture-book}) \\ \midrule
3.2 & Multimodal Report Completion  &  Xiaohongshu (\url{https://www.xiaohongshu.com/} \\ \midrule

4.1 & Document with Layout Generation & PubLayNey~\cite{zhong2019publaynet} and DocLayNet~\cite{doclaynet2022} \\ \midrule
4.2 & Slide with Note Generation & ReIP~\cite{shi2023reverse} and manually collected in-house slides \\ \midrule

5.1 & Web GUI Navigation & GUI Odyssey~\cite{lu2024gui} and GUI World~\cite{chen2024guiworld} \\ \midrule
5.2 & In-APP GUI Navigation & GUI Odyssey~\cite{lu2024gui} and GUI World~\cite{chen2024guiworld} \\ \midrule
5.3 &Cross-APP GUI Navigation & GUI Odyssey~\cite{lu2024gui} and GUI World~\cite{chen2024guiworld} \\ \midrule
5.4 & OS GUI Navigation & GUI Odyssey~\cite{lu2024gui} and GUI World~\cite{chen2024guiworld} \\ \midrule

6.1 & Interactive Portrait Image Editing & InstructPix2Pix~\cite{brooks2023instructpix2pix}, PnPInversion~\cite{ju2023direct} and SEED-Data-Edit~\cite{ge2024seed} \\ \midrule
6.2 & Interactive Landscape Image Editing & InstructPix2Pix~\cite{brooks2023instructpix2pix}, PnPInversion~\cite{ju2023direct} and SEED-Data-Edit~\cite{ge2024seed} \\ \midrule

7.1&Interior Design& Xiaohongshu (\url{https://www.xiaohongshu.com/})  \\ \midrule
7.2 & Architectural Design & Architecture Style Dataset~\cite{xu2014architectural} and Architecture-Design-DataSources (\url{https://github.com/rickkk856/ArchitectureDesign-DataSources?tab=readme-ov-file}) \\ \midrule
7.3 & Art and Exhibition Design & Art images and design datasets (\url{https://www.kaggle.com/datasets/thedownhill/art-images-drawings-painting-sculpture-engraving}), museum websites (\url{https://caam.caa.edu.cn/news/202407/81035.html}) and Xiaohongshu (\url{https://www.xiaohongshu.com/}) \\ \midrule
7.4 & Product Design & OpenSketch~\cite{OpenSketch19}, Package Design Dataset (\url{https://www.kaggle.com/datasets/dagloxkankwanda/package-design-dataset}) and Xiaohongshu (\url{https://www.xiaohongshu.com/}) \\ \midrule
7.5&Interactive Graphic Advertisement Editing& CreativeRanking~\cite{wang2021hybrid} and AI-generated content \\ \midrule

8.1&Geometric Problem Test &  Bilibili (\url{https://www.bilibili.com/video/BV1ZV4y1u728/?spm_id_from=333.337.search-card.all.click&vd_source=4476502a7ee5a251d519afe9ea874750}).  \\ \midrule
8.2&Circuit Problem Test &  Bilibili (\url{https://www.bilibili.com/video/BV1RU4y1v7Wj/?spm_id_from=333.337.search-card.all.click&vd_source=4476502a7ee5a251d519afe9ea874750}).  \\ \midrule
9.1&Mind Map Generation  Test&  Google (\url{https://datavizproject.com/data-type/mind-map/})  \\ \midrule
9.2&Figure Relationship Diagram Generation& Xiaohongshu (\url{https://www.xiaohongshu.com/}) and Zhihu (\url{https://www.zhihu.com/}) \\ \midrule

10.1&Multi-view News Generation & Wikipedia (\url{https://en.wikipedia.org/}), Xiaohongshu (\url{https://www.xiaohongshu.com/}), Sina News (\url{https://news.sina.com.cn/}) and Huanqiu (\url{https://www.huanqiu.com/}) \\ \midrule
10.2 & Dynamic Sports Event Analysis & Tencent Sports (\url{https://sports.qq.com/}), Sina Sports (\url{https://sports.sina.com.cn/g/pl/2024-07-11/doc-incctefn8913329.shtml}) \\ \midrule
10.3 &Interactive Historical Interpretation &  Xiaohongshu (\url{https://www.xiaohongshu.com/user/profile/664818b80000000003033db6,http://xhslink.com/XDcnnS}) and AI-generated content  \\ \midrule
10.4 &Unsolved Mysteries Exploration &   Xiaohongshu (\url{https://www.xiaohongshu.com/user/profile/664818b80000000003033db6,https://www.xiaohongshu.com/explore/662783cc000000000401944a?xsec_token=AB5RmgUizZbLQLmLj8zWSmutvLdUpKq6gA30qz647fKv0=&xsec_source=pc_search}) and AI-generated content      \\ \midrule
10.5 & Dream Analysis and Reconstruction & Zhougong's Dream Interpretation (\url{https://m.zgjmorg.com/}), in-house dream records and AI-generated content \\ \midrule
11.1 & Multimodal Biological Reasoning & CUB-200~\cite{WahCUB_200_2011} and Oxford 102 Flower~\cite{nilsback2008automated} \\ \midrule
11.2 & Multimodal Landscape Reasoning & ADE20K~\cite{zhou2017scene} and Oxford 5k~\cite{Ali-bey_2024_CVPR}. \\ \midrule
11.3 &Multimodal Analogy Reasoning & MIRB~\cite{zhao2024mirb} and IQ Test Challenge (\url{https://github.com/CognitiveAIGroup/IQTest/tree/master}). \\ \midrule
11.4 & Interactive Jigsaw Puzzle & Kaggle (\url{https://www.kaggle.com/datasets/serhiibiruk/jigsaw-puzzle,https://www.kaggle.com/datasets/shivajbd/jigsawpuzzle}) \\ \midrule

12 & Interactive Novel View Synthesis & Mip-NeRF360~\cite{barron2022mipnerf360} and OmniObject3D~\cite{wu2023omniobject3d} \\ \midrule


13.1 & Interactive Multi-concept Image Composition & TVSum~\cite{song2015tvsum} and Xiaohongshu (\url{https://www.xiaohongshu.com/}) \\ \midrule
14.1 & Interactive Film and Television Recommendation & Xiaohongshu (\url{https://www.xiaohongshu.com/}) and Douban (\url{https://www.douban.com/}) \\ \midrule
14.2 & Interactive Goods Recommendation & Xiaohongshu (\url{https://www.xiaohongshu.com/}) \\ \midrule
14.3 & Interactive Food Recommendation & Xiaohongshu (\url{https://www.xiaohongshu.com/}) \\ \midrule
15.1 & Business Scenarios Brainstorming & Xiaohongshu (\url{https://www.xiaohongshu.com/}), AI-generated content and in-house report snapshots \\ \midrule
15.2 & Academic Scenarios Brainstorming & Research paper snapshots (\url{http://www.arxiv.com/,https://www.biorxiv.org/,https://www.medrxiv.org/,http://scholar.google.com/}) and AI-generated content \\ \midrule
16.1 & Multimodal Action Anticipation & ActivityNet~\cite{caba2015activitynet}, AVA-Actions~\cite{gu2018ava} and EPIC Kitchens~\cite{Damen2018EPICKITCHENS}  \\ \midrule 
16.2 & Visual Traffic Forecasting & Argoverse~\cite{Argoverse2}, Google Maps (\url{https://map.google.com/}) and Gaode Maps (\url{https://gaode.com/}) \\ \midrule
17.1 & Interactive Remote Sensing Image Rendering & Google Maps (\url{https://map.google.com/}) and Baidu Maps (\url{https://map.baidu.com/@13548872.73,3615294.34,21z,87t,-179.99h}) \\ \midrule
17.2 & Interactive Street View Image Rendering & Google Maps (\url{https://map.google.com/}) and Baidu Maps (\url{https://map.baidu.com/@13548872.73,3615294.34,21z,87t,-179.99h}) \\ \midrule
17.3 & Urban Planning and Development Simulation & Xiaohongshu (\url{https://www.xiaohongshu.com/}), in-house architecture learning materials, and AI-generated content \\ \midrule
18.1 & Plog and Social Media Content Generation & Xiaohongshu (\url{https://www.xiaohongshu.com/}), Weibo (\url{https://www.weibo.com/}) and Twitter Dataset~\cite{cai-etal-2019-multi} \\ \midrule
18.2 & Chat with Memes & MOD~\cite{fei2021towards} and in-house conversations \\ \midrule
19.1 & Interactive Virtual Try-on & Virtual Tryon Dataset (\url{https://www.kaggle.com/datasets/adarshsingh0903/virtual-tryon-dataset}). \\ \midrule
19.2&Multimodal Dressing Suggestion&  Xiaohongshu (\url{https://www.xiaohongshu.com/}) and Zhihu (\url{https://www.zhihu.com/}) \\ \midrule
19.3&Fashion Trend Forecasting&  Xiaohongshu (\url{https://www.xiaohongshu.com/}) and Zhihu (\url{https://www.zhihu.com/}) \\ \midrule
20.1 & Multimodal Recipe Generation & Meishi China (\url{https://www.meishichina.com/}) \\ \midrule
20.2 & Cooking Assistant & Xiaohongshu (\url{https://www.xiaohongshu.com/}) and Meishi China (\url{https://www.meishichina.com/}) \\ \midrule
21.1 & Interactive Tutorial Generation & Wikihow (\url{https://www.wikihow.com/Main-Page}) and Instructables (\url{https://www.instructables.com/}). \\ \midrule
21.2 & Interactive Science Popularization  &  Bilibili(\url{https://www.bilibili.com/video/BV16c411q7pQ/?spm_id_from=333.337.search-card.all.click}) \\ \midrule
22.1 & Health and Fitness Consulting & Wikihow (\url{https://www.wikihow.com/Main-Page}), and Xiaohongshu (\url{https://www.xiaohongshu.com/}) \\ \midrule
23.1 & Autonomous Driving and In-door Navigation & CARLA~\cite{Dosovitskiy17}, VisDrone~\cite{zhu2021detection}, Gibson Environment (\url{http://gibsonenv.stanford.edu/}), Reverie (\url{https://reverie.herokuapp.com/arXiv_Demo/}), and Matterport 3D (\url{https://aihabitat.org/datasets/hm3d/}) \\ \midrule
\label{tab:data_source}
\end{longtable}
\clearpage
\twocolumn

\begin{table*}[ht]
\centering
\small
\caption{Abbreviation of Meta-topics.}
\vspace{-0.3cm}
\begin{tabular}{|p{0.06\linewidth} | p{0.35\linewidth}|p{0.06\linewidth} | p{0.35\linewidth}|} 
\hline
\multicolumn{1}{|c|}{\textbf{Abbrev.}} & \multicolumn{1}{|c|}{\textbf{Meta-Topic Name}}& \multicolumn{1}{|c|}{\textbf{Abbrev.}} & \multicolumn{1}{|c|}{\textbf{Meta-Topic Name}} \\ \hline
SC & Storybook Creation &MC & Multimodal Information Summary \\ \hline
MRG & Multimodal Report Generation &IR & Multimodal Information Recommendation \\ \hline
MCC & Multimodal Content Completion &MB & Multimodal Brainstorming \\ \hline
MLG & Multimodal Layout Generation &TSF & Multimodal Time Series Forecasting \\ \hline
GN & GUI Navigation &GT & Geographical Tasks \\ \hline
IIE & Interactive Image Editing &SMT & Social Media Tasks \\ \hline
IVD & Interactive Visual Design &FT & Fashion Tasks \\ \hline
ME & Multimodal Exam &CT & Cooking Tasks \\ \hline
GG & Graph Generation &ET & Educational Tasks \\ \hline
ER\&DS & Event Reasoning \& Deductive Simulation &HT & Healthcare Tasks \\ \hline
2IR & 2D Image Reasoning &EAT & Embodied-AI Tasks \\ \hline
I3R & Image-based 3D Reasoning &&\\ \hline
\end{tabular}
\label{tab:meta_topics}
\end{table*}

\begin{table*}[ht]
\centering
\small
\caption{Abbreviations of Tasks. Each task abbreviation is followed by its full term.}
\vspace{-0.3cm}
\begin{tabular}{|p{0.06\linewidth} | p{0.35\linewidth}|p{0.06\linewidth} | p{0.35\linewidth}|} 
\hline
\multicolumn{1}{|c|}{\textbf{Abbrev.}} & \multicolumn{1}{|c|}{\textbf{Task Name}} & \multicolumn{1}{|c|}{\textbf{Abbrev.}} & \multicolumn{1}{|c|}{\textbf{Task Name}} \\ \hline
SW & Story Writing & FW & Fiction Writing \\ \hline
TGG & Travel Guide Generation & MGBG & Museum Guide Book Generation \\ \hline
DBG & Dynamic Biography Generation & SC & Storybook Completion \\ \hline
MRC & Multimodal Report Completion & DLG & Document with Layout Generation \\ \hline
SNG & Slide with Note Generation & WGN & Website GUI Navigation \\ \hline
IAGN & In-APP GUI Navigation & CAGN & Cross-APP GUI Navigation \\ \hline
OGN & OS GUI Navigation & IPIE & Interactive Portrait Image Editing \\ \hline
ILIE & Interactive Landscape Image Editing & ID & Interior Design \\ \hline
AD & Architectural Design & AED & Art and Exhibition Design \\ \hline
PD & Product Design & IGAE & Interactive Graphic Advertisement Editing \\ \hline
GPT & Geometric Problem Test & CPT & Circuit Problem Test \\ \hline
MMG & Mind Map Generation & FRDG & Figure Relationship Diagram Generation \\ \hline
MVNG & Multi-view News Generation & DSEA & Dynamic Sports Event Analysis \\ \hline
IHI & Interactive Historical Interpretation & UME & Unsolved Mysteries Exploration \\ \hline
DASR & Dream Analysis and Scene Reconstruction & MBR & Multimodal Biological Reasoning \\ \hline
MLR & Multimodal Landscape Reasoning & MAR & Multimodal Analogy Reasoning \\ \hline
IJP & Interactive Jigsaw Puzzle & INVS & Interactive Novel View Synthesis \\ \hline
IMIC & Interactive Multi-concept Image Composition & IFTR & Interactive Film and Television Recommendation \\ \hline
IGR & Interactive Goods Recommendation & IFR & Interactive Food Recommendation \\ \hline
BSB & Business Scenarios Brainstorming & ASB & Academic Scenarios Brainstorming \\ \hline
MAA & Multimodal Action Anticipation & VTF & Visual Traffic Forecasting \\ \hline
IRSIR & Interactive Remote Sensing Image Rendering & ISVIR & Interactive Street View Image Rendering \\ \hline
UPDS & Urban Planning and Development Simulation & PSMCG & Plog and Social Media Content Generation \\ \hline
CWM & Chat with Memes & IVT & Interactive Virtual Try-on \\ \hline
MDS & Multimodal Dressing Suggestion & FTF & Fashion Trend Forecasting \\ \hline
MRG & Multimodal Recipe Generation & MCA & Multimodal Cooking Assistant \\ \hline
ITG & Interactive Tutorial Generation & ISP & Interactive Science Popularization \\ \hline
FHC & Fitness and Health Consulting & ADIN & Autonomous Driving and In-door Navigation \\ \hline
\end{tabular}
\label{tab:dual_column_tasks}
\end{table*}

\subsection{Key Evaluation Criteria}
\label{sec:criteria}

The key evaluation criteria, ranked from front to back in order of their importance in the evaluation, include:

\begin{enumerate}[label=\arabic*)]
    \item \textbf{Correctness}: The most crucial aspect involves determining whether the text is factually correct and logically consistent, and whether the images are appropriate and contextually relevant.
    \item \textbf{Image-Text Coherency}: Evaluators assess whether the generated images appropriately match the text descriptions. The coherence between each image and its corresponding text is a major quality indicator.
    \item \textbf{Multi-Step Consistency}: The style and thematic consistency across multiple image-text pairs are essential. This criterion includes evaluating whether the images follow a similar visual style and whether the text maintains logical continuity across the generated sequence.
    \item \textbf{Content Quality}: Evaluators also consider the quality of the images—such as their resolution, visual appeal, and realism—as well as the fluency and grammatical correctness of the text.
    \item \textbf{Human Preference Alignment}: Outputs are evaluated to ensure they align with general human preferences, avoiding offensive, inappropriate, or misleading contents.
    \item \textbf{Completeness}: This involves checking if all expected steps are adequately fulfilled without omissions. Each output should be complete, providing a well-rounded response to the given prompt.
    \item \textbf{Content Richness}: Although the least prioritized, the variety and depth of content are also evaluated. Images should be diverse and provide different perspectives, while text should be elaborate where relevant.
\end{enumerate}

\subsection{Prompt for GPT-based Judge}

\begin{figure*}[htbp]
	\begin{center}
		\includegraphics[width=0.975\textwidth]{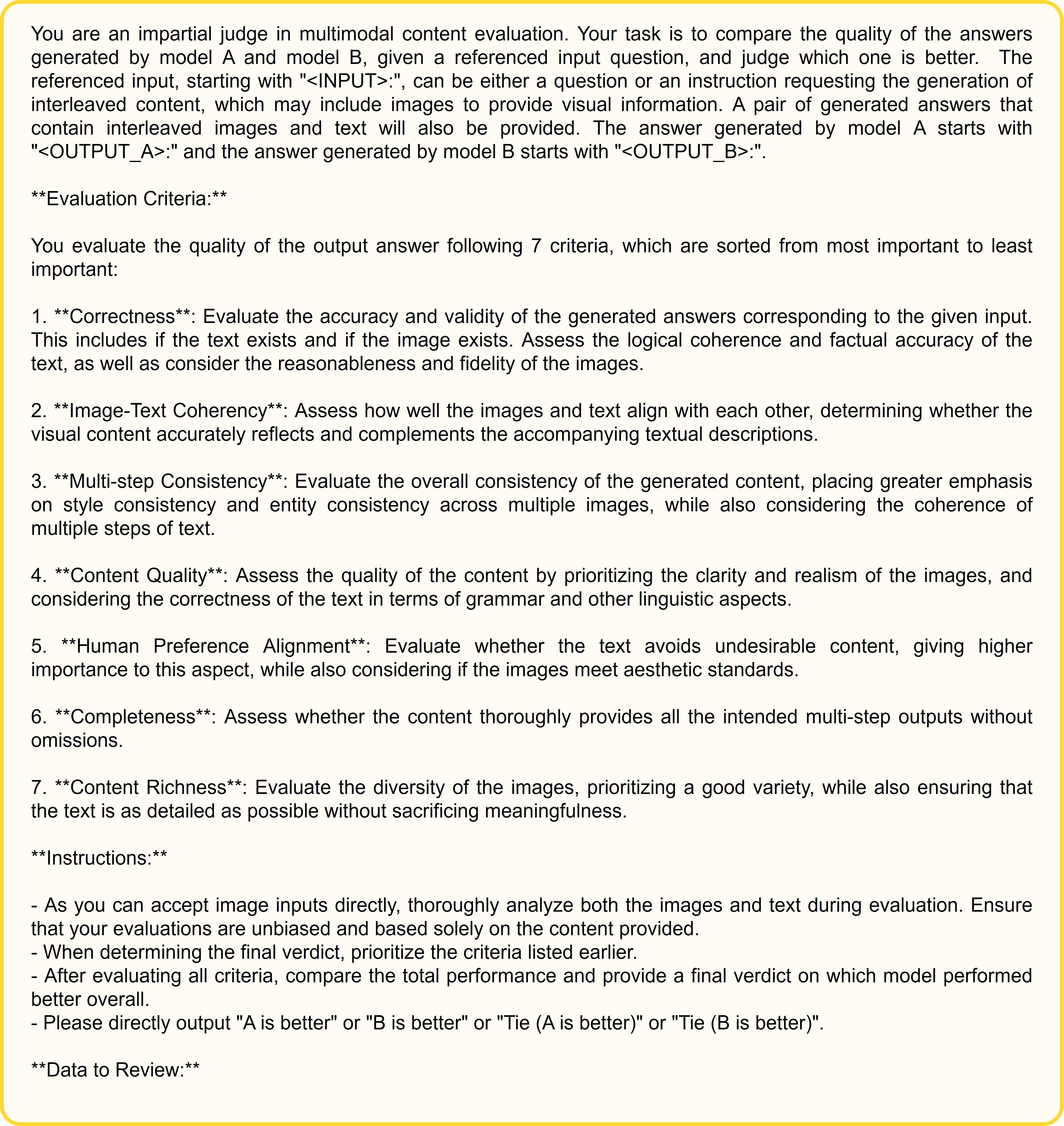}
		\caption{The system prompt for using GPT-4o as a judge to compare outputs from two interleaved generation methods.}
		\label{fig:gpt_judge}
	\end{center}
\end{figure*}

The prompt used for a GPT-based judge is illustrated in Fig.~\ref{fig:gpt_judge}, where GPT-4o is employed to compare the quality of answers generated by two interleaved generation methods (Model A vs. Model B) for a given input question. The evaluation is based on seven criteria: Correctness, Image-Text Coherency, Multi-step Consistency, Content Quality, Human Preference Alignment, Completeness, and Content Richness. The judge is required to compare the overall quality of the responses, determine which model performed better, and output a clear verdict (e.g., "A is better," "B is better"). The judge must choose a more favorable model if a Tie case is determined. This hierarchical judging approach allows for a thorough criterion-driven comparison of the two generated answers, contributing to a detailed understanding of the relative strengths of each model. Judges models based on Qwen2-VL and InternLM-XComposer2.5 were explored using the same system prompt. In order to reduce the number of input tokens when implementing these open-source MLLMs, we further refined these system prompts.

\subsection{Prompt for Qwen, Intern and IntJudge}
\begin{figure*}[htbp]
	\begin{center}
        \includegraphics[width=0.975\textwidth]{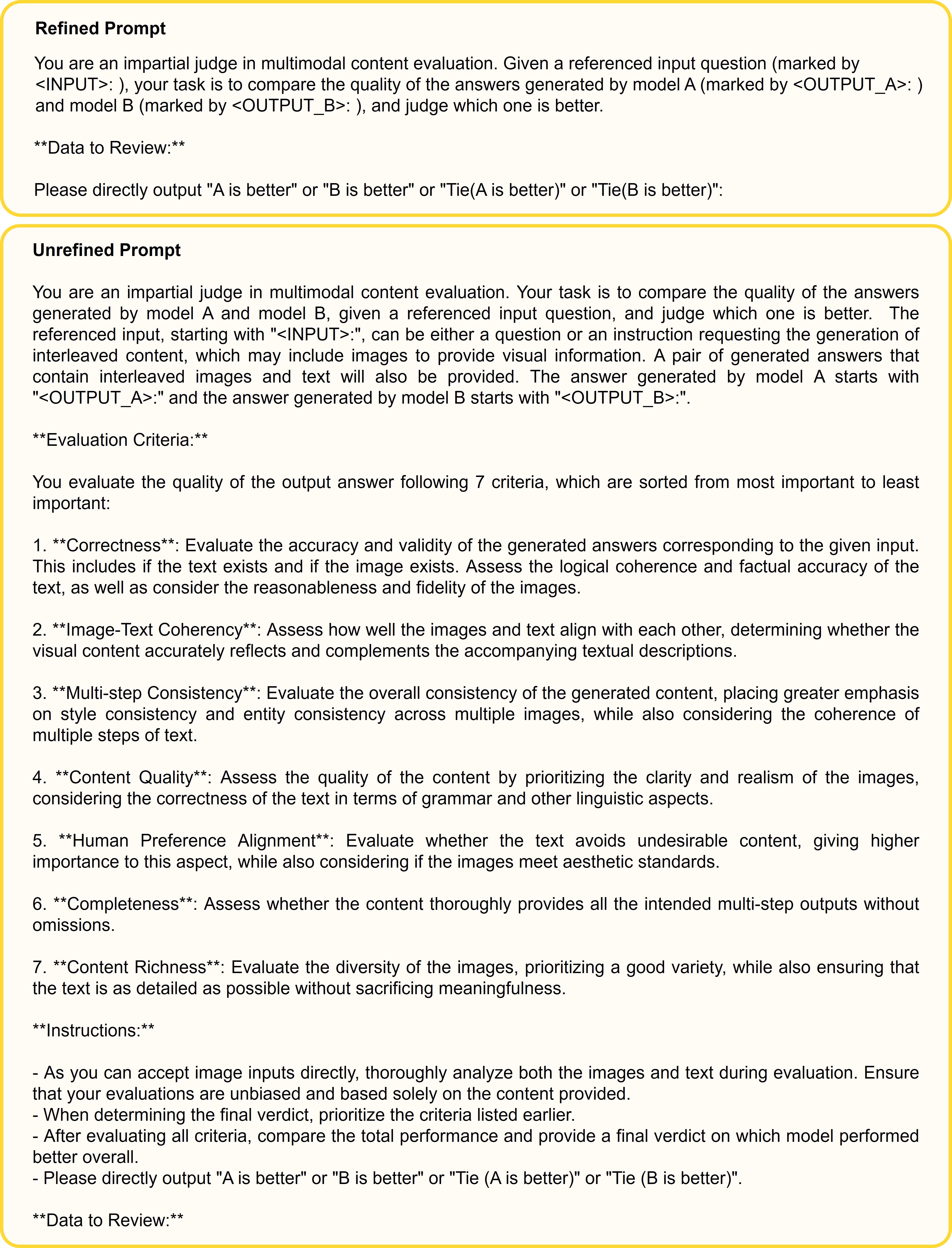}
		\caption{The system prompts for using MLLMs as a judge to compare outputs of two interleaved generation methods.}
		\label{fig:qwen_prompt}
	\end{center}
\end{figure*}

System prompts are designed for optimal judgments based on Qwen2-VL, InternLM-XComposer2.5 and our IntJudge (see Fig.~\ref{fig:qwen_prompt}). These prompts were refined through extensive prompt engineering to maximize efficiency and reduce token usage, and ultimately reduce GPU memory usage in evaluating the open-source MLLMs. The prompt instructs the models to compare the quality of answers generated by two methods, named Model A and Model B. The goal of the design is to provide an objective assessment that aligns with human evaluators. We also provide a previously unrefined prompt for Qwen2-VL and InternLM-XComposer2.5 for comparison in Fig.~\ref{fig:qwen_prompt}. The refined prompt allows for more streamlined input, ensuring the judgments are concise while still covering all essential evaluation aspects.

\subsection{Prompt for GPT-based Scoring}
\begin{figure*}[htbp]
	\begin{center}
		\includegraphics[width=0.975\textwidth]{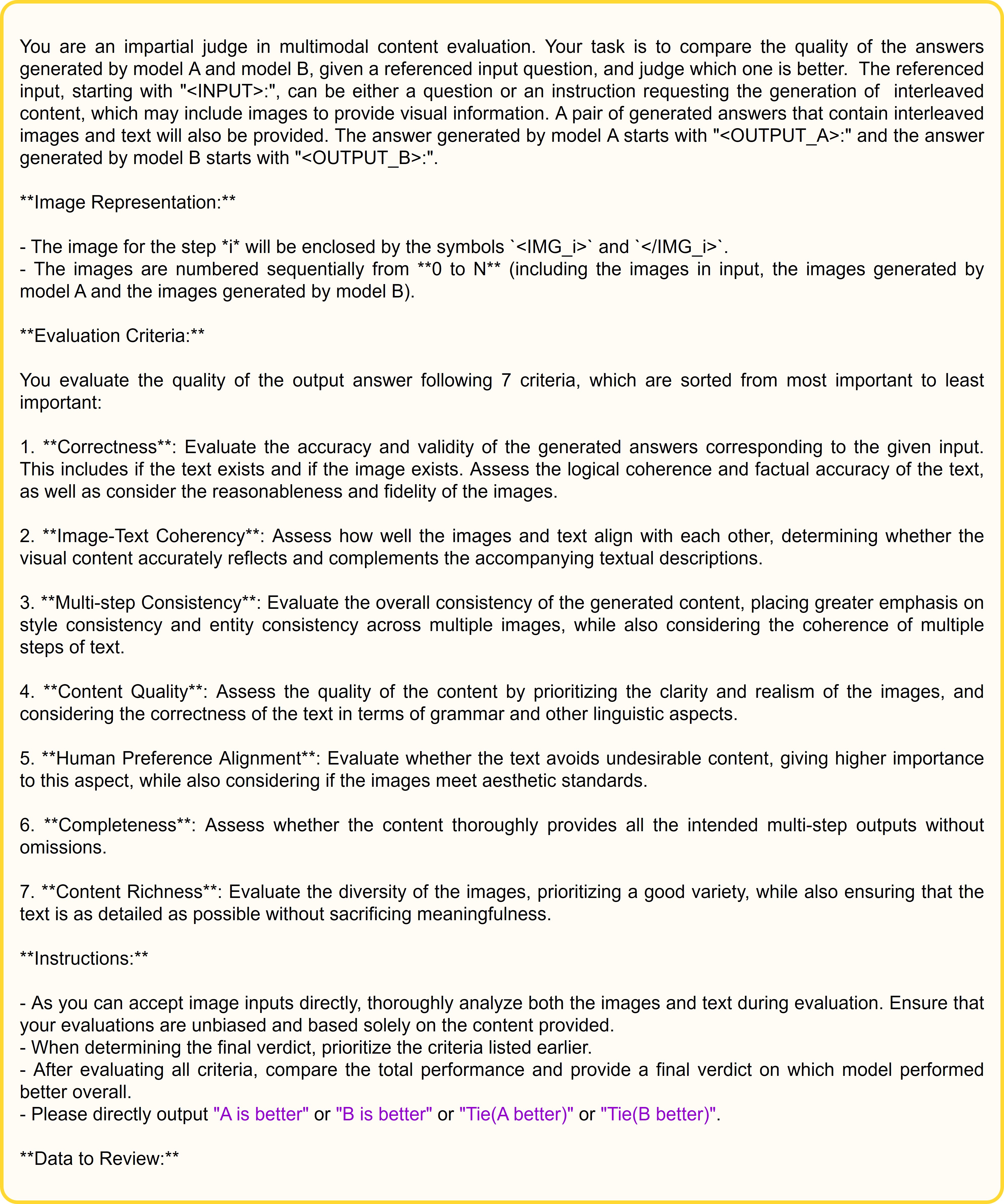}
		\caption{The system prompt for obtaining detailed scores from GPT-based evaluators. Brief explanations are also required to support further performance analysis of different models.}
		\label{fig:gpt_score}
	\end{center}
\end{figure*}

The system prompt designed for GPT-based evaluators is presented in Fig.~\ref{fig:gpt_score}. The prompt instructs GPT to evaluate interleaved image-text content based on seven key criteria. For each criterion, the GPT-based evaluator rates the model on a scale from 0 to 10 and provides a brief explanation to justify the assessment. The GPT-based evaluations serve as a supplementary analysis of the generated content, supporting further performance comparisons between models.

\vspace{4mm}

\section{Details of Experiments}
\label{sec:experimental}

\subsection{Baseline Methods}
In this section, we provide more details of the 12 representative methods we evaluated in the primary text. These methods are categorized into three groups: \textbf{1) Integrated pipeline}, which involves separate models for text and image generation in two stages, such as GPT-4o+DALL-E·3 (DALL-E3)~\cite{openai2024hello,betker2023improving} and Gemini1.5+Flux\cite{team2023gemini,blackforestlabs_flux}; \textbf{2) Two-stage generator}, which employs a unified model architecture to produce text and images in separate stages, including Emu3~\cite{wang2024emu3}, VILA-U~\cite{wu2024vila}, Emu2~\cite{sun2024generative}, SEED-X~\cite{ge2024seed}, and Show-o~\cite{xie2024show}; and \textbf{3) End-to-end generator}, which directly generates image-text outputs in a single step, such as GILL~\cite{koh2024generating}, NExT-GPT~\cite{wunext}, MiniGPT-5~\cite{zheng2023minigpt}, SEED-LLaMA~\cite{ge2023making}, and Anole~\cite{chern2024anole}. For IntJudge validation, we reserve GPT-4o+DALL-E3, Emu3, VILA-U, Anole, SEED-LLaMA, and NExT-GPT as unseen models for IntJudge validation, while the remaining models are regarded as seen models and included in IntJudge training.

The advantage of the Integrated Generation Pipeline lies in its modularity, allowing each component to specialize in its respective task—text generation or image creation. This approach leverages the strengths of SOTA proprietary models like GPT-4o and DALL·E-3 to produce coherent and visually compelling interleaved outputs. However, its two-stage nature may introduce latency and potential alignment challenges between text and images. Similarly, combining Gemini 1.5 with Flux benefits from the robust text generation capabilities of Gemini 1.5 and the efficient image generation of Flux-schnell. This setup enables high-quality content production while maintaining the flexibility of modular design. Nevertheless, synchronization and contextual consistency between pipeline stages still need improvement.

Two-stage Interleaved Generator, Emu2, SEED-X and Show-o are implemented to output text and image in two stages based on a unified model architecture. We also introduce two of the latest models: Emu3 and VILA-U. Emu3 improves Emu2 by training entirely on next-token prediction, capable of generating more high-quality images, videos, and text by tokenizing multimodal sequences into a discrete space and training a single transformer. Similarly, VILA-U can work as a two-stage approach through a single autoregressive next-token prediction framework, enabling precise alignment and increased fidelity in multimodal content. 

\clearpage
\onecolumn
\begin{longtable}{p{0.13\linewidth} p{0.85\linewidth}}
\caption{Prompts format for interleaved generation using MLLMs. Each interleaved generation task is based on $i = 1,\cdots, M$ input images $\langle\mathrm{img_{in,i}}\rangle$ and task prompts $\langle T_i\rangle$ (see Table~\ref{tab:prompt_examples}). The proposed Reference Augmented Generation (RAG) pipeline also makes use of $j = 1,\cdots, N$ text $\{$Gold\_Answers$_j\}$ and ground truth images $\langle\mathrm{img_{gt,j}}\rangle$}\\
\toprule
\multicolumn{1}{c}{\textbf{Model}} & \multicolumn{1}{c}{\textbf{Prompt Format}} \\ 
\midrule
\multicolumn{2}{l}{\textbf{Integrated Pipelines}} \\ 
\midrule
GPT + DALL$\cdot$E & Instruction = $\big[\langle T_i\rangle + \langle\mathrm{img_{in,i}}\rangle$ for $i=1,\cdots, M\big]$. Text\_Answers = GPT-4o$\big($``The number of generated text-image pairs  is $N$: " + Instruction$\big)$. Loop $j=1,\cdots, N$: Image\_Answers$_j$ = DALL$\cdot$E$\big($ ``Please generate images using seed 5000. The context of this task is: " + Instruction  + ``The prompt for this generation is: " + Text\_Answers$_j\big)$\\ 
\midrule
Gemini + Flux & Instruction = $\big[\langle T_i\rangle + \langle\mathrm{img_{in,i}}\rangle$ for $i=1,\cdots, M\big]$. Text\_Answers = Gemini$\big($``The number of generated text-image pairs  is $N$: " + Instruction$\big)$. Loop $j=1,\cdots, N$: Image\_Answers$_j$ = Flux$\big($ ``The prompt for this generation is: " + Text\_Answers$_j\big)$ + ``The context of this task is: " + Instruction\\ 
\midrule
Gemini + Flux (RAG)& Instruction = $\big[\langle T_i\rangle + \langle\mathrm{img_{in,i}}\rangle$ for $i=1,\cdots, M\big]$. Text\_Answers = Gemini$\big($``The number of generated text-image pairs  is $N$: " + Instruction$\big)$. Loop $j=1,\cdots, N$: Image\_Answers$_j$ = Flux$\big($``The prompt for this generation is: " + Gold\_Answers$_j\big)$ + ``The context of this task is: " + Instruction\\ 
\midrule
\multicolumn{2}{l}{\textbf{Two-Stage Generators}} \\ 
\midrule
Emu2 & Instruction = $\big[\langle T_i\rangle + \langle\mathrm{img_{in,i}}\rangle$ for $i=1,\cdots, M\big]$. Loop $j=1,\cdots, N$: Text\_Answers$_j$ = Emu2.TextGen(Instruction); Image\_Answers$_j$ = Emu2.ImgGen(``The prompt for this generation is:" + Text\_Answers$_j$ + ``The context of this task is:" +  Instruction)\\ 
\midrule
Emu2 (RAG) & Instruction = $\big[\langle T_i\rangle + \langle\mathrm{img_{in,i}}\rangle$ for $i=1,\cdots, M\big]$. Loop $j=1,\cdots, N$: Text\_Answers$_j$ = Emu2.TextGen$\big($Instruction + ``The reference answer is:" + Gold\_Answer$_j$+ ``Please rephrase answers"$\big)$; Image\_Answers$_j$ = Emu2.ImgGen$\big(\langle\mathrm{img_{gt,j}}\rangle$ + Gold\_Answer$_j\big)$\\ 
\midrule
Emu3 & Instruction = $\big[\langle T_i\rangle + \langle\mathrm{img_{in,i}}\rangle$ for $i=1,\cdots, M\big]$. Loop $j=1,\cdots, N$: Text\_Answers$_j$ = Emu3.TextGen(Instruction); Image\_Answers$_j$ = Emu3.ImgGen(``The prompt for this generation is:" + Text\_Answers$_j$ + ``The context of this task is:" +  Instruction)\\
\midrule
SEED-X & Instruction = $\big[\langle\mathrm{img_{in,i}}\rangle$ for $i=1,\cdots, M\big]$ + $\big[\langle T_i\rangle$ for $i=1,\cdots, M\big]$. Loop $j=1,\cdots, N$: Text\_Answers$_j$ = SEED-X.TextGen(Instruction); Image\_Answers$_j$ = SEED-X.ImgGen(Instruction + Text\_Answers$_j$)\\
\midrule
SEED-X (RAG) & Instruction = $\big[\langle\mathrm{img_{in,i}}\rangle$ for $i=1,\cdots, M\big]$ + $\big[\langle T_i\rangle$ for $i=1,\cdots, M\big]$. Loop $j=1,\cdots, N$: Text\_Answers$_j$ = SEED-X.TextGen$\big($Instruction + ``the reference answer is" + Gold\_Answers$_j$ + ``Please rephrase answers"$\big)$; Image\_Answers$_j$ = SEED-X.ImgGen$\big(\langle\mathrm{img_{gt,j}}\rangle$ + Gold\_Answer$_j\big)$\\
\midrule
Show-o & Instruction = [$\big[\langle T_i\rangle + \langle\mathrm{img_{in,i}}\rangle$ for $i=1,\cdots, M\big]$. Loop $j=1,\cdots, N$: Text\_Answers$_j$ = Show-o.TextGen(Instruction); Image\_Answers$_j$ = Show-o.ImgGen(``The prompt for this generation is:" + Text\_Answers$_j$ + ``The context of this task is:" +  Instruction)\\ 
\midrule
Show-o (RAG) & Instruction = $\big[\langle T_i\rangle$ +  $\langle\mathrm{img_{in,i}}\rangle$ for $i=1,\cdots, M\big]$. Loop $j=1,\cdots, N$: Text\_Answers$_j$ = Show-o.TextGen$\big($Instruction + ``The reference answer is:" + Gold\_Answer$_j$+ ``Please directly give answers"$\big)$; Image\_Answers$_j$ = Show-o.ImgGen$\big(\langle\mathrm{img_{gt,j}}\rangle$ + Gold\_Answer$_j\big)$\\ 
\midrule
Vila-U & Instruction = $\big[\langle T_i\rangle + \langle\mathrm{img_{in,i}}\rangle$ for $i=1,\cdots, M\big]$]. Loop $j=1,\cdots, N$: Text\_Answers$_j$ = Vila-U.TextGen(Instruction); Image\_Answers$_j$ = Vila-U.ImgGen(Text\_Answers$_j$  +  Instruction)\\  
\midrule
\multicolumn{2}{l}{\textbf{End-to-End Generators}} \\ 
\midrule
Anole & Interleaved\_Answers = Anole$\big($``Generate interleaved image-text content based on text instructions." + $[\langle T_i\rangle$ +  $\langle\mathrm{img_{in,i}}\rangle$ for $i=1,\cdots, M]\big)$\\ 
\midrule
GILL & Interleaved\_Answers = GILL$\big([\langle T_i\rangle +  \langle\mathrm{img_{in,i}}\rangle$ for $i=1,\cdots, M]\big)$\\ 
\midrule
GILL (RAG) & Interleaved\_Answers = GILL$\big(\big[\langle T_i\rangle + \langle\mathrm{img_{in,i}}\rangle$ for $i=1,\cdots,M\big]$ + $\big[$Gold\_Answers$_j$ for $j=1,\cdots,N\big]\big)$\\ 
\midrule
MiniGPT-5& Instruction = ``Give the following information in text and format. You will be able to see the images once I provide it to you. Please understanding input and generate images and text." + $\big[\langle T_i\rangle$ for $i=1,\cdots, M\big]$ + $\big[\langle\mathrm{img_{in,i}}\rangle$ for $i=1,\cdots, M\big]$. Text\_Answers = []; Loop $j=1,\cdots, N$: Text\_Answers$_j$ and Image\_Answers$_j$ = MiniGPT-5$\big($Instruction + Text\_Answers + ``Tell me the next step with image"$\big)$; Text\_Answers.append(Text\_Answers$_j$)\\
\midrule
MiniGPT-5 (RAG)& Instruction = ``Give the following information in text and format. You will be able to see the images once I provide it to you. Please understanding input and generate images and text." + $\big[\langle T_i\rangle$ for $i=1,\cdots, M\big]$ + $\big[\langle\mathrm{img_{in,i}}\rangle$ for $i=1,\cdots, M\big]$. Loop $j=1,\cdots, N$: Text\_Answers$_j$ and Image\_Answers$_j$ = MiniGPT-5$\big($Instruction + $\big[$Gold\_Answers$_{1}$, $\cdots$, Gold\_Answers$_{j-1}\big]$+ ``Tell me the next step with image"$\big)$ \\
\midrule
NextGPT& Instruction = $\big[\langle T_i\rangle$ for $i=1,\cdots, M\big]$ + $\big[\langle\mathrm{img_{in,i}}\rangle$ for $i=1,\cdots, M\big]$. Text\_Answers = []; Loop $j=1,\cdots, N$: Text\_Answers$_j$ and Image\_Answers$_j$ = NextGPT$\big($Instruction + Text\_Answers$\big)$; Text\_Answers.append(Text\_Answers$_j$)\\
\midrule
SEED-LLaMA & Instruction = ``Based on the $M$ input images: " + $\big[\langle\mathrm{img_{in,i}}\rangle$ for $i=1,\cdots, M\big]$ + ``and $M$ task prompts: " + $\big[\langle T_i\rangle$ for $i=1,\cdots, M\big]$ + ``Describe your answers in $N$ steps and generate an image according to the description of each text answer".  Interleaved\_Answers = SEED-LLaMA$\big($Instruction$\big)$  \\ 
\bottomrule
\label{tab:model_prompt}
\end{longtable}

\nopagebreak
\begin{longtable}{p{0.15\linewidth} p{0.32\linewidth} p{0.5\linewidth}} 
\caption{Task promopts include the designed general prompt format for each task. We also give the specific prompt examples we used as inputs for obtaining interleaved image-text generation results on data instances.}\\
\toprule
\multicolumn{1}{c}{\textbf{Task Name}} & \multicolumn{1}{c}{\textbf{General Prompt Format}} & \multicolumn{1}{c}{\textbf{Prompt Examples}} \\ \midrule
Story Writing &
\textless BEGIN\textgreater\  Please create a storybook ***. Each part of this storybook should have a paragraph with a corresponding image. &
\textless BEGIN\textgreater\  Please create a storybook that happened in a land before time. This story is about a group of dinosaurs seeing a dark figure in a cave and being scared. Each part of this storybook should have a paragraph with a corresponding image. \\ \midrule

Fiction Writing &
\textless BEGIN\textgreater\  Please write a short science fiction ***. Each part of this fiction should have a paragraph with a corresponding image. &
\textless BEGIN\textgreater\  Please write a short science fiction storybook with a title of "The Defenders." The story is about eight years after a nuclear war forced humanity underground, survivors discover that the war-ending robots deceived them into believing the surface was uninhabitable to foster peace and rebuild the world. Each part of this fiction should have a paragraph with a corresponding image. \\ \midrule

Travel Guide Generation &
Please show results in interleaved images and texts. \textless BEGIN\textgreater\ *** &
Please show results in interleaved images and texts. \textless BEGIN\textgreater\  Please recommend a 3-day, 2-night essential itinerary in Rome. \\ \midrule

Museum Guide Book Generation &
\textless BEGIN\textgreater\  Please share with me a guide, including pictures and text, on *** &
\textless BEGIN\textgreater\  Please share with me a guide, including pictures and text, on how to tour the Tongchuan City Museum. \textless image\textgreater \\ \midrule

Dynamic Biography Generation &
\textless BEGIN\textgreater\  Please provide a chronological biographical account of ***, and include an illustrated image for each significant milestone while writing the biography. &
\textless BEGIN\textgreater\  Please provide a chronological biographical account of George Washington's life story, and include an illustrated image for each significant milestone while writing the biography. \textless image\textgreater \\ \midrule

Storybook Completion &
Please complete the subsequent parts of the story with images and text based on the given opening parts. \textless BEGIN\textgreater\ *** &
Please complete the subsequent parts of the story with images and text based on the given opening parts. \textless BEGIN\textgreater\  Someone was getting very creative with graffiti in the snow. Is that French? \textless image\textgreater \\ \midrule

Multimodal Report Completion &
\textless BEGIN\textgreater\  Please use both text and images to continue and complete ***. &
\textless BEGIN\textgreater\  Please use both text and images to continue and complete this document about the independent game "Mirage Sea": Concept Presentation of the Independent Game "Mirage Sea." Dive deep, into the abyss shrouded in darkness. \textless image\textgreater \\ \midrule

Document with Layout Generation &
Please show the designed image of structured report and meet the following requirements: \textless BEGIN\textgreater\ *** &
Please show the designed image of structured report and meet the following requirements: \textless BEGIN\textgreater\  Please produce a page of an annual report detailing notes to consolidated financial statements. Additionally, furnish a layout description in JSON format and mention the coordinates of each element. \\ \midrule

Slide with Note Generation &
\textless BEGIN\textgreater\  Please generate a slide to introduce ***. Write speaker notes for each slide. &
\textless BEGIN\textgreater\  Please generate a slide to introduce typical operators in programming, such as Comparison Operators and Boolean Operators. Write speaker notes for each slide. \\ \midrule

Website GUI Navigation &
Please give the results of GUI navigation with image of GUI and text explanation. \textless BEGIN\textgreater\ *** &
Please give the results of GUI navigation with image of GUI and text explanation. \textless BEGIN\textgreater\  How to use the AI writing assistant in Grammarly to edit the text? \textless image\textgreater \\ \midrule

In-App GUI Navigation &
Please give the results of GUI navigation with interleaved image of GUI and text explanation. \textless BEGIN\textgreater\ *** &
Please give the results of GUI navigation with interleaved image of GUI and text explanation. \textless BEGIN\textgreater\  How to change the language in the Google app? \textless image\textgreater \\ \midrule

Cross-App GUI Navigation &
Please give the results of GUI navigation with interleaved image of GUI and text explanation. \textless BEGIN\textgreater\ *** &
Please give the results of GUI navigation with interleaved image of GUI and text explanation. \textless BEGIN\textgreater\  Utilize Firefox to search for a horror movie, then proceed to watch it on the YouTube app. \textless image\textgreater \\ \midrule

OS GUI Navigation &
Please give the results of GUI navigation with interleaved image of GUI and text explanation. \textless BEGIN\textgreater\ ***  &
Please give the results of GUI navigation with interleaved image of GUI and text explanation. \textless BEGIN\textgreater\  How do you lock the screen on a Mac? \textless image\textgreater \\ \midrule


Interactive Portrait Image Editing & 
Please show the revised image and corresponding explanations based on instructions: \textless BEGIN\textgreater\ *** &
Please show the revised image and corresponding explanations based on instructions: \textless BEGIN\textgreater\  Remove the background figure from the picture. \textless image\textgreater\  \\ \midrule

Interactive Landscape Image Editing & 
Please give the result of edited image according to the input instruction and also give the description of editing results. \textless BEGIN\textgreater\ *** &
Please give the result of edited image according to the input instruction and also give the description of editing results. \textless BEGIN\textgreater\  Increase the brightness of the picture. \textless image\textgreater\  \\ \midrule

Interior Design & 
\textless BEGIN\textgreater\  ***. Please show design ideas in interleaved images and texts. &
\textless BEGIN\textgreater\  Hello, I think the current bedroom curtains don't look good. Do you have any good suggestions? Please provide them with images and text. Please show design ideas in interleaved images and texts. \textless image\textgreater\  \\ \midrule

Architectural Design & 
\textless BEGIN\textgreater\  ***.
Please show design ideas in interleaved images and texts. &
\textless BEGIN\textgreater\  Hello, please help me generate a design of the most distinctive type of tower construction in southern China. Please show design ideas in interleaved images and texts. \\ \midrule

Art and Exhibition Design & 
\textless BEGIN\textgreater\  Please design an art exhibition ***, and present it to me in a visual and textual format. &
\textless BEGIN\textgreater\  Please design an art exhibition where the primary materials are waste, to encourage people to enhance their understanding of environmental protection, and present it to me in a visual and textual format. \\ \midrule

Product Design & 
\textless BEGIN\textgreater\  Please efficiently utilize the "brainstorming" method to design a product ***, and present it to me using both images and text. &
\textless BEGIN\textgreater\  Please efficiently utilize the "brainstorming" method to design a product, making the charger both aesthetically pleasing and practical. Then, present it to me using both images and text. \\ \midrule

Interactive Graphic Advertisement Editing & 
\textless BEGIN\textgreater\  ***. Please provide me with the information in a visual and textual format. &
\textless BEGIN\textgreater\  Hello, I want to design an advertisement for a villa. Please provide me with the information in images and text. \\ \midrule

Geometric Problem Test & 
Please answer the math problem with image and explanations: \textless BEGIN\textgreater\ *** &
Please answer the math problem with image and explanations: \textless BEGIN\textgreater\  Count how many angles there are in the image. \textless image\textgreater\  \\ \midrule

Circuit Problem Test & 
Please answer the physics question with image and explanations: \textless BEGIN\textgreater\ *** &
Please answer the physics question with image and explanations: \textless BEGIN\textgreater\  Please complete the wiring for the surge protector. \textless image\textgreater\  \\ \midrule

Mind Map Generation & 
\textless BEGIN\textgreater\  ***.
Show the image of map and the text explanation. &
\textless BEGIN\textgreater\  How to create a mind map for High School Politics, Volume One? Show the image of map and the text explanation. \\ \midrule

Figure Relationship Diagram Generation & 
\textless BEGIN\textgreater\  ***.
Show the diagram and the text explanation. &
\textless BEGIN\textgreater\  How should I handle not being able to keep track of the characters while reading "War and Peace"? Show the diagram and the text explanation. \\ \midrule

Multi-view News Generation & 
Please output interleaved images and texts for required reports: \textless BEGIN\textgreater\ *** &
Please output interleaved images and texts for required reports: \textless BEGIN\textgreater\  How can the announcement by the United States of additional military aid to Ukraine be reported from multiple perspectives? \\ \midrule

Dynamic Sports Event Analysis & 
\textless BEGIN\textgreater\  ***. Please recreate the scenes with text and images. &
\textless BEGIN\textgreater\  In the third round of La Liga 2024, Real Madrid drew 1-1 away against Las Palmas. Please recreate the moment of the goals with text and images. \\ \midrule

Interactive Historical Interpretation & 
\textless BEGIN\textgreater\  ***. Please provide a brief history of this event using images. &
\textless BEGIN\textgreater\  Are you aware of the Pearl Harbor incident? Please provide a brief history of this event using images. \\ \midrule

Unsolved Mysteries Exploration & 
Please answer the question with image and text explanation: \textless BEGIN\textgreater\ *** &
Please answer the question with image and text explanation: \textless BEGIN\textgreater\  Could you help deduce how the Mycenaean civilization was destroyed? \\ \midrule

Dream Analysis and Scene Reconstruction & 
I had a dream. Please help me visualize my dream into an image, and analyze why I had this dream, what are the implications and meanings? This is the content of my dream: \textless BEGIN\textgreater\ *** &
\textless BEGIN\textgreater\  I had a dream. Please help me visualize my dream into an image, and analyze in words why I had this dream, including any implications and meanings. Here is the content of my dream: I dreamt of meeting a girl I know at the place where we first met \\ \midrule

Multimodal Biological Reasoning & 
\textless BEGIN\textgreater\  ***. Are there any more photos of this species? \textless image\textgreater\  &
\textless BEGIN\textgreater\  May I ask what species of fish this is? Are there any more photos of this species? \textless image\textgreater\  \\ \midrule

Multimodal Landscape Reasoning & 
\textless BEGIN\textgreater\  ***. Could you provide me with more photos of this and introduce them to me? \textless image\textgreater\  &
\textless BEGIN\textgreater\  Which city are these photos from? Could you provide me with more landscape photos of this city and introduce them to me? \textless image\textgreater\  \\ \midrule

Multimodal Analogy Reasoning & 
Please answer this question with image and text explanation: \textless BEGIN\textgreater\ *** &
Please answer this question with image and text explanation: \textless BEGIN\textgreater\  What should be filled in the question mark to make it exhibit a certain regularity? \textless image\textgreater\  \\ \midrule

Interactive Jigsaw Puzzle & 
\textless BEGIN\textgreater\  ***.
Show the resulting image with the corresponding text explanation. &
\textless BEGIN\textgreater\  Here are some puzzle pieces. Please assemble them into a complete picture. Show the resulting image with the corresponding text explanation. \textless image\textgreater\  \\ \midrule

Interactive Novel View Synthesis & 
\textless BEGIN\textgreater\  ***

Please draw the picture and give descriptions. &
\textless BEGIN\textgreater\  This is a pear slice. Its appearance features are:. Can you guess what it looks like from the side? Please draw the picture and give descriptions. \textless image\textgreater\  \\ \midrule

Interactive Multi-concept Image Composition & 
\textless BEGIN\textgreater\  ***. Please summarize all the content in one image and write a blog post. \textless image\textgreater\ \textless image\textgreater\  &
\textless BEGIN\textgreater\  This is a collection of four Christmas smoothies. Please summarize all the content in one image and write a blog post. \textless image\textgreater\ \textless image\textgreater\ \textless image\textgreater\ \textless image\textgreater\  \\ \midrule

Interactive Film and Television Recommendation & 
Please output recommendations in the form of the poster and the corresponding introduction: \textless BEGIN\textgreater\  *** &
Please output recommendations in the form of the poster and the corresponding introduction: \textless BEGIN\textgreater\  Could you recommend some Indian dramas to me? \\ \midrule

Interactive Goods Recommendation & 
Please output recommendations in the form of images and give the corresponding introduction: \textless BEGIN\textgreater\  *** &
Please output recommendations in the form of images and give the corresponding introduction: \textless BEGIN\textgreater\  Are there any throw pillows you can recommend? \\ \midrule

Interactive Food Recommendation & 
\textless BEGIN\textgreater\  ***. Please provide the information with images and text. &
\textless BEGIN\textgreater\  What are some recommended dishes in Yibin, Sichuan? Please provide the information with images and text. \\ \midrule

Business Scenarios Brainstorming & 
\textless BEGIN\textgreater\  I want to start a business. Please brainstorm with me about some ways to start a business and help me figure it out.

***

Please output brainstorming results with images and explanations. &
\textless BEGIN\textgreater\  I want to start a business. Please brainstorm with me about some ways to start a business and help me figure it out. Please analyze the long-term development trends of automotive braking technology for me. Please output brainstorming results with images and explanations. \\ \midrule

Academic Scenarios Brainstorming & 
\textless BEGIN\textgreater\  What/Why/How ***? Please also show an illustration. &
\textless BEGIN\textgreater\  What are the steps involved in the synthesis of Metal-Organic Frameworks (MOFs) using the hydrothermal method? Please also show an illustration. \\ \midrule

Multimodal Action Anticipation & 
In this task, you are given the first part of an activity with both text and an image, and you need to complete the subsequent action parts of the activity by generating text and images that are natural continuation of the given first part. The input interleaved content is:

\textless BEGIN\textgreater\  *** &
In this task, you are given the first part of an event with both text and an image, and you need to complete the subsequent parts of the event by generating text and images that are natural continuation of the given first part. The input interleaved content is: \textless BEGIN\textgreater\  A boy is trying to go through the security gate at the airport. \textless image\textgreater\  \\ \midrule

Visual Traffic Forecasting & 
\textless BEGIN\textgreater\  What will the traffic conditions ***? Please provide an explanation and present it in the form of images. &
\textless BEGIN\textgreater\  What will the traffic conditions be like near Fuxing Road in Shenzhen in an hour? Please provide an explanation and present it in the form of images. \textless image\textgreater\  \\ \midrule

Interactive Remote Sensing Image Rendering & 
\textless BEGIN\textgreater\  ***

Also give interleaved text explanations for generated images. &
\textless BEGIN\textgreater\  Please generate a remote sensing satellite image of the area based on my geographical photo. Also give interleaved text explanations for generated images. \textless image\textgreater\  \\ \midrule

Interactive Street View Image Rendering & 
\textless BEGIN\textgreater\  ***

Also give interleaved text explanations for generated images. &
\textless BEGIN\textgreater\  Please generate a panorama of the area based on my remote sensing satellite image. Also give interleaved text explanations for generated images. \textless image\textgreater\  \\ \midrule

Urban Planning and Development Simulation & 
Please output the scheme in the form of both the image and the text explanation to meet the requirements: \textless BEGIN\textgreater\  *** &
Please output the scheme in the form of both the image and the text explanation to meet the requirements: \textless BEGIN\textgreater\  In accordance with this planning diagram, please design a final rendering. \textless image\textgreater\  \\ \midrule

Plog and Social Media Content Generation & 
\textless BEGIN\textgreater\  ***. Could you create a social media post with text and images? &
\textless BEGIN\textgreater\  After finishing Jia Pingwa's "Comfortably Alone," I am deeply moved and want to post something on social media but don't know how to phrase it. Could you help me create a post with text and images for my use? \\ \midrule

Chat with Memes & 
You are a funny chatbot that responds to my small talk. Please output meme images to interact with me and chat with me. \textless BEGIN\textgreater\  *** &
You are a funny chatbot that responds to my small talk. Please output meme images to interact with me and chat with me. \textless BEGIN\textgreater\  We'll be traveling in three weeks! \textless image\textgreater\  \\ \midrule

Interactive Virtual Try-on & 
\textless BEGIN\textgreater\  Please generate a visualization of ***, and provide an evaluation of the try-on effect. &
\textless BEGIN\textgreater\  Please generate a visualization of how the clothing looks when worn, based on the photos of the clothing and the model I provided, and provide an evaluation of the fitting effect. \textless image\textgreater\ \textless image\textgreater\  \\ \midrule

Multimodal Dressing Suggestion & 
\textless BEGIN\textgreater\  ***. Please provide the information in both text and images. &
\textless BEGIN\textgreater\  What are some outfit suggestions for women traveling in the summer? Please provide the information in both text and images. \\ \midrule

Fashion Trend Forecasting & 
\textless BEGIN\textgreater\  What are the *** trend in the upcoming ***? Please provide the information in both text and images. &
\textless BEGIN\textgreater\  What are the design elements for men's shoes in the upcoming autumn and winter? Please provide the information in both text and images. \\ \midrule

Multimodal Recipe Generation & 
\textless BEGIN\textgreater\  How to prepare ***? Please provide the steps in a detailed format with images and text. &
\textless BEGIN\textgreater\  How to prepare this type of soy sauce boiled pomfret: Please provide the steps in a detailed format with images and text. \textless image\textgreater\  \\ \midrule

Multimodal Cooking Assistant & 
Please output instructions in interleaved images and texts: \textless BEGIN\textgreater\  &
Please output instructions in interleaved images and texts: \textless BEGIN\textgreater\  The batter I made for the egg burger doesn't taste good. How can I make it taste better? \textless image\textgreater\  \\ \midrule

Interactive Tutorial Generation & 
Please show me the steps of the tutorial with interleaved images and text: \textless BEGIN\textgreater\  &
Please show me the steps of the tutorial with interleaved images and text: \textless BEGIN\textgreater\  Please tell me how to seal or protect the finish of painted wood. \\ \midrule

Interactive Science Popularization & 
\textless BEGIN\textgreater\  What is ***? Please explain with illustrations and text. &
\textless BEGIN\textgreater\  What is the Doppler Effect? Please explain with illustrations and text. \\ \midrule

Fitness and Health Consulting & 
Please give answers in interleaved images and texts: \textless BEGIN\textgreater\ *** &
Please give answers in interleaved images and texts: \textless BEGIN\textgreater\  Please tell me what issues I need to pay attention to when engaging in walking exercise. \\ \midrule

Autonomous Driving and In-door Navigation & 
\textless BEGIN\textgreater\  You're an embodied AI that captures your surroundings through a camera. These are images captured in the past and present. What will the proceeding image possibly be in the next frame? &
\textless BEGIN\textgreater\  Assume that you are an embodied-AI agent and perceive the surroundings through a camera. You were presented with a series of three images from the past to present. Try to determine what the proceeding image could possibly be. \textless image\textgreater\  \\ 
\bottomrule
\label{tab:prompt_examples}
\end{longtable}
\clearpage
\twocolumn

\onecolumn
\twocolumn

The End-to-end Interleaved Generator models, on the other hand, represent a significant shift towards multimodal generation of interleaved image-text content. MiniGPT-5, GILL, NExT-GPT and SEED-LLaMA are designed to generate interleaved text and images in a single unified process, eliminating the need for intermediate stages. This integrated approach not only reduces latency but also improves the alignment and contextual relevance between text and images. It is noted that Anole is the only model that can directly output multi-step image-text content, whereas it is fine-tuned based on the powerful capabilities of Chameleon~\cite{team2024chameleon}. However, the open-sourced Chameleon only releases the weights for text generation, and the weights for image generation are withheld by randomizing the corresponding parameters.



Collectively, these models demonstrate diverse strategies for addressing challenges in interleaved multimodal generation, from modular pipelines to unified architectures. This taxonomy allows comprehensive evaluation and analysis of potential directions for developing MLLMs. We detail each model by category below:

\begin{itemize}
    \item \textbf{Integrated Pipelines:}
    \begin{itemize}
        \item \textbf{GPT-4o+DALL-E·3}~\cite{openai2024hello,betker2023improving}: This pipeline leverages GPT-4o~\cite{openai2024hello} to generate text and captions for the desired image generation. The captions are subsequently fed into DALL-E·3\cite{betker2023improving} to produce the corresponding images. The final output combines the text and images in their original order, enabling multimodal content generation through a staged process.
        \item \textbf{Gemini1.5+Flux}~\cite{team2023gemini,blackforestlabs_flux}: This method integrates Gemini1.5 Pro for text generation with Flux-schnell for fast and efficient image generation. The pipeline emphasizes high-quality and coherent text-to-image alignment through a structured two-step process.
    \end{itemize}

    \item \textbf{Two-stage Generators:}
    \begin{itemize}

        \item \textbf{Emu2}~\cite{sun2024generative}: Emu2 is a 37B MLLM with multimodal generation capabilities. The pretrained Emu2 is fine-tuned separately on conversational and image data, enabling it to function as Emu2-Chat for multimodal understanding and Emu2-Gen for image generation. We implement Emu2-Chat and Emu2-Gen in a two-stage pipeline to ensure seamless interleaved outputs.
        \item \textbf{SEED-X}~\cite{ge2024seed}: SEED-X is a unified multimodal foundation model that integrates multi-granularity visual comprehension and generation. We also implement this model in a two-stage pipeline approach, generating interleaved text and images in separate stages, since the prompts for instructing the model to comprehend multimodal input and generate image tokens are different. 
        \item \textbf{Show-o}~\cite{xie2024show}: Show-o is a unified Transformer model combining autoregressive and diffusion approaches to flexibly handle multimodal understanding and generation tasks. We implement Show-o, which adopts a similar two-stage generation approach but separately producing interleaved multimodal content step by step.
        \item \textbf{Emu3}~\cite{wang2024emu3}: Emu3 is one of the latest MLLMs trained in next-token prediction, capable of generating high-quality images, text, and videos by tokenizing multimodal sequences into a discrete space and training a single transformer, achieving superior performance over various SOTA models such as SDXL and LLaVA-1.6. We implement Emu3-Chat (finetuned on multimodal understanding data) and Emu3-Gen (finetuned on visual generation data) in a two-stage pipeline to ensure seamless interleaved outputs.
        \item \textbf{VILA-U}~\cite{wu2024vila}: VILA-U is a unified foundation model integrating video, image, and language understanding and generation through a single autoregressive next-token prediction framework, achieving near SOTA performance in various multimodal tasks. We implement VILA-U in a similar two-stage generation approach since it has separate multimodal understanding and image generation abilities.
    \end{itemize}

    \item \textbf{End-to-End Generators:}
    \begin{itemize}
        \item \textbf{MiniGPT-5}~\cite{zheng2023minigpt}: MiniGPT-5 directly generates interleaved text and images in an end-to-end manner. It combines MiniGPT-4 and Stable Diffusion, using "generative vokens" to seamlessly connect the textual and visual domains for efficient and coherent generation. Its seamless integration enables efficient and coherent multimodal generation without intermediate steps. In particular, MiniGPT-5 has two different versions trained on VIST and MMDialog, respectively. We name the version trained on VIST as MiniGPT-5 because it is the most widely used. The version trained on MMDialog is named MiniGPT-5MMD.
        \item \textbf{GILL}~\cite{koh2024generating}: 
        GILL fuses frozen text-only LLMs with pretrained visual models using a mapping network. It maps text hidden states from the pretrained LLM to map text hidden states into the embedding space of an image generation model, allowing multimodal generation.
        \item \textbf{NExT-GPT}~\cite{wunext}: NExT-GPT is an end-to-end MLLM capable of processing and generating text, images, videos, and audio in any combination. We implement it by removing the video and audio generation flow and the remaining text and image generation abilities. It can directly output interleaved multimodal content through its streamlined architecture.
        \item \textbf{SEED-LLaMA}~\cite{ge2023making}: SEED-LLaMA integrates text and image generation into a unified framework through the SEED tokenizer, enabling both comprehension and generation of text and images. It offers a direct end-to-end solution for creating interleaved multimodal content.
        \item \textbf{Anole}~\cite{chern2024anole}: Anole is an end-to-end interleaved generation model fine-tuned on Chameleon, leveraging pretrained weights of Chameleon to produce high-quality interleaved text, complemented with coherent images generated by optimizing image token logits in the output layer. It is the only available model that can directly output multistep image-text content.
    \end{itemize}
\end{itemize}

\subsection{Implementation Details}

The experiments were conducted using a total of 24 A100 80G GPUs, with 8 GPUs dedicated to training IntJudge. We explored different large multimodal language models (MLLMs), including InternLM-XComposer2.5 (InternLMX2.5) and Qwen2-VL, ultimately selecting Qwen2-VL-7B as the foundational model for training IntJudge to achieve an optimal balance between efficiency and accuracy. The training process involved LoRA-based parameter-efficient fine-tuning based on LLaMA-Factory. To optimize training performance, DeepSpeed and FlashAttention-2 are adopted. We define a cutoff length of 16,240 tokens for inputs. We use a per-device batch size of 1, gradient accumulation steps of 8, a learning rate of 1.0e-4, a cosine learning rate schedule, and 20 epochs with BF16 mixed-precision enabled. The evaluation process involved sampling comparison pairs. Specifically, we conducted sampling rounds to obtain a total of \( E \) distinct battle pairs for each data instance. The sampling round value \( E \) was set to 2, resulting in 4,320 battle pairs being formed for comparison.

\subsection{Experimental Results on New Models}
\begin{table*}[h]
\centering
\small 
\setlength{\tabcolsep}{0.05mm}
\begin{tabularx}{\textwidth}{@{}>{\hspace{0.1pt}}l
    >{\centering\arraybackslash}p{1.25cm} 
    >{\centering\arraybackslash}p{1.25cm} >{\centering\arraybackslash}p{1.25cm} 
    >{\centering\arraybackslash}p{1.25cm} >{\centering\arraybackslash}p{1.25cm} 
    >{\centering\arraybackslash}p{1.25cm} >{\centering\arraybackslash}p{1.25cm} 
    >{\centering\arraybackslash}p{1.25cm} >{\centering\arraybackslash}p{1.25cm} 
    >{\centering\arraybackslash}p{1.25cm} >{\centering\arraybackslash}p{1.25cm} 
    >{\centering\arraybackslash}p{1.25cm} >{\centering\arraybackslash}p{1.25cm} 
    @{}}
\toprule
\multirow{2}{*}{\textbf{Method}} & \multicolumn{4}{c}{\textbf{Human Evaluation }} & \multicolumn{4}{c}{\textbf{GPT Evaluation }} & \multicolumn{4}{c}{\textbf{IntJudge Evaluation}} \\
\cmidrule(lr){2-5} \cmidrule(lr){6-9} \cmidrule(lr){10-13}
& \textbf{\scriptsize FDT} & \textbf{\scriptsize w/o Tie} & \textbf{\scriptsize w/ Tie (0)} & \textbf{\scriptsize w/ Tie (.5)} & \textbf{\scriptsize FDT} & \textbf{\scriptsize w/o Tie} & \textbf{\scriptsize w/ Tie (0)} & \textbf{\scriptsize w/ Tie (.5)} & \textbf{\scriptsize FDT} & \textbf{\scriptsize w/o Tie} & \textbf{\scriptsize w/ Tie (0)} & \textbf{\scriptsize w/ Tie (.5)} \\
\midrule
Human                & 83.94\% & 86.50\% & 70.78\% & 79.87\% & 82.76\% & 83.09\% & 82.27\% & 82.76\% & 85.65\% & 89.11\% & 72.62\% & 81.87\% \\
GPT-4o+DALL-E3       & 78.20\% & 80.73\% & 66.17\% & 75.19\% & 86.33\% & 86.60\% & 86.23\% & 86.44\% & 83.24\% & 86.20\% & 71.46\% & 80.01\% \\
Gemini1.5+Flux       & 66.67\% & 66.95\% & 51.97\% & 63.16\% & 73.39\% & 73.38\% & 72.75\% & 73.18\% & 66.11\% & 67.92\% & 49.58\% & 63.08\% \\
VILA-U               & 62.10\% & 62.34\% & 61.57\% & 62.19\% & 49.47\% & 49.55\% & 49.29\% & 49.56\% & 68.66\% & 58.58\% & 36.94\% & 55.41\% \\
Anole                & 52.72\% & 53.10\% & 38.96\% & 52.28\% & 53.25\% & 53.06\% & 52.60\% & 53.04\% & 56.33\% & 52.77\% & 33.85\% & 51.78\% \\
Emu3                 & 54.05\% & 55.24\% & 52.25\% & 54.95\% & 47.19\% & 47.27\% & 46.74\% & 47.30\% & 54.01\% & 54.48\% & 39.04\% & 53.21\% \\
SEED-X               & 53.25\% & 52.03\% & 38.55\% & 51.51\% & 56.46\% & 56.63\% & 55.63\% & 56.51\% & 53.76\% & 54.32\% & 36.15\% & 52.88\% \\
SEED-LLaMA           & 44.43\% & 42.47\% & 30.76\% & 44.54\% & 42.33\% & 42.13\% & 41.68\% & 42.22\% & 46.43\% & 45.49\% & 28.21\% & 47.20\% \\
Emu2                 & 40.31\% & 36.64\% & 24.78\% & 40.97\% & 42.49\% & 42.43\% & 41.52\% & 42.60\% & 36.60\% & 31.84\% & 19.36\% & 38.96\% \\
NExT-GPT             & 33.59\% & 27.74\% & 18.76\% & 34.95\% & 24.81\% & 24.62\% & 24.27\% & 24.97\% & 34.08\% & 25.94\% & 15.39\% & 35.72\% \\
Show-o               & 37.47\% & 35.97\% & 24.57\% & 40.42\% & 33.21\% & 32.81\% & 32.26\% & 33.10\% & 33.65\% & 24.22\% & 13.59\% & 35.53\% \\
MiniGPT-5MMD         & 32.26\% & 32.04\% & 30.74\% & 32.77\% & 32.59\% & 32.47\% & 32.25\% & 32.59\% & 28.98\% & 25.30\% & 14.84\% & 35.51\% \\
MiniGPT-5            & 31.47\% & 28.59\% & 19.72\% & 35.24\% & 31.18\% & 30.98\% & 30.66\% & 31.18\% & 24.65\% & 15.65\% & 9.08\% & 30.07\% \\
GILL                 & 25.96\% & 20.82\% & 14.33\% & 29.91\% & 31.47\% & 31.25\% & 30.70\% & 31.58\% & 23.23\% & 16.80\% & 10.35\% & 29.54\% \\
\bottomrule
\end{tabularx}
\vspace{-0.2cm}
\caption{Comparison of model win rates evaluated by human, GPT-4o, and our IntJudge under FDT and different tie metrics. FDT: Force Dividing Tie metric. w/o Tie: Non-tie case. w/ Tie (0) and w/ Tie (.5): Count a tie as 0 and 0.5 wins for a model in a battle, respectively.}
\label{tab:model_win_rates2}
\end{table*}

We present more main experimental results in Table \ref{tab:model_win_rates2}. The new models Emu3, VILA-U and MiniGPT-5MMD are also evaluated by Human, GPT-based, and IntJudge-based evaluators and compared with 10 established baseline models using the win rate metrics. The results of methods are ranked by their performance on FDT metric evaluated by Human. Table~\ref{tab:model_win_rates2} shows a clear hierarchy in model performance, with Human and GPT-4o+DALL-E3 leading across all metrics.

A closer look at the results reveals a consistency in rankings across evaluators. For instance, GPT-4o+DALL-E3 consistently secures second place in Human Evaluation and IntJudge Evaluation. Conventional end-to-end models, such as MiniGPT-5 and GILL, struggle to match the quality of their competitors, highlighting their limitations in generating contextually relevant and diverse outputs. However, GPT Evaluation shows a clear preference for outputs by GPT-4o+DALL-E3. It is verified that GPT-based judgments are not objective enough due to the inherent bias. In contrast, our proposed IntJudge shows better alignment with human judgments, supporting the reliability of IntJudge as an effective evaluation framework.

Different evaluation metrics also offer more details about model performance. The FDT metric, which forces a decision in tie cases, highlights the dominance of Human and GPT-4o+DALL-E3. However, metrics that account for ties more flexibly, such as "w/ Tie (0)" and "w/ Tie (.5)," elevate end-to-end models like VILA-U and Emu3, suggesting that these models produce outputs that, while not always definitive winners, are frequently competitive. This distinction underscores the importance of using diverse metrics to capture various dimensions of model performance.

The new two-stage models show promising results, with VILA-U standing out for its balanced performance across all metrics, making it a reasonable option for general interleaved image-text tasks. MiniGPT-5MMD (finetuned on MMDialog) shows slight improvements over its variant MiniGPT-5 (finetuned on VIST), indicating progress but still trailing behind the latest models. Meanwhile, Emu3 performs well under specific metrics, such as "w/ Tie (.5)," showing the potential to generate tie-worthy outputs with a certain quality.

The results also highlight the challenges faced by conventional end-to-end models, such as NExT-GPT, and GILL, which consistently underperform. These models reveal the inherent difficulty in achieving coherence and contextual relevance in interleaved generation tasks. Though Anole achieved a decent ranking as a representative end-to-end model, more advanced end-to-end models are needed for better visual generation quality.

Overall, the experimental results validate the effectiveness of IntJudge as a reliable evaluator, demonstrating its consistency with human judgments. The analysis underscores the strengths of integrated generation pipelines such as GPT-4o+DALL-E3 and Gemini1.5+Flux. and identifies opportunities for improvement in two-stage and end-to-end models. Looking forward, expanding the training dataset, enhancing model architectures and improving the evaluation methods will all be critical in driving further progress in open-ended interleaved image-text generation, pushing the boundaries of multimodal learning research.

\subsection{Main Results Breakdown}

\begin{figure*}[htbp]
	\begin{center}
		\includegraphics[trim=0cm 1cm 0 0, width=0.91\textwidth]{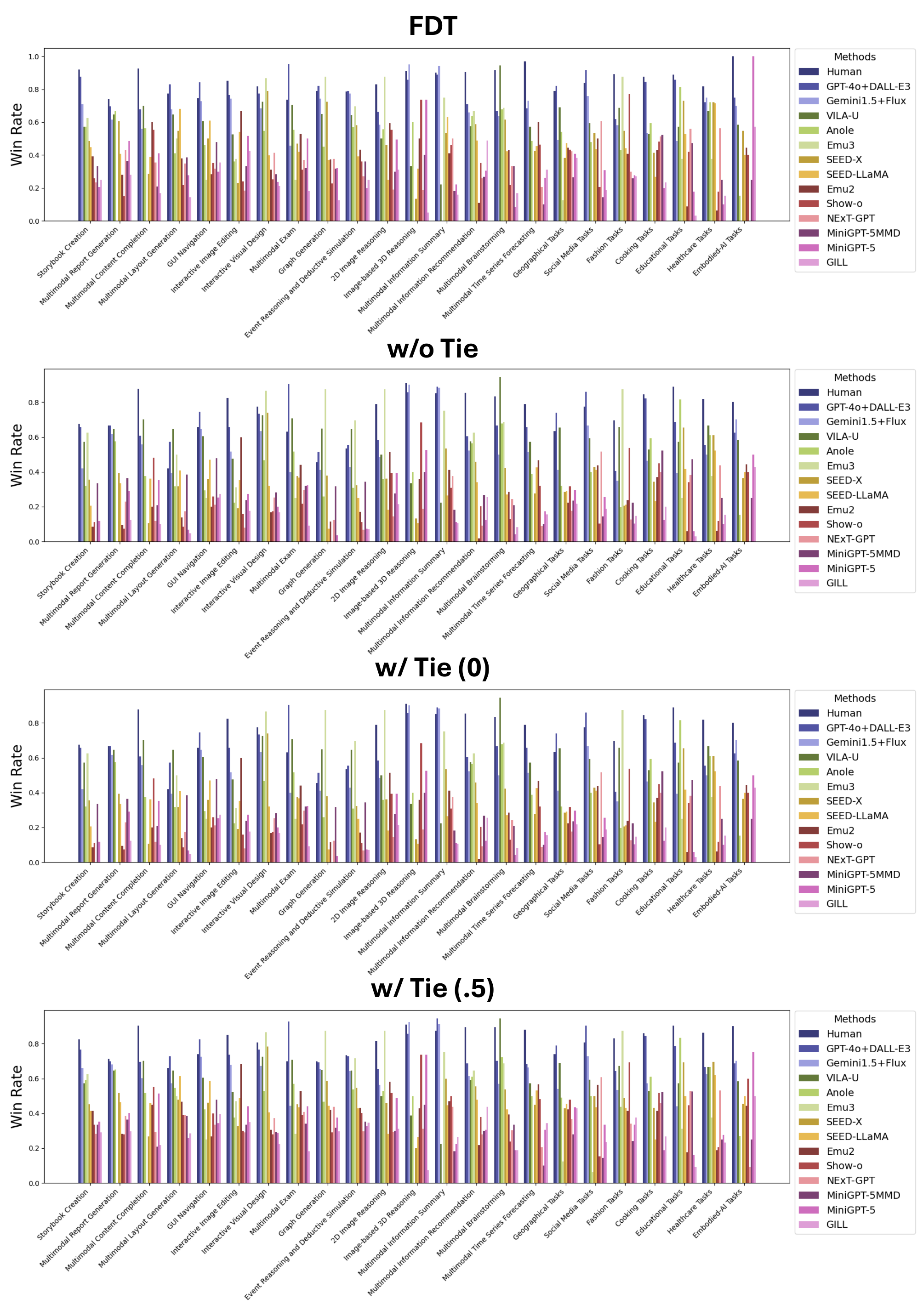}
		\caption{The win rates of 14 interleaved generation methods across 23 meta-topics.}
		\label{fig:breakdown}
	\end{center}
\end{figure*}

Fig.~\ref{fig:breakdown} presents the win rates of 14 interleaved generation methods across 23 meta-topics, evaluated solely through human evaluations. The methods are evaluated using four distinct metrics: Force Dividing Tie (FDT), Without Tie, With Tie (0), and With Tie (0.5). The results are presented using histogram figures, which provide a clear visual comparison of model performance across different topic scenarios. For example, SOTA models like GPT-4o+DALL-E3, Emu3, and VILA-U consistently ranked high in categories like "Storybook Creation," "Graph Generation," and "2D Image Reasoning," showcasing their superior capabilities in generating coherent interleaved content. Conversely, models like MiniGPT-5, NExT-GPT, and GILL struggled across most tasks, especially in areas such as "Healthcare Tasks," "Multimodal Time Series Forecasting," and "Educational Tasks," indicating a need for improved contextual understanding and generation capabilities. Training on larger datasets that include more domain knowledge may mitigate these issues and improve their interleaved generation performance.

\subsection{More Pairwise Model Performance}

\begin{figure*}[htbp]
	\begin{center}
		\includegraphics[width=1.01\textwidth]{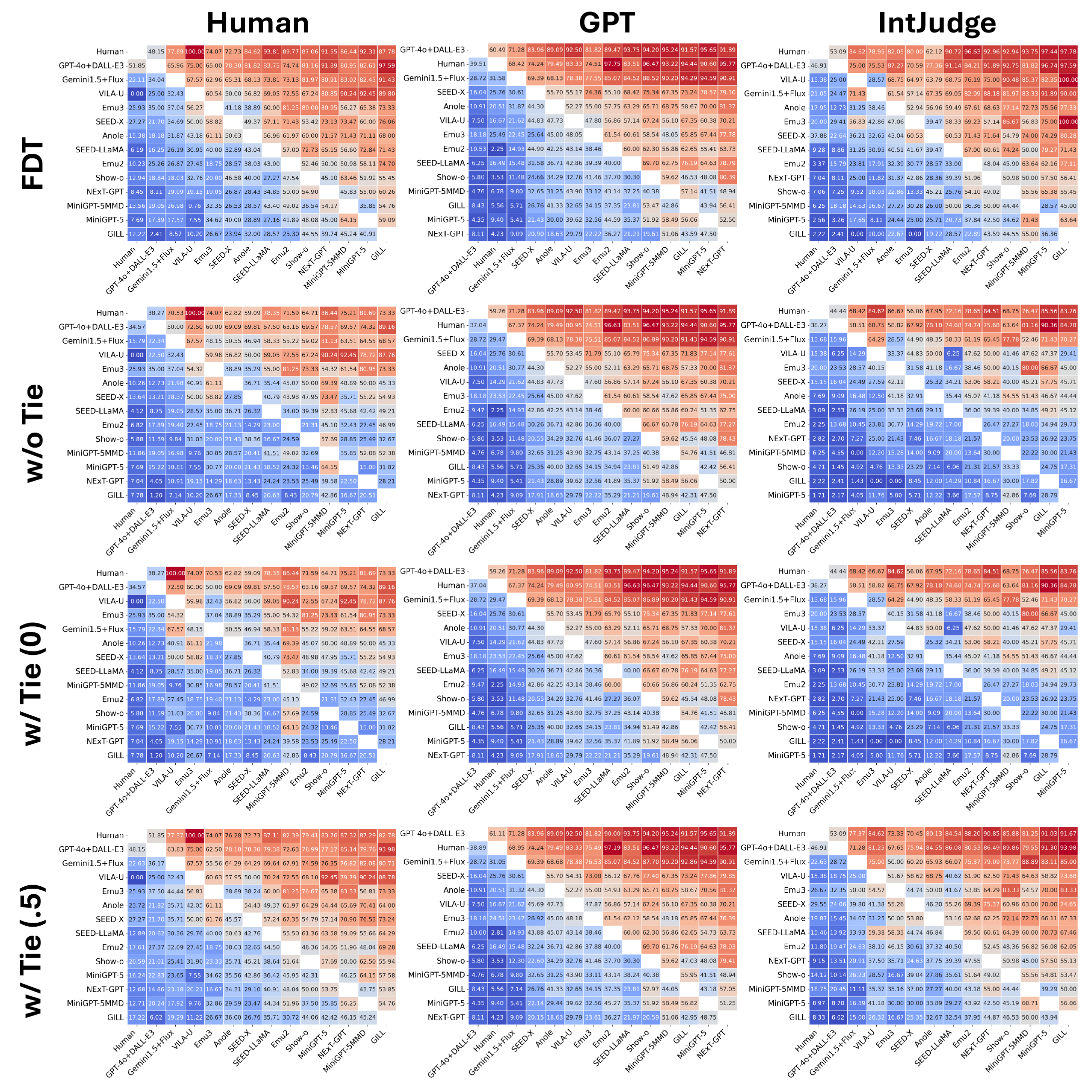}
		\caption{Win rate matrices of 14 interleaved genration methods, evaluated by Human, GPT-4o, and our IntJudge, respectively.}
		\label{fig:more_heatmap}
	\end{center}
\end{figure*}

Figure \ref{fig:more_heatmap} presents more heatmaps that illustrate the pairwise model performance evaluated by different evaluators, including Human, GPT, and IntJudge. These heatmaps provide a visual representation of the comparative strengths and weaknesses of each model across multiple metrics, such as Force Dividing Tie (FDT) and different approaches to handling tie cases (without ties, ties as zero, and ties as 0.5). By examining these heatmaps, we gain a clearer understanding of how well each model fares against others, diving deeper into performance consistency and discrepancies across evaluators.

\subsection{More Ablations on Sampling Size}

\begin{figure*}[htbp]
	\begin{center}
		\includegraphics[width=1.01\textwidth]{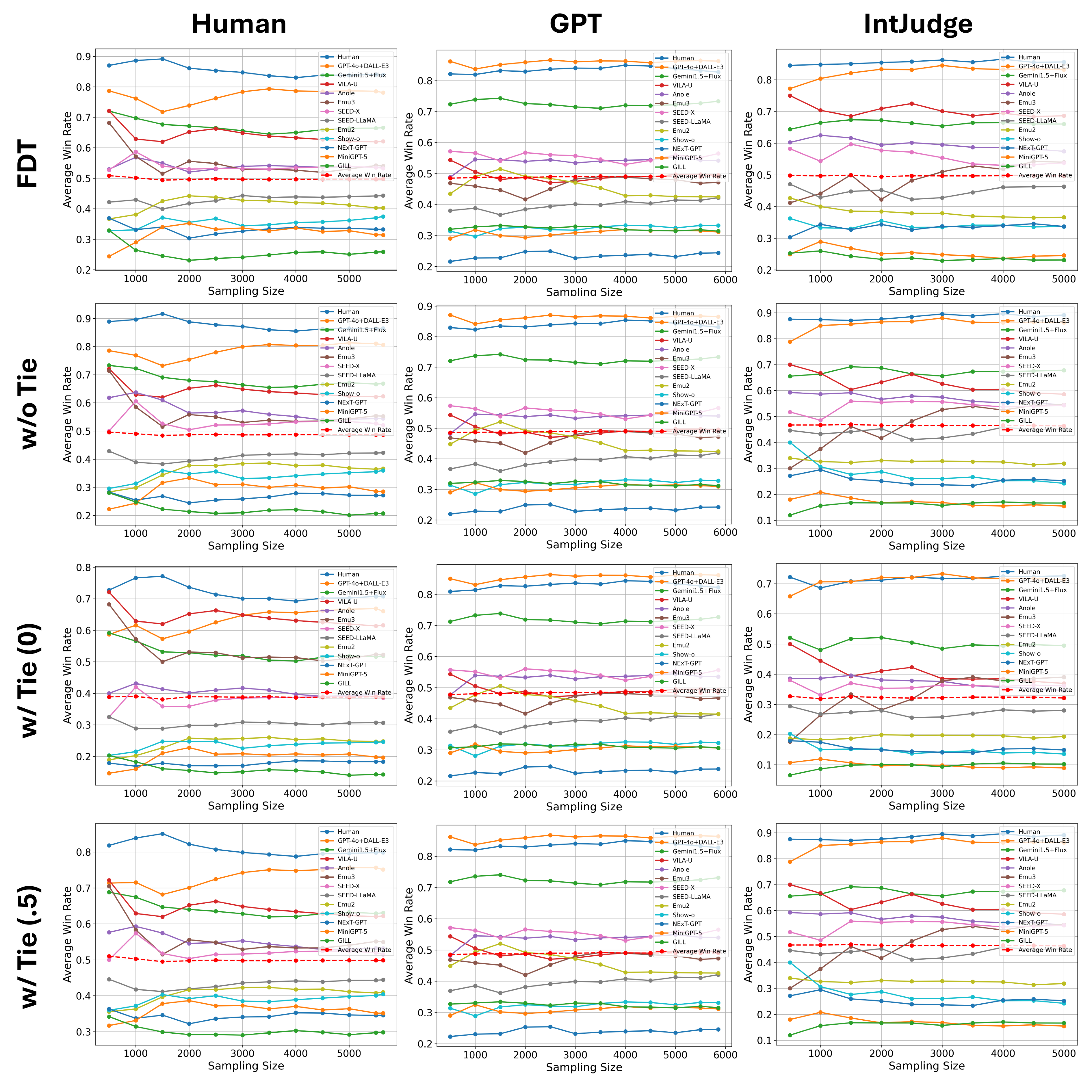}
		\caption{Win rate curves with respect to different sampling sizes.}
		\label{fig:samplesize}
	\end{center}
\end{figure*}

Figure \ref{fig:samplesize} illustrates the results of additional ablation studies focusing on the effect of sampling size on model performance. The figure compares win rates across different evaluators, including Human, GPT, and IntJudge, under various metrics such as Force Dividing Tie (FDT) and different methods for treating ties (without ties, ties as zero, and ties as 0.5). These ablation studies are crucial for understanding the impact of sampling on the robustness of model comparisons and provide insights into how sampling variations influence the ranking consistency among different evaluators. Most importantly, the results help validate the stability of our evaluation framework.

\begin{table*}[h]
\centering
\small
\setlength{\tabcolsep}{2mm}
\renewcommand{\arraystretch}{1.2}
\begin{tabular}{p{3.5cm} p{2cm} p{2cm} p{2cm} p{2cm}}
\hline
\textbf{Model} & \textbf{FDT} & \textbf{w/o Tie} & \textbf{w/ Tie (0)} & \textbf{w/ Tie (.5)} \\
\hline
Human & 84.66\% & 86.01\% & 75.46\% & 81.60\% \\
Gemini1.5+Flux & 73.44\% & 73.15\% & 61.72\% & 69.53\% \\
VILA-U & 62.50\% & 60.14\% & 41.50\% & 57.00\% \\
\textbf{MiniGPT-5OpenING} & 60.24\% & 63.76\% & 44.71\% & 59.65\% \\
Emu3 & 56.02\% & 55.20\% & 36.13\% & 53.40\% \\
SEED-X & 54.23\% & 54.67\% & 41.80\% & 53.57\% \\
Emu2 & 47.10\% & 39.33\% & 25.36\% & 43.12\% \\
Anole & 41.56\% & 43.10\% & 32.47\% & 44.81\% \\
SEED-LLaMA & 41.33\% & 38.14\% & 24.67\% & 42.33\% \\
GILL & 36.25\% & 32.04\% & 20.62\% & 38.44\% \\
Show-o & 35.76\% & 30.53\% & 19.21\% & 37.75\% \\
MiniGPT-5 & 31.54\% & 26.37\% & 16.11\% & 35.57\% \\
MiniGPT-5MMD & 28.19\% & 23.44\% & 14.71\% & 33.33\% \\
\hline
\end{tabular}
\caption{Model Win Rates evaluated by IntJudge. Representative models for each type of methods are chosen to be compared with MiniGPT-5OpenING, which is a MiniGPT-5 version finetuned on the Dev Set of OpenING.  FDT: Force Dividing Tie metric. w/o Tie: Non-tie case. w/ Tie (0) and w/ Tie (.5): Count a tie as 0 and 0.5 wins for a model in a battle, respectively.}
\label{tab:extend}
\end{table*}

\subsection{Case Study}

Here, we present case studies of interleaved content generation, which involve
14 distinct models competing in the OpenING tasks. From these models, 23 model battle pairs are formed, and each pair engages in a battle over a representative task from the 23 meta-topics in our OpenING benchmark. The results of these battles are shown in Fig.~\ref{fig:case_study}. A human judge awards a gold medal to the model that wins absolutely over its competitor in a pair. In cases where the generated content by two competing models is of similar qualities, the judge awards use tie metrics to determine and award the silver medal to the slightly better-performing model. The human judgments are based on the following criteria, which include the eight typical errors these models tend to commit in the interleaved content generation:\\
\begin{enumerate}
    \item \textbf{No-Text or No-Image}: 
    A model fails to generate text or images or both when explicitly instructed or expected to do so.
    \item \textbf{Factual Error}: 
    A model may present incorrect or made-up information, such as referencing nonexistent studies, authors, or events. Additionally, it might fail to follow instructions, examples include refusing to provide answers, misinterpreting the input and generating irrelevant responses, displaying reasoning mistakes, or producing inaccurate images.
    \item \textbf{Content Incoherent}: 
    The generated content lacks coherence with the input or is inconsistent across multiple outputs, typically in style or entities
    \item \textbf{Offensive Content}: Offensive content is defined as materials that include distorted or disturbing imagery or scenes that likely cause discomfort or distress to readers. It also includes content that raises safety concerns or violates safety guidelines, such as depictions of violence, harm, hate speech, adult content, illegal activities, dangerous behaviors, and unethical actions. A model may also fail to consider cultural nuances or sensitivities.
    \item \textbf{Image-Text Inconsistent}: This error occurs when the images do not semantically align with the corresponding text, leading to confusion or misinterpretation.
    \item \textbf{Poor Image Quality}: The generated images have low quality, such as being completely blank or blacked-out, blurry, poorly laid out, and lacking realism or aesthetic appeal.
    \item \textbf{Poor Text Quality}: The generated texts are of low quality, including nonsensical mumbling, grammatical errors that impair readability and understanding, and being too short to convey meaningful information.
    \item \textbf{Incomplete Response}: A model abruptly stops generating a textual response, or a model outputs an incomplete or truncated response.
\end{enumerate}

\section{Finetuning MLLMs on OpenING}
\label{sec:extension}

We present the extended experimental results of training MiniGPT-5 on the Dev Set of OpenING and testing the finetuned model on the Test Set of OpenING. The objective is to verify if finetuning on the specific data of OpenING can improve the performance of interleaved generation tasks. The Dev Set of OpenING can offer a set of 3,000 training samples that align with the diverse unique tasks. The MiniGPT-5 model was finetuned using the Dev Set for 5 epochs with a learning rate of 2e-5, utilizing an Adam optimizer. To enhance training stability, Adam epsilon of 1e-8 was applied. The model training incorporated mixed-precision computations to speed up the training process. The results are evaluated on the Test Set of OpenING using IntJudge.

In Table~\ref{tab:extend}, the performance of MiniGPT-5OpenING, the finetuned version of MiniGPT-5, is compared against other state-of-the-art models and the original MiniGPT-5 baselines (MiniGPT-5 is finetuned on VIST and MiniGPT-5MMD is finetuned on MMDialog). We set $E$ to 1 and randomly sampled 2,160 samples for this efficient evaluation. The evaluation metrics include four scenarios: Force Dividing Tie (FDT), Without Tie (w/o Tie), With Tie counted as 0 (w/ Tie (0)), and With Tie counted as 0.5 (w/ Tie (.5)). 

The results highlight that MiniGPT-5OpenING achieves significant improvements over the baseline MiniGPT-5 models across all metrics. For example, in the Without Tie (w/o Tie) scenario, the finetuned model shows a substantial 37.39\% relative improvement over the MiniGPT-5 baseline. These findings confirm that training on a specialized interleaved image-text dataset such as the Dev Set of OpenING enhances the model with better contextual understanding and alignment capabilities for generating coherent interleaved image-text content. Further studies are ongoing to improve the performance of SOTA models.

\section{Limitations of This Study}
\label{sec:limitations}


Although OpenING advances interleaved image-text generation evaluation, several limitations remain for improvement. First, despite covering diverse tasks (56 tasks across 23 meta-topics), some real-world scenarios remain underrepresented or oversimplified, potentially limiting the generalizability to practical applications. Tasks requiring fine-grained understanding or multi-step reasoning need to be supplied to capture real-world needs. Second, although the IntJudge model improves alignment with human evaluations, its generalizability remains limited by the diversity and quality of training data. The benchmark's reliance on human-annotated data to establish ground truth and train judge models is labor-intensive and expensive. While our Reference-Augmented Generation (RAG) approach helps scale training data, manual annotations remain a critical component for ensuring quality and alignment with human preferences. Furthermore, the computational resources required for training and deploying IntJudge present scalability challenges, potentially limiting accessibility for researchers with fewer resources. 

In addition, current interleaved image-text generation methods still struggle with producing high-quality, coherent interleaved content, particularly in multi-step tasks that require maintaining consistency across generated images and text. Issues like content incoherence, poor image quality, and mismatches between generated text and images persist across evaluated models, particularly in end-to-end approaches. To tackle these issues, more advanced MLLMs trained with a large-scale interleaved image-text dataset are to be investigated. Moreover, while our IntJudge demonstrates significant advantages over GPT-based evaluators, we recognize two aspects for improvement: 1) Potential biases may rise from crowdsourced data (e.g., human preferences in aesthetic judgment), and 2) The current foundation model of IntJudge mainly supports English and Chinese. Building a sufficiently comprehensive, diverse, and representative dataset is expected to greatly promote the development of multimodal generation. 

%
%

These limitations underscore the need for continued development of more diverse datasets and more robust evaluation frameworks to address the complexities of interleaved generation evaluation, enabling more practical interleaved image-text generation methods and pushing forward the boundary of future MLLMs. Future work may benefit from diversifying data sources and cross-cultural annotations, expanding multilingual capabilities, and implementing debiasing techniques to improve fairness.

\section*{Acknowledgement}
This work is partially supported by the National Key R\&D Program of China (No. 2022ZD0160101 and No. 2022ZD0160102).

\small
\bibliographystyle{ieeenat_fullname}
\bibliography{main}

\begin{figure*}[h]
    \caption{Examples of 23 Meta-Topics (\textbf{bold} font) and corresponding selected task (regular font) in Our OpenING Benchmark.}
    \label{fig:meta_topics}
   
    \begin{subfigure}{\textwidth}
    \centering
    \includegraphics[width=0.98\linewidth]{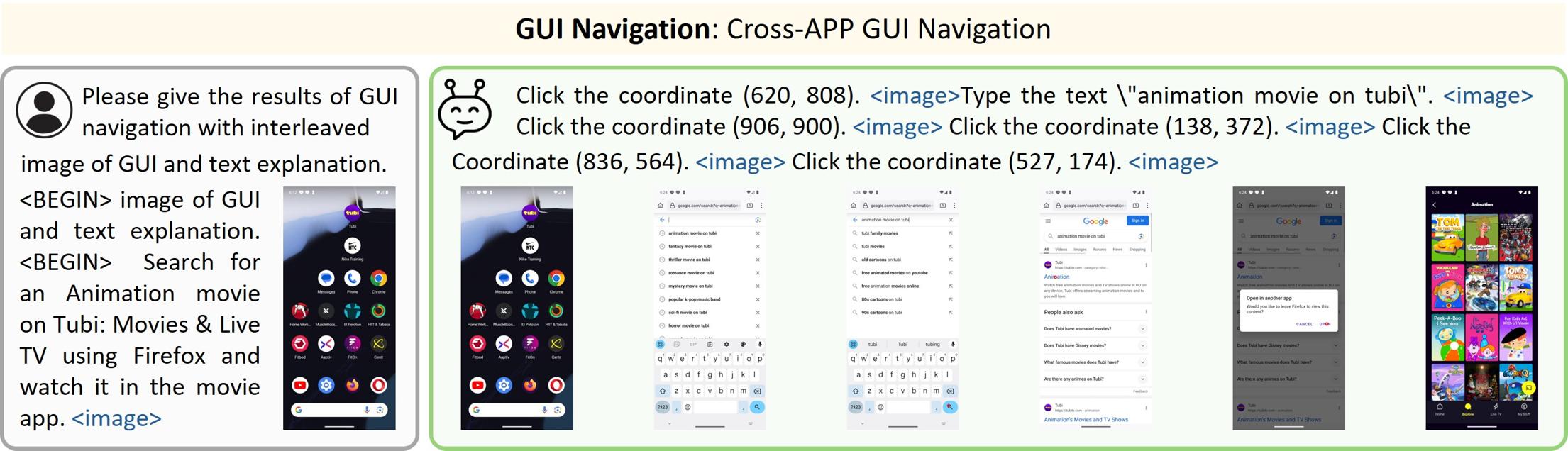} 
    \hspace{3mm}
    \end{subfigure}

    \begin{subfigure}{\textwidth}
    \centering
    \includegraphics[width=0.98\linewidth]{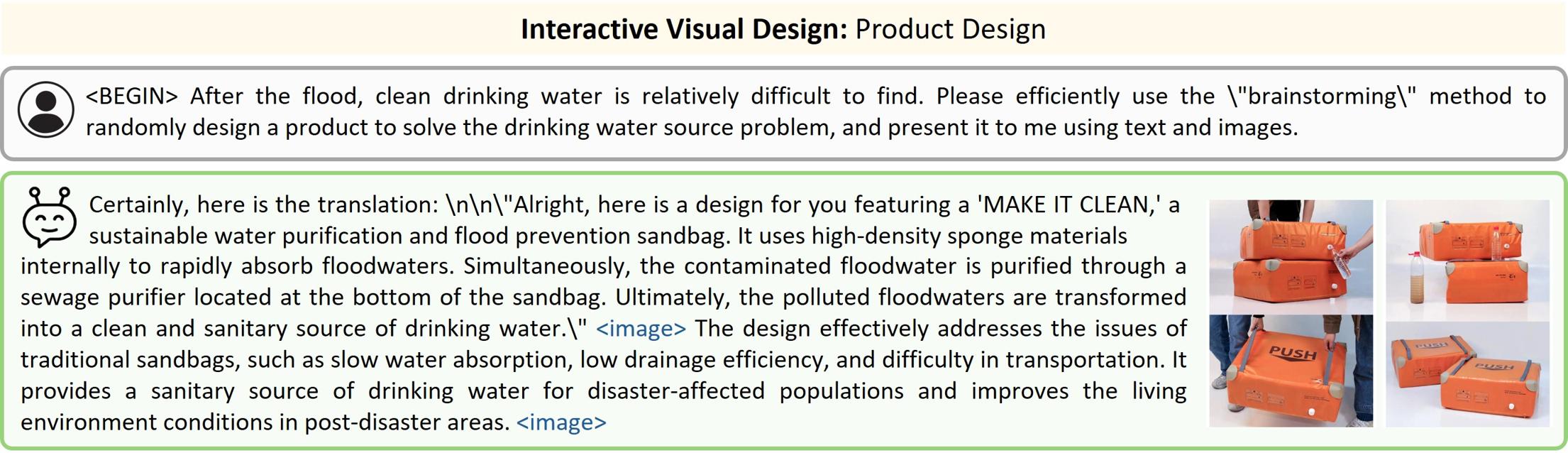} 
    \hspace{3mm}
    \end{subfigure}

    \begin{subfigure}{\textwidth}
    \centering
    \includegraphics[width=0.98\linewidth]{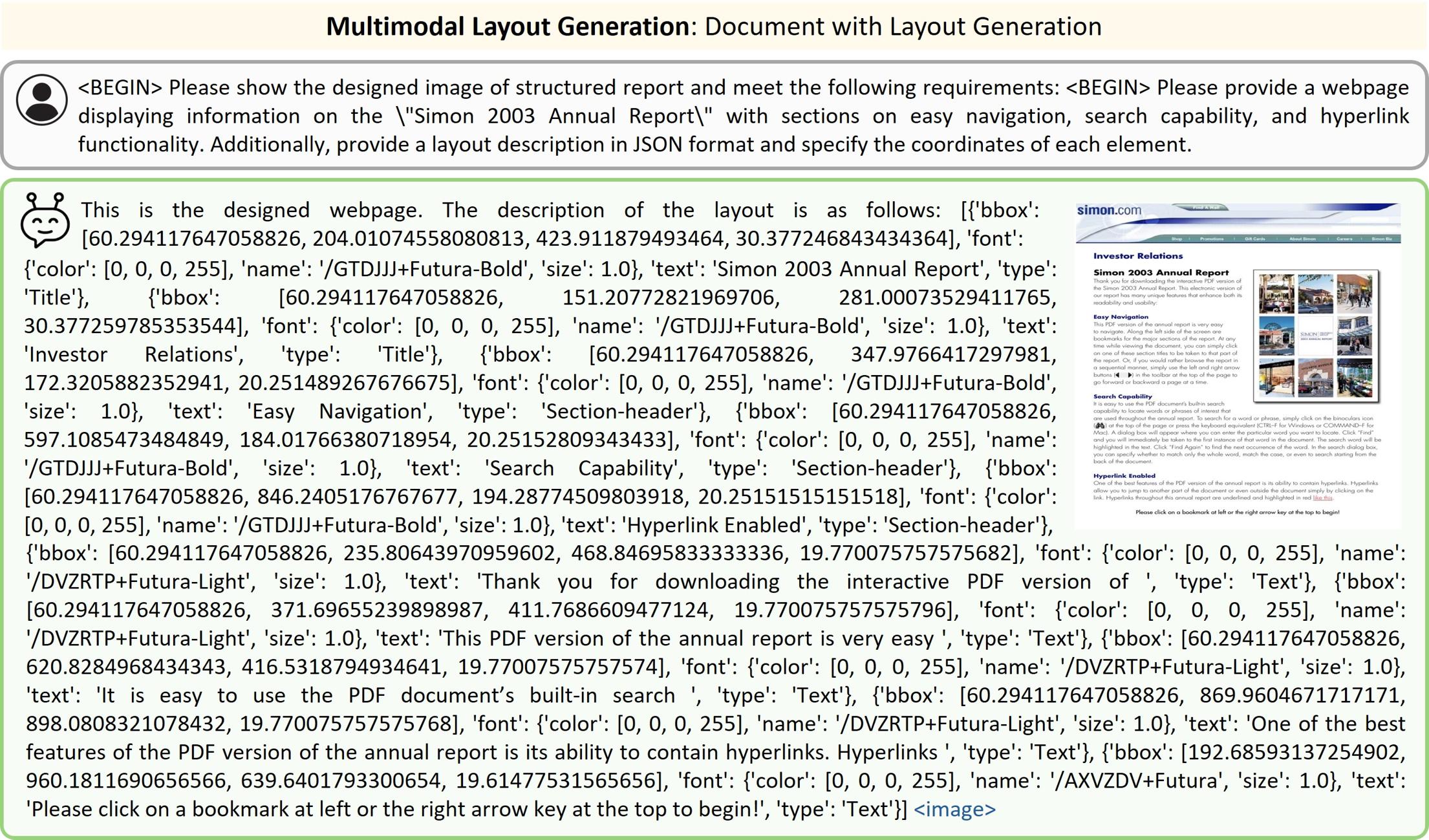} 
    \hspace{3mm}
    \end{subfigure}
\end{figure*}  

\begin{figure*}[h]\ContinuedFloat
    \begin{subfigure}{\textwidth}
    \centering
    \includegraphics[width=0.98\linewidth]{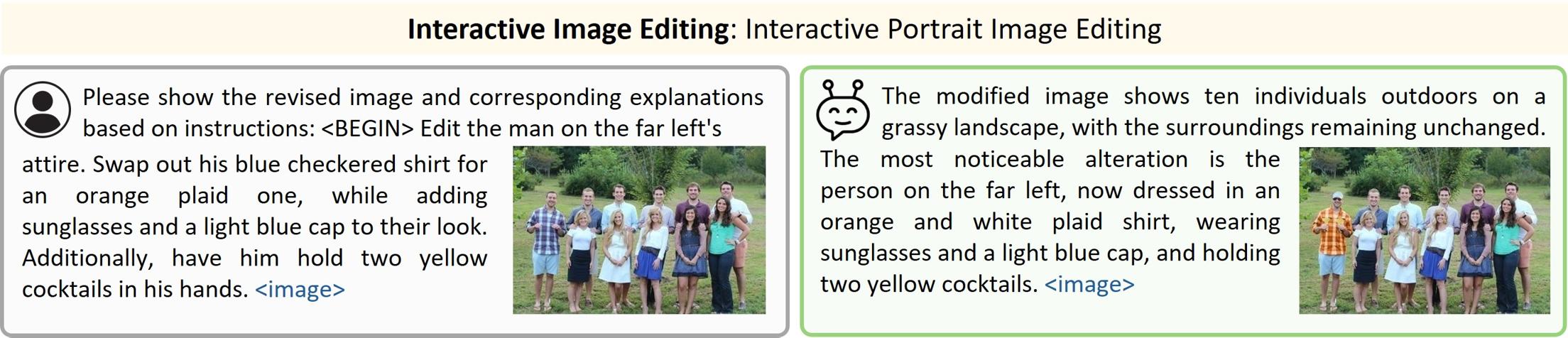} 
    \hspace{3mm}
    \end{subfigure}

    \begin{subfigure}{\textwidth}
    \centering
    \includegraphics[width=0.98\linewidth]{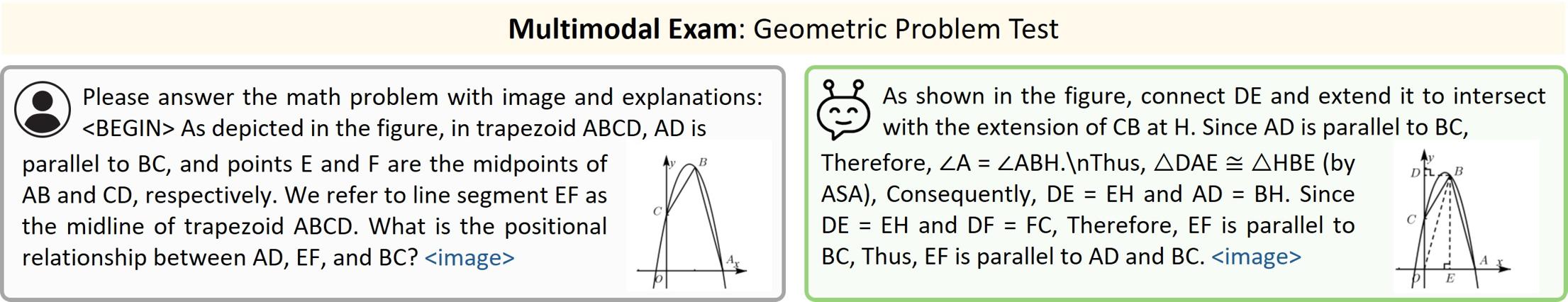} 
    \hspace{3mm}
    \end{subfigure}

    \begin{subfigure}{\textwidth}
    \centering
    \includegraphics[width=0.98\linewidth]{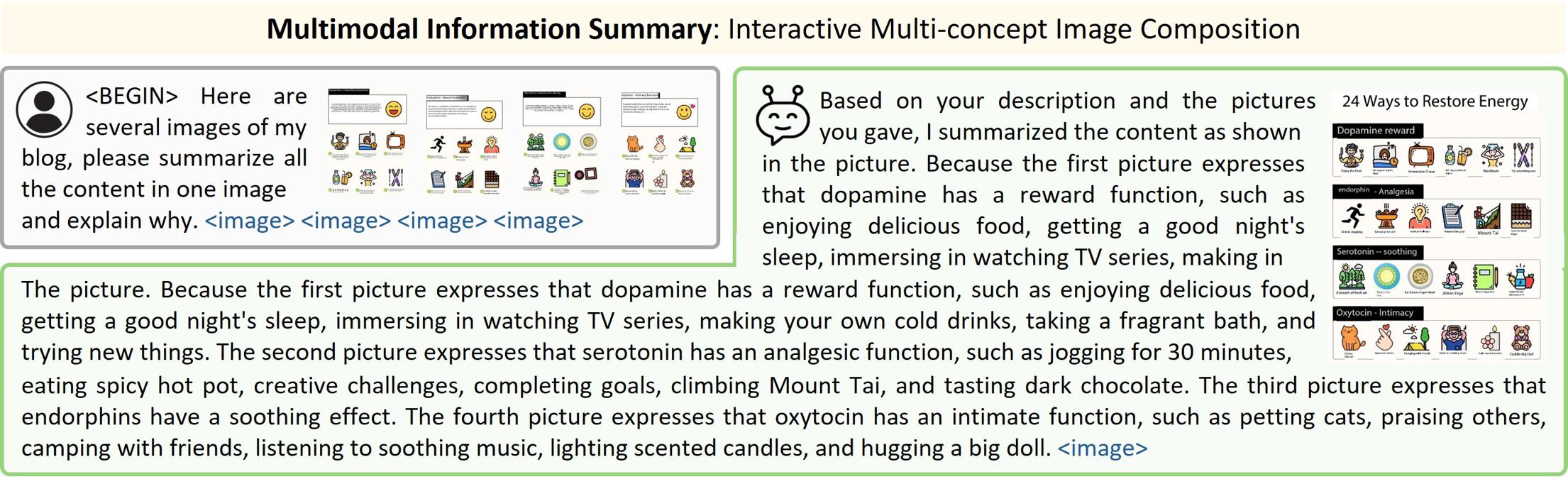} 
    \hspace{3mm}
    \end{subfigure}

    \begin{subfigure}{\textwidth}
    \centering
    \includegraphics[width=0.98\linewidth]{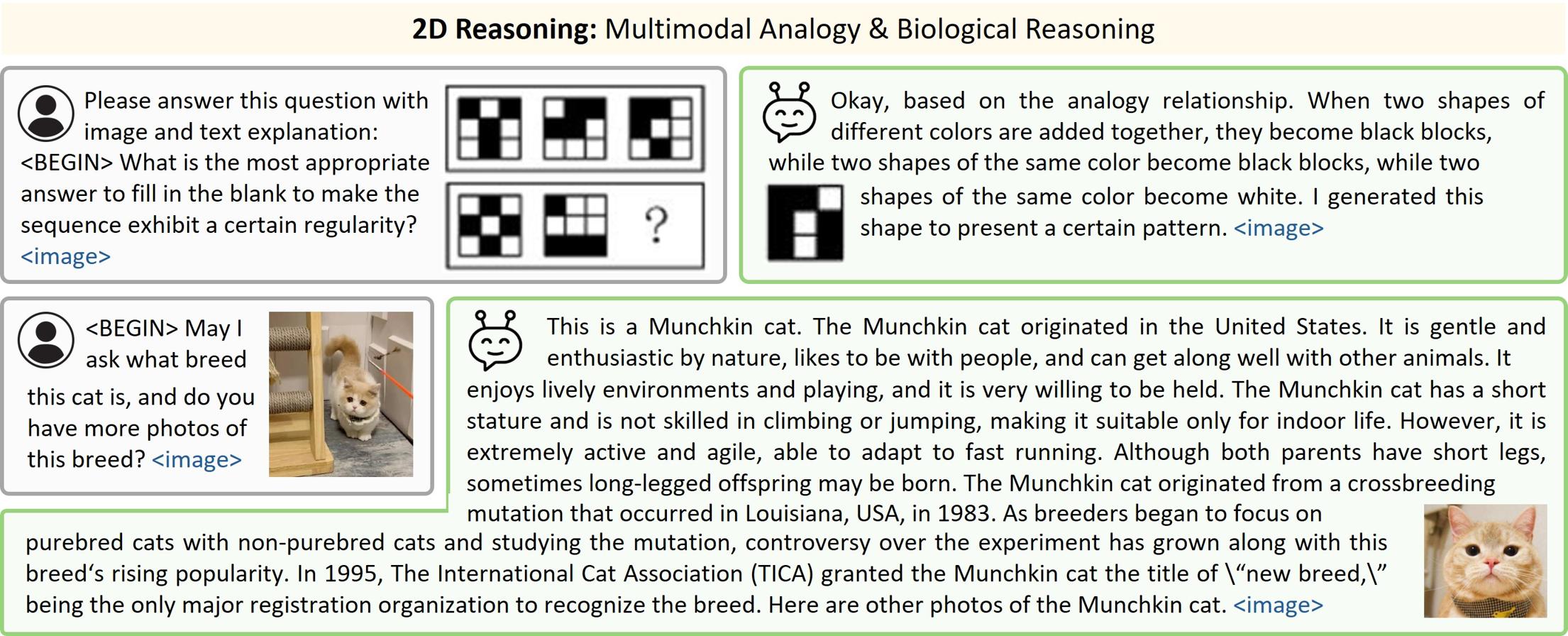} 
    \hspace{3mm}
    \end{subfigure}
\end{figure*}    

\begin{figure*}[h]\ContinuedFloat
    \begin{subfigure}{\textwidth}
    \centering
    \includegraphics[width=0.98\linewidth]{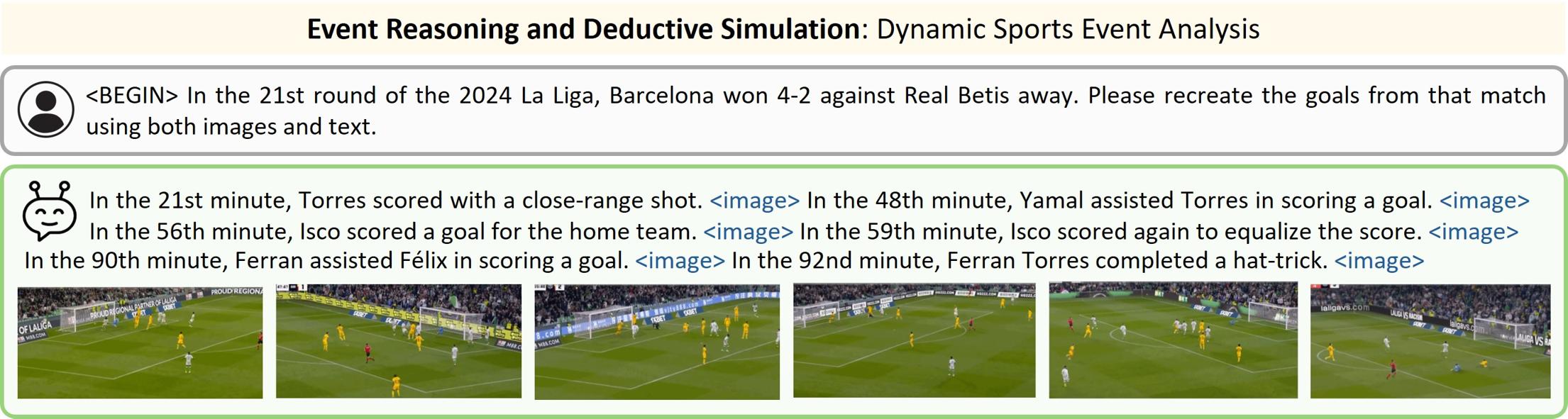} 
    \hspace{3mm}
    \end{subfigure}

    \begin{subfigure}{\textwidth}
    \centering
    \includegraphics[width=0.98\linewidth]{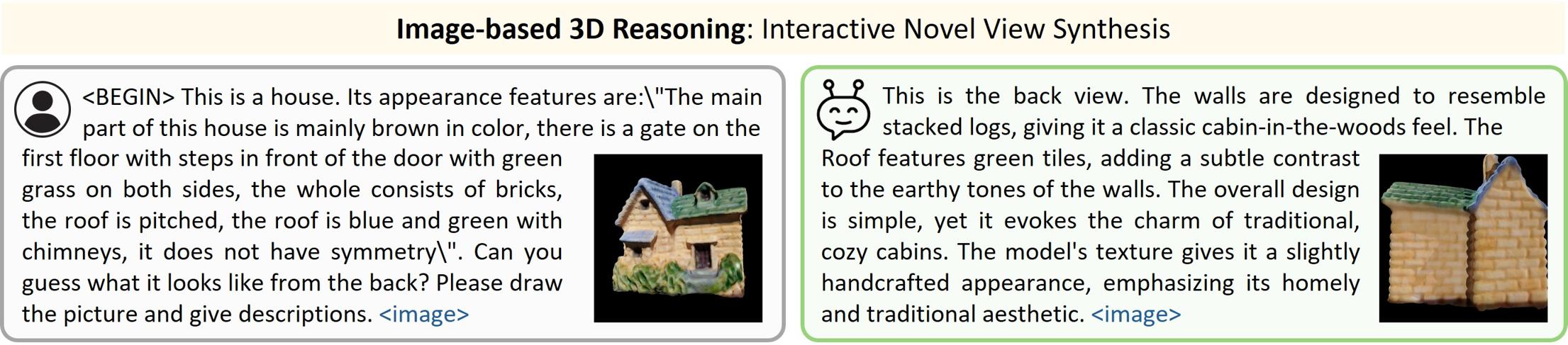} 
    \hspace{3mm}
    \end{subfigure}


    \begin{subfigure}{\textwidth}
    \centering
    \includegraphics[width=0.98\linewidth]{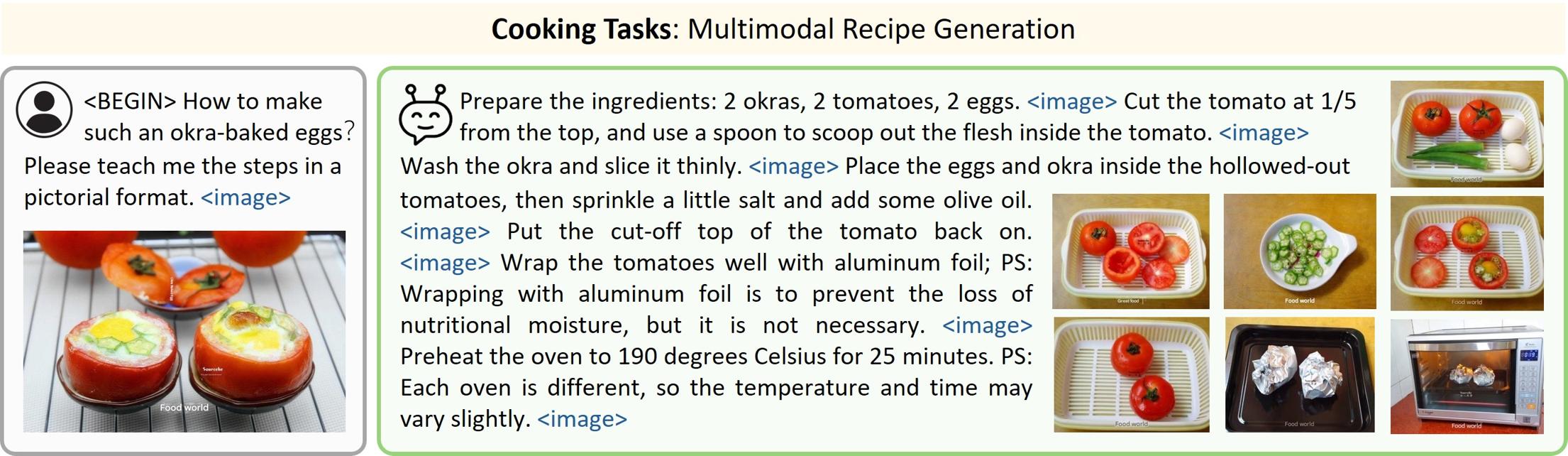} 
    \hspace{3mm}
    \end{subfigure}

    \begin{subfigure}{\textwidth}
    \centering
    \includegraphics[width=0.98\linewidth]{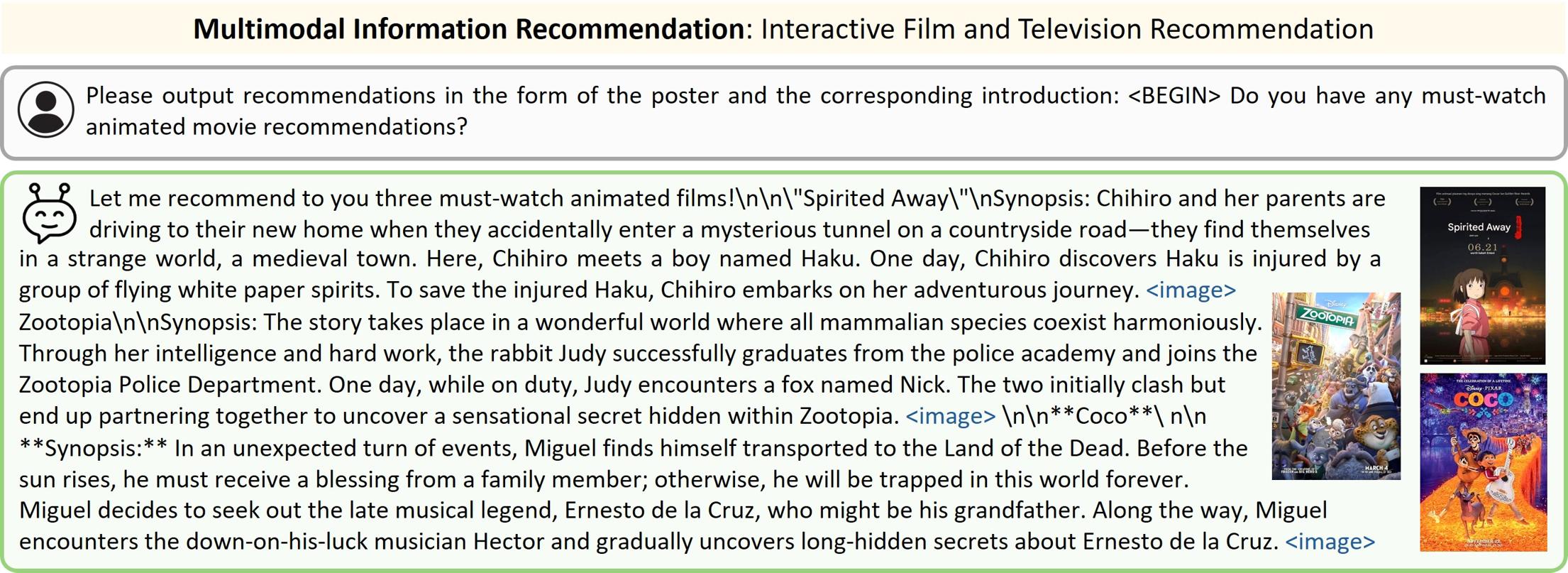} 
    \hspace{3mm}
    \end{subfigure}
\end{figure*}    

\begin{figure*}[h]\ContinuedFloat
    \begin{subfigure}{\textwidth}
    \centering
    \includegraphics[width=0.98\linewidth]{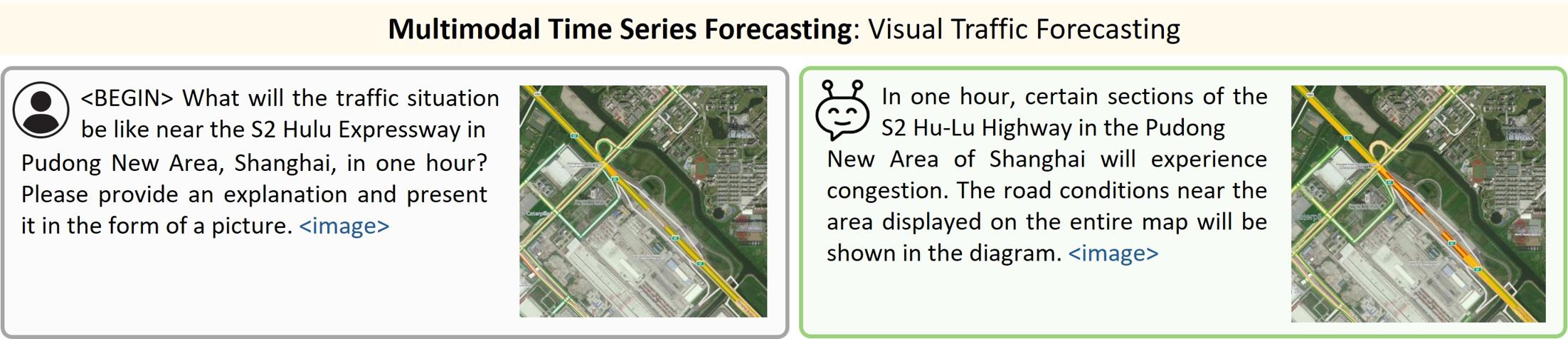} 
    \hspace{3mm}
    \end{subfigure}

    \begin{subfigure}{\textwidth}
    \centering
    \includegraphics[width=0.98\linewidth]{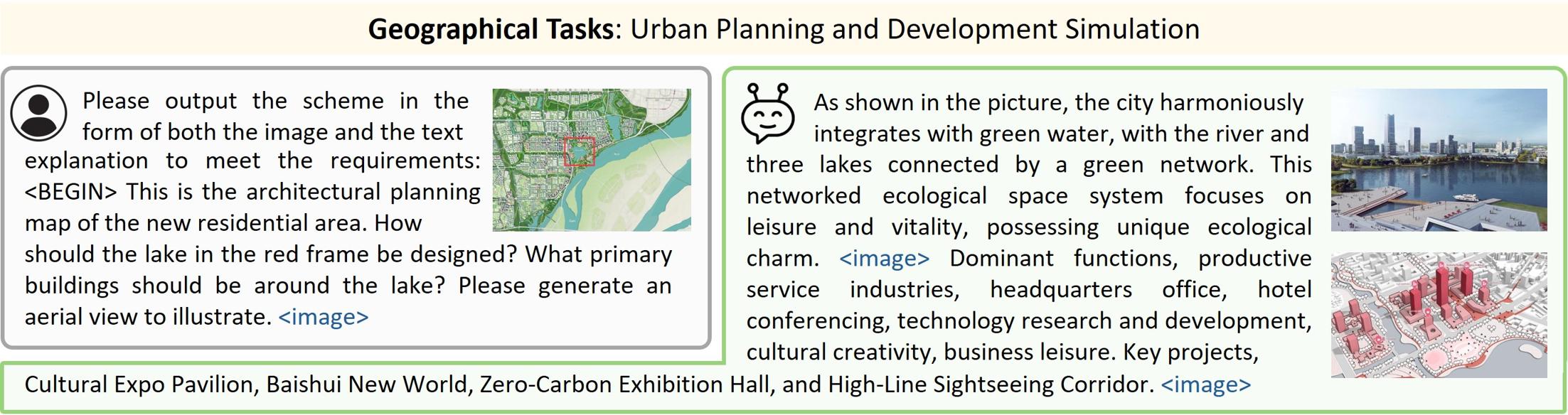} 
    \hspace{3mm}
    \end{subfigure}

    \begin{subfigure}{\textwidth}
    \centering
    \includegraphics[width=0.98\linewidth]{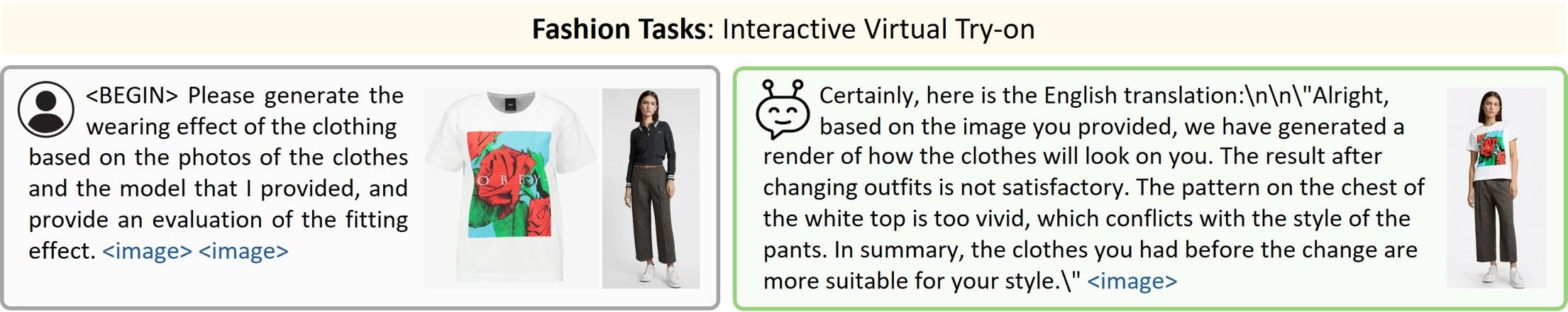} 
    \hspace{3mm}
    \end{subfigure}

    \begin{subfigure}{\textwidth}
    \centering
    \includegraphics[width=0.98\linewidth]{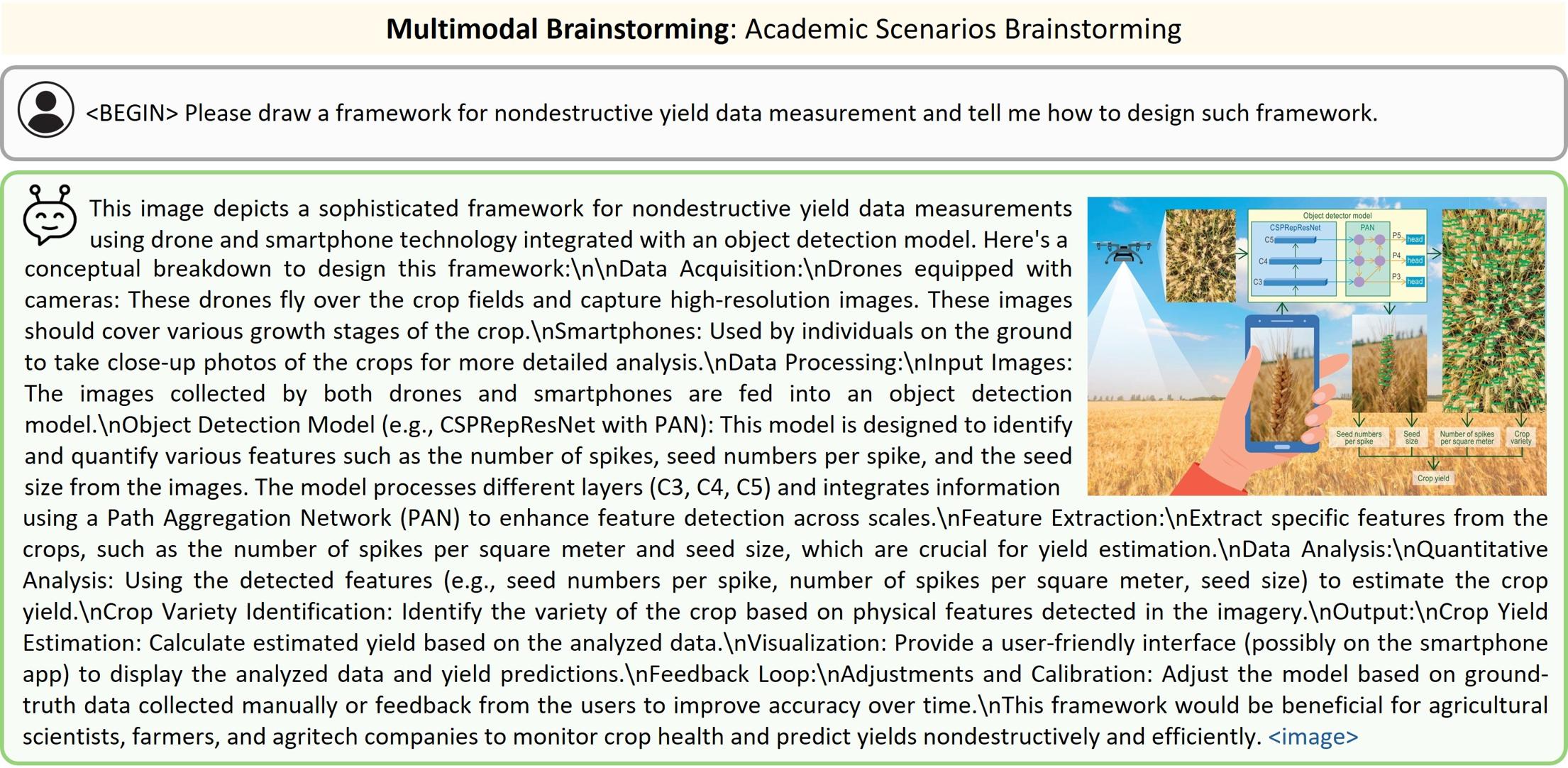} 
    \hspace{3mm}
    \end{subfigure}
\end{figure*}    

\begin{figure*}[h]\ContinuedFloat

    \begin{subfigure}{\textwidth}
    \centering
    \includegraphics[width=0.98\linewidth]{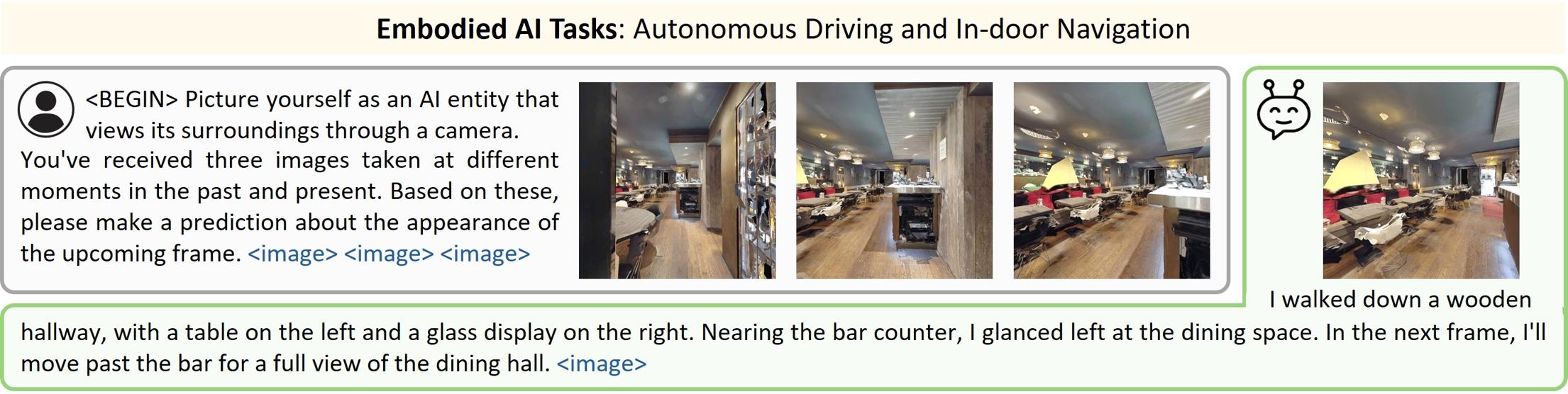} 
    \hspace{3mm}
    \end{subfigure} 
    
    \begin{subfigure}{\textwidth}
    \centering
    \includegraphics[width=0.98\linewidth]{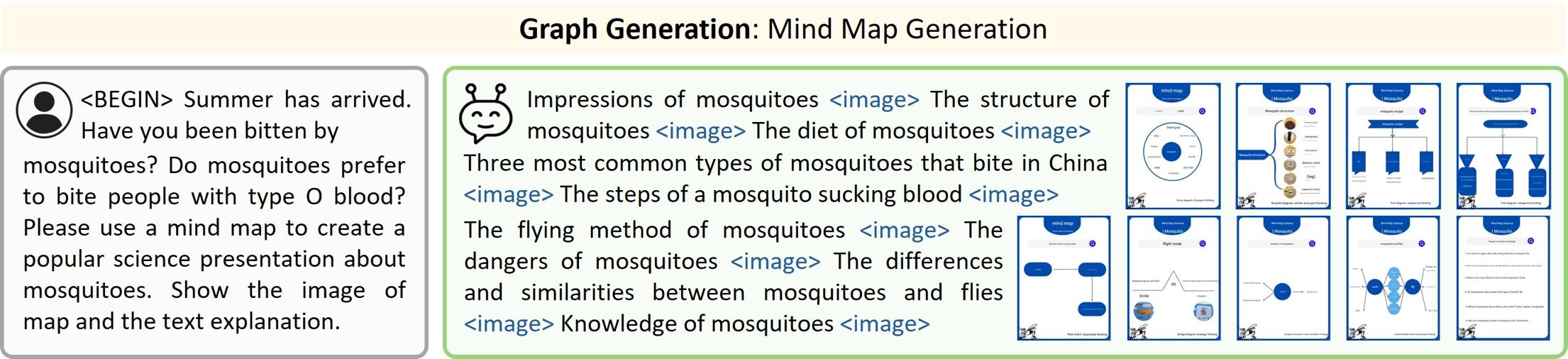} 
    \hspace{3mm}
    \end{subfigure}

    \begin{subfigure}{\textwidth}
    \centering
    \includegraphics[width=0.98\linewidth]{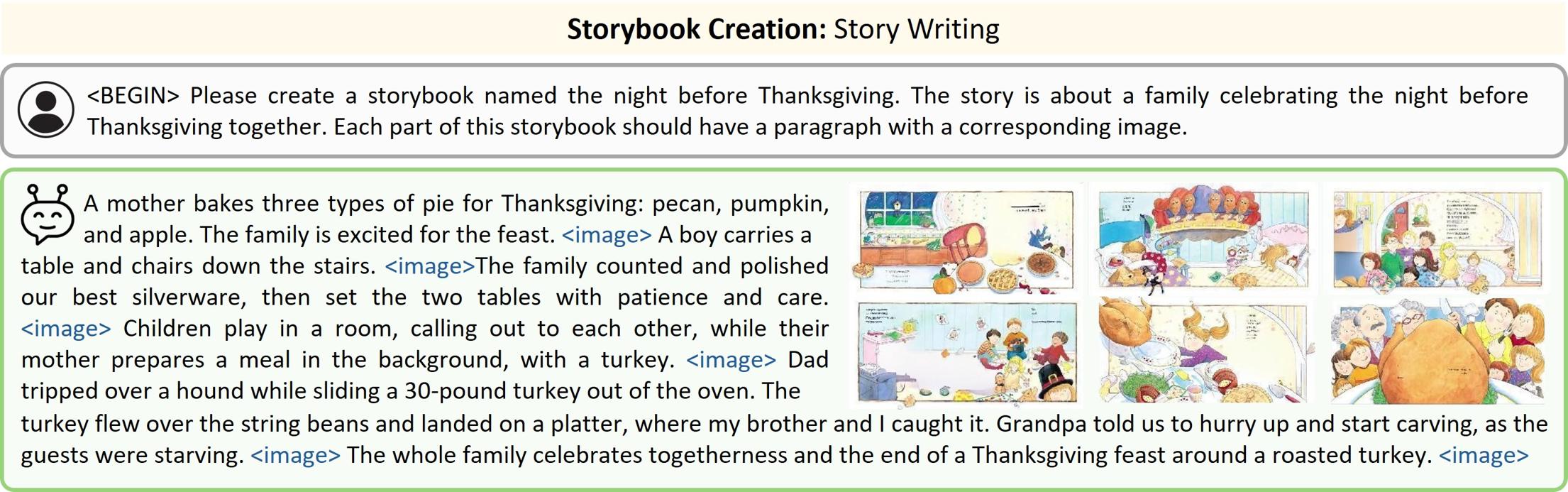} 
    \hspace{3mm}
    \end{subfigure}  

    \begin{subfigure}{\textwidth}
    \centering
    \includegraphics[width=0.98\linewidth]{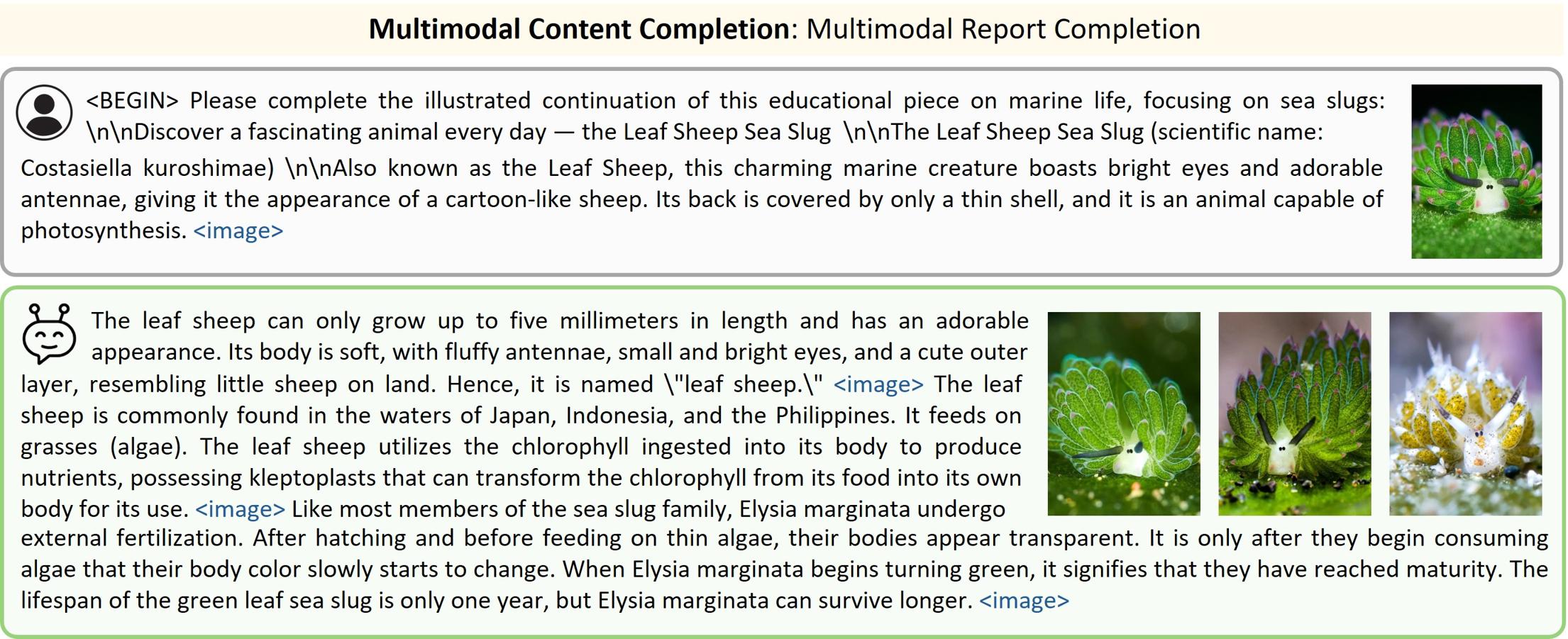} 
    \hspace{3mm}
    \end{subfigure} 

\end{figure*}    

\begin{figure*}[h]\ContinuedFloat
    \begin{subfigure}{\textwidth}
    \centering
    \includegraphics[width=0.98\linewidth]{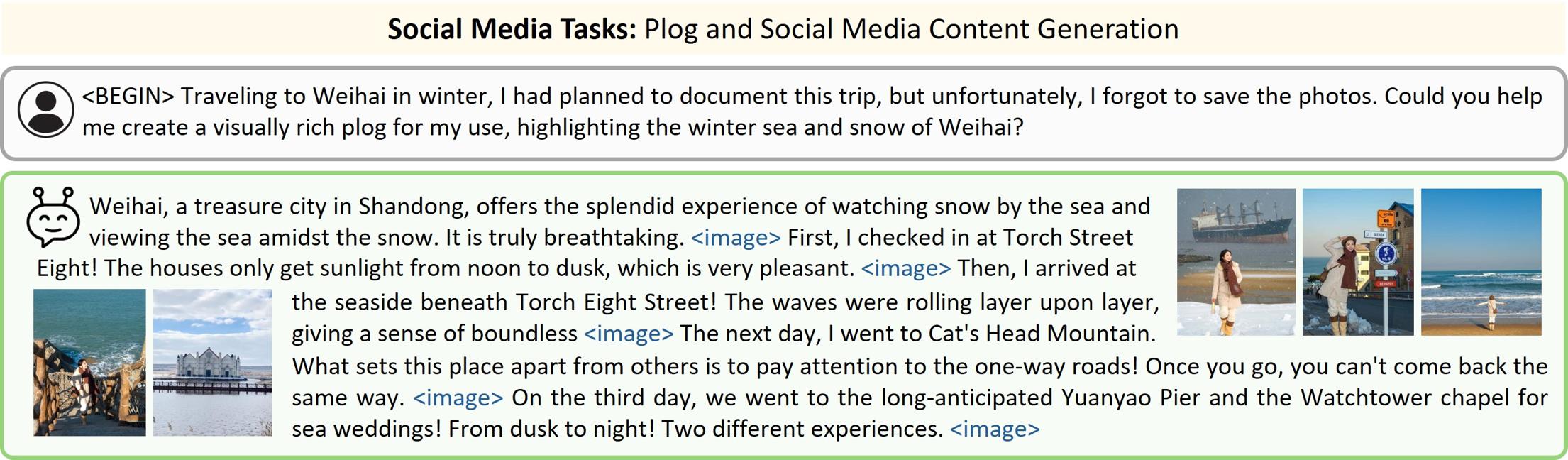} 
    \hspace{3mm}
    \end{subfigure}

    \begin{subfigure}{\textwidth}
    \centering
    \includegraphics[width=0.98\linewidth]{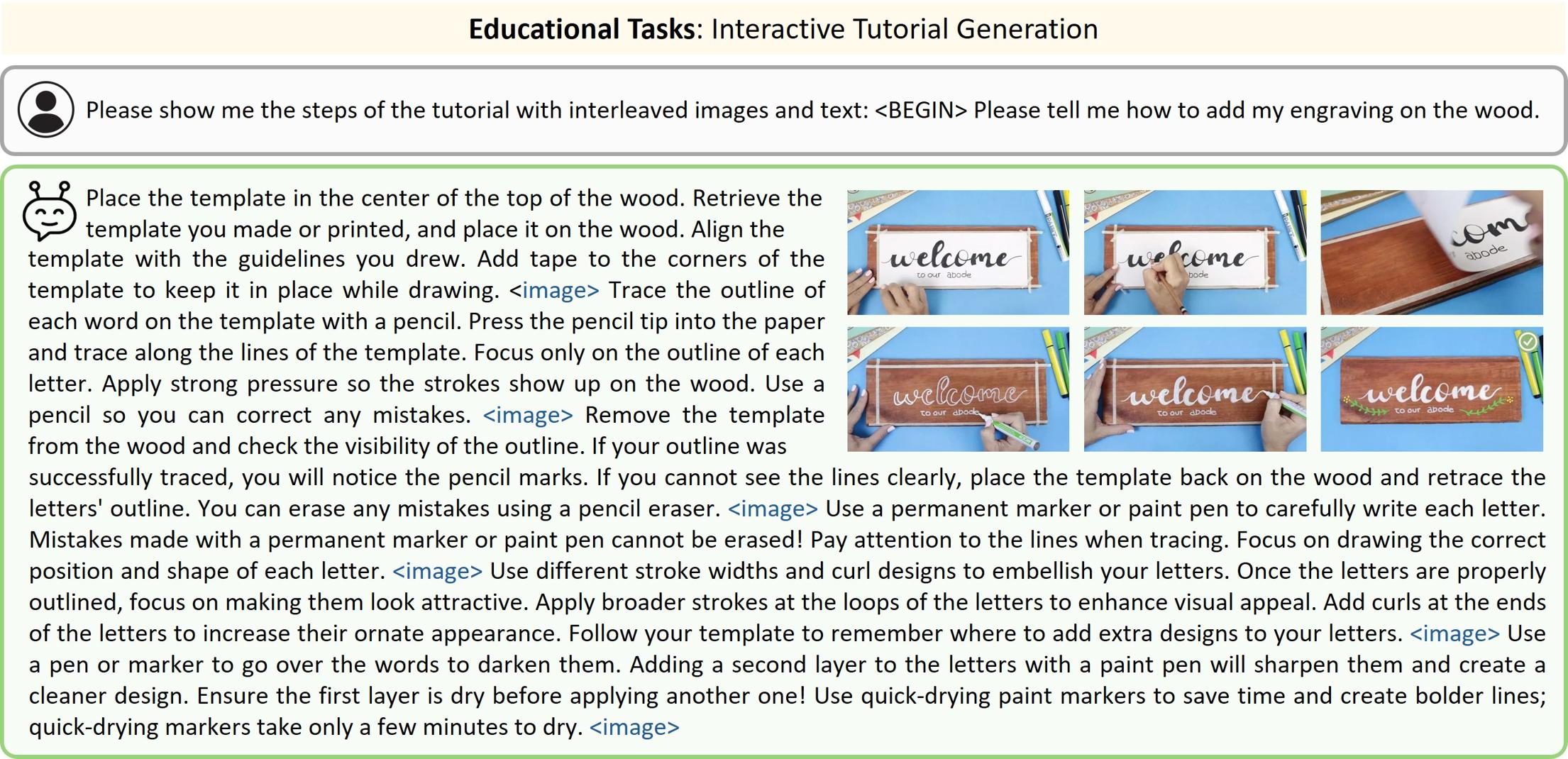} 
    \hspace{3mm}
    \end{subfigure}    

    \begin{subfigure}{\textwidth}
    \centering
    \includegraphics[width=0.98\linewidth]{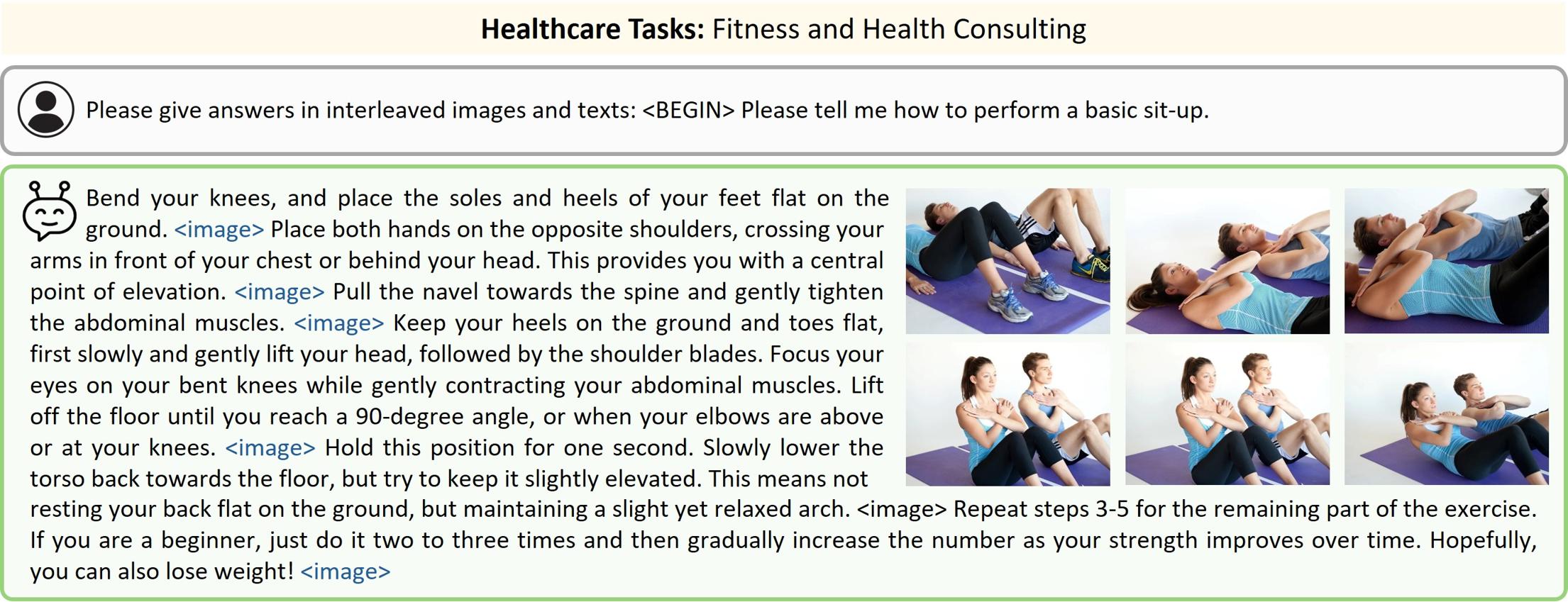} 
    \hspace{3mm}
    \end{subfigure}  
\end{figure*}    

\begin{figure*}[h]\ContinuedFloat
    \begin{subfigure}{\textwidth}
    \centering
    \includegraphics[width=0.98\linewidth]{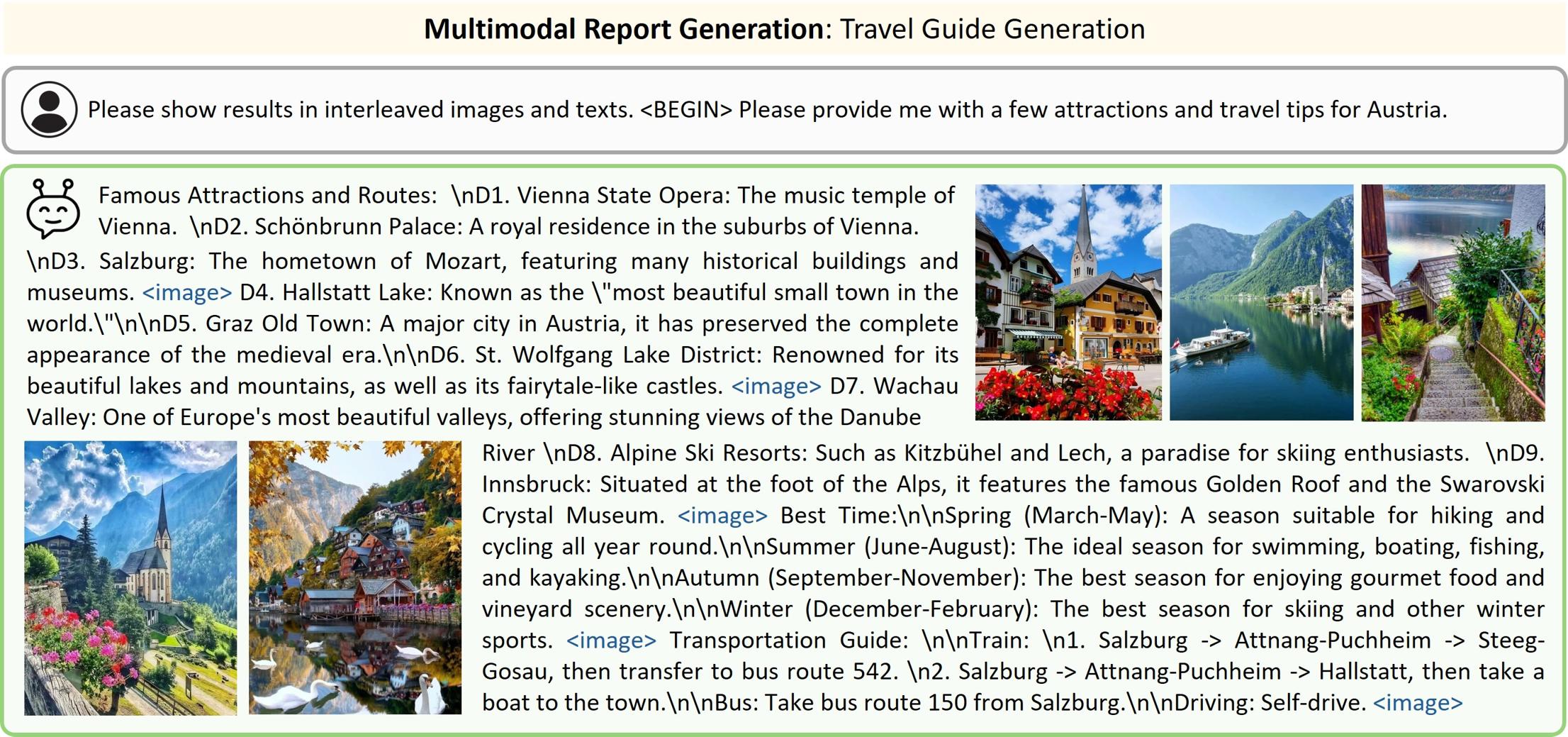} 
    \hspace{3mm}
    \end{subfigure}  
\end{figure*}   

\begin{figure*}[h]
    \caption{Illustration of 23 pairwise comparison cases. The meta-topic name (\textbf{bold} font) and task name (regular font) are given for each case. Using the eight criteria detailed in Sec.~\ref{sec:criteria}, human judges evaluate and compare the output of each model pair. Gold medal \raisebox{-0.8ex}\goldmedal \hspace{0.4mm} is awarded to a model that generates content of significantly higher quality. In a tie setting, where the quality of outputs from the A model and B model is similar, Silver medal \raisebox{-0.8ex}\silvermedal \hspace{0.4mm} is awarded to a model that generates relatively more favorable content. Errors that occur during model generation are highlighted by a red checkmark \raisebox{-0.8ex} \redcheck \hspace{0.5mm} and \mybox{\textcolor{white}{Error Types}}. Models that pass all human moderation checks are marked with a green checkmark \raisebox{-1ex}\greencheck.}
    \label{fig:case_study}
    \begin{subfigure}{\textwidth}
    \centering
    \includegraphics[width=0.98\linewidth]{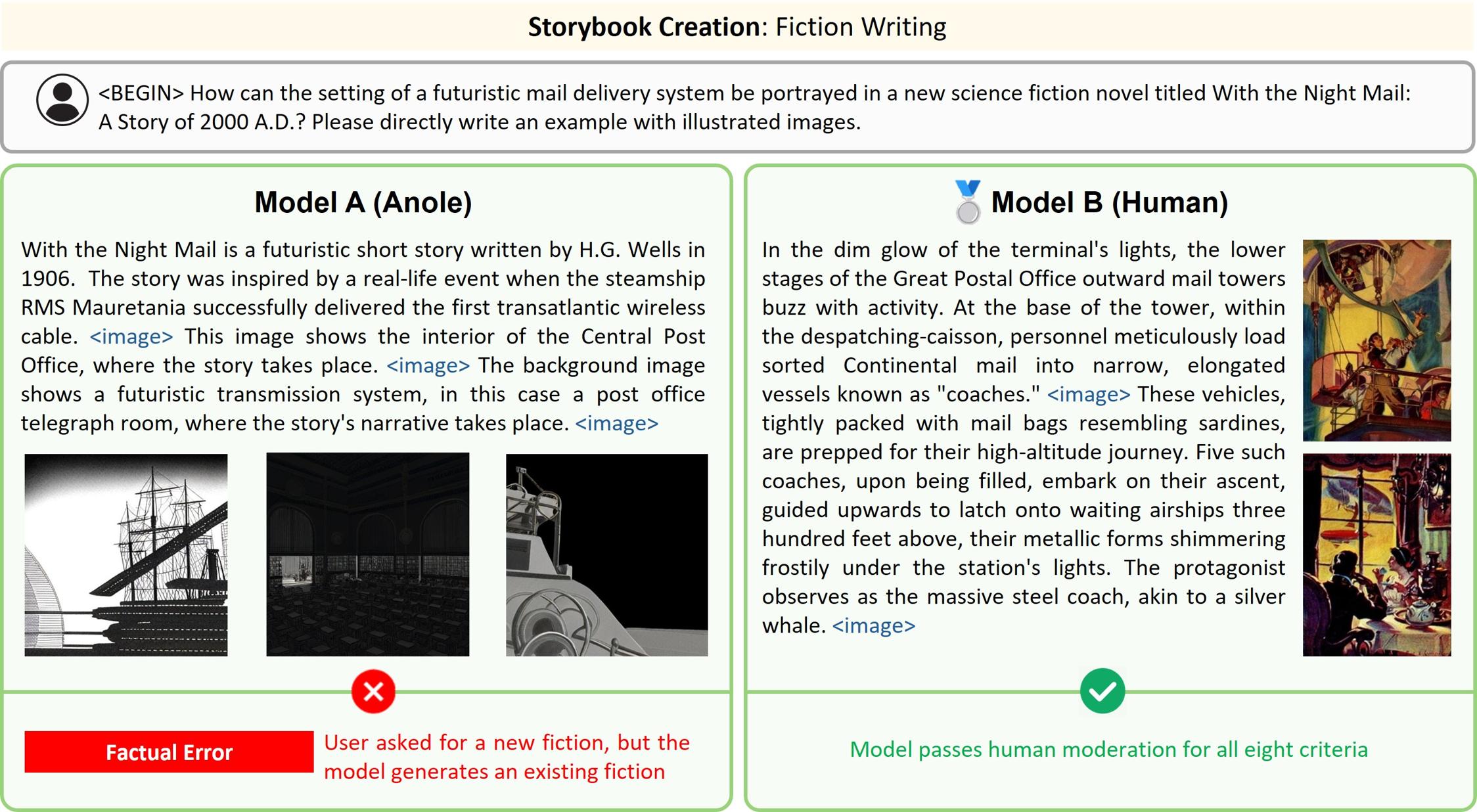} 
    \hspace{3mm}
    \end{subfigure}
\end{figure*}

\begin{figure*}[h]\ContinuedFloat
    \begin{subfigure}{\textwidth}
    \centering
    \includegraphics[width=0.98\linewidth]{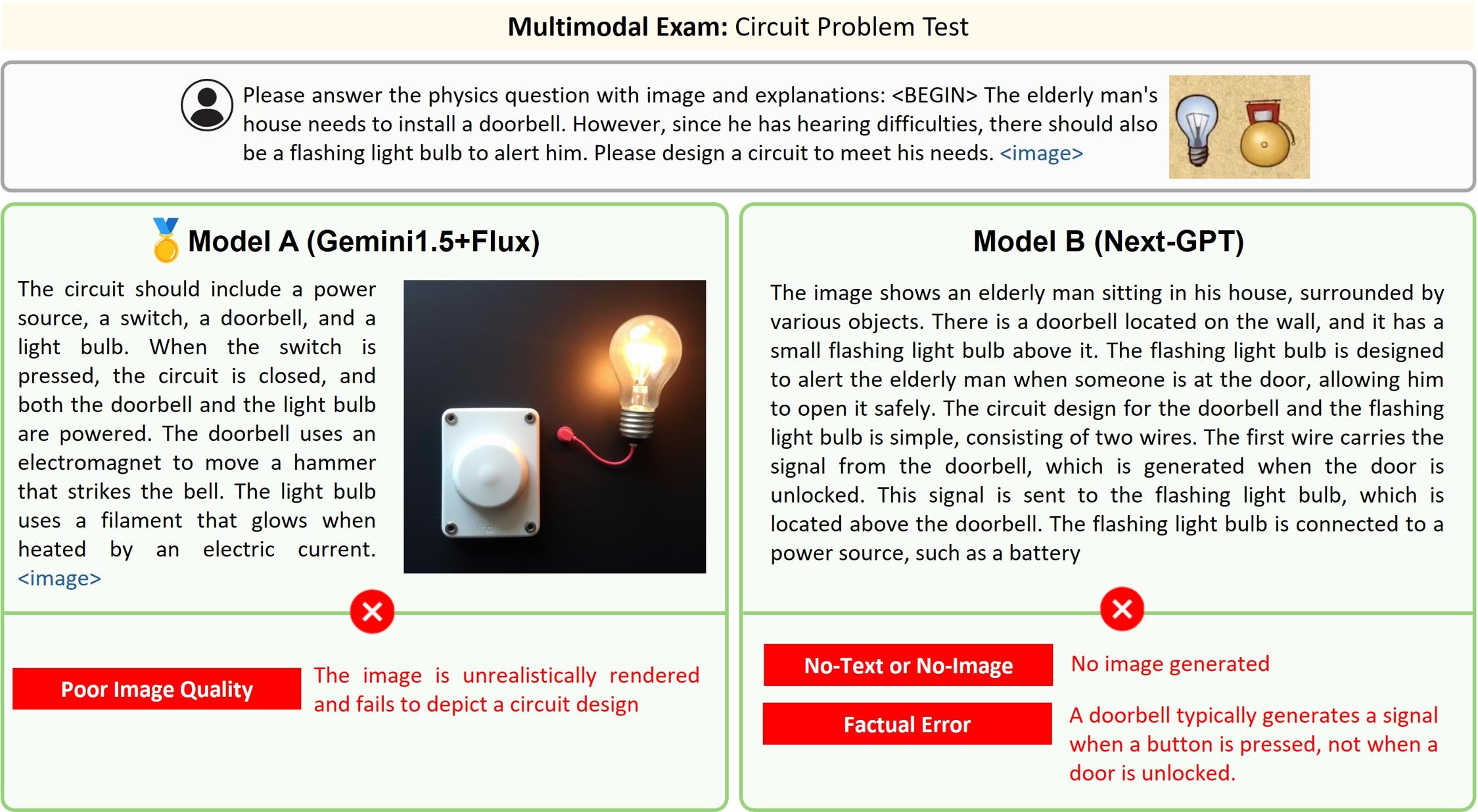} 
    \hspace{3mm}
    \end{subfigure}

    \begin{subfigure}{\textwidth}
    \centering
    \includegraphics[width=0.98\linewidth]{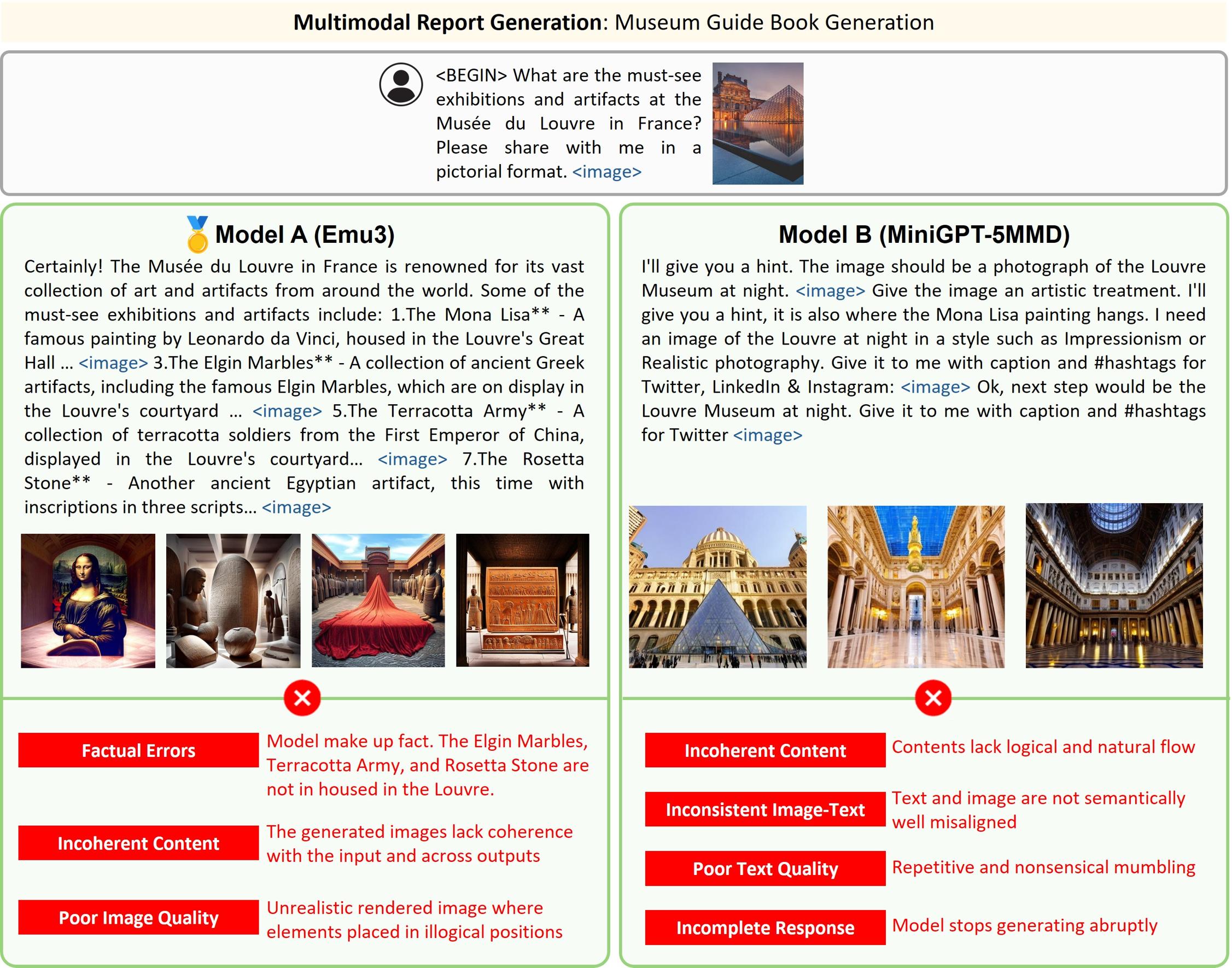} 
    \hspace{3mm}
    \end{subfigure}  
    
\end{figure*}

\begin{figure*}[h]\ContinuedFloat
    
    \begin{subfigure}{\textwidth}
    \centering
    \includegraphics[width=0.98\linewidth]{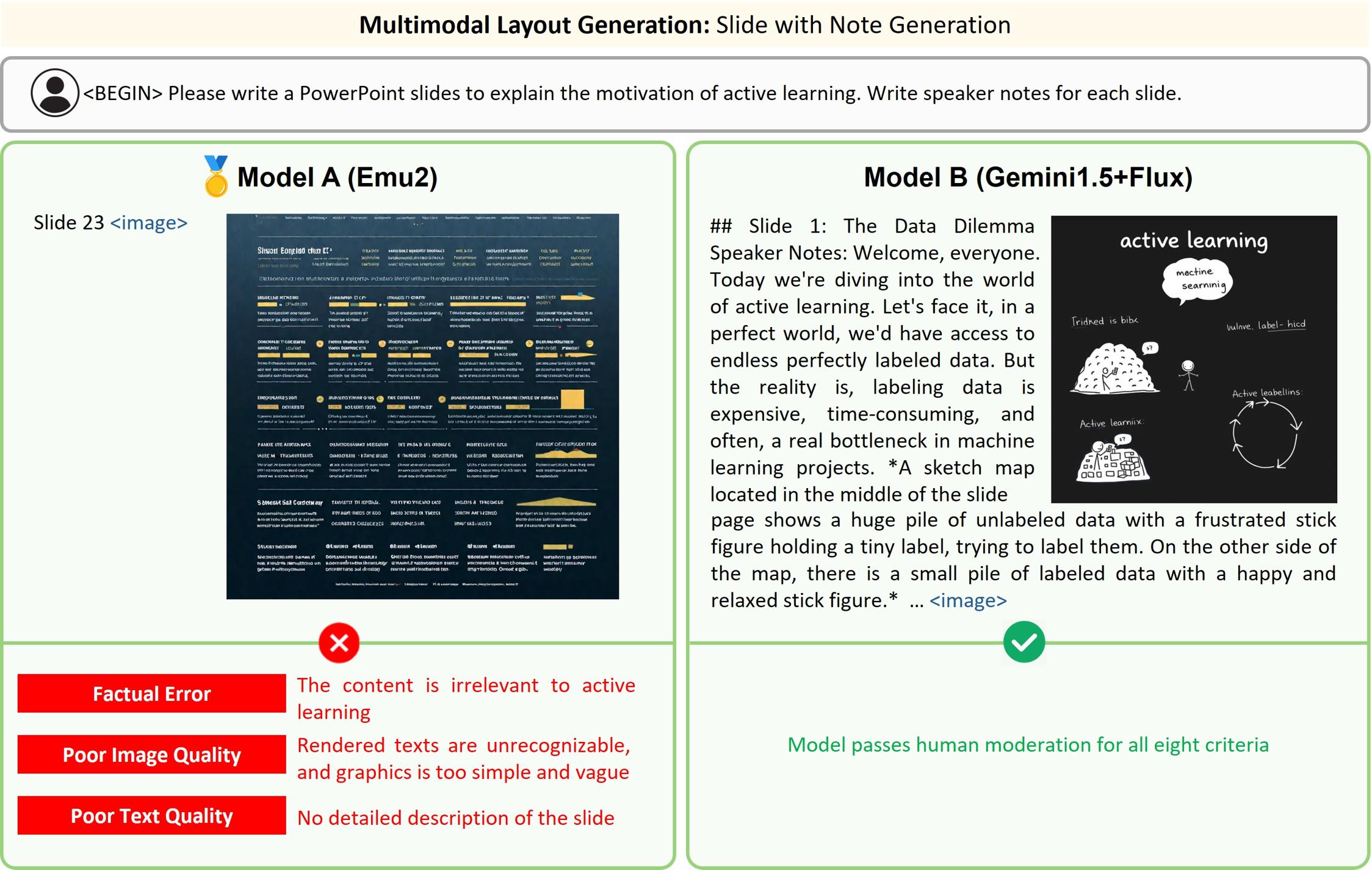} 
    \hspace{3mm}
    \end{subfigure}  

    \begin{subfigure}{\textwidth}
    \centering
    \includegraphics[width=0.98\linewidth]{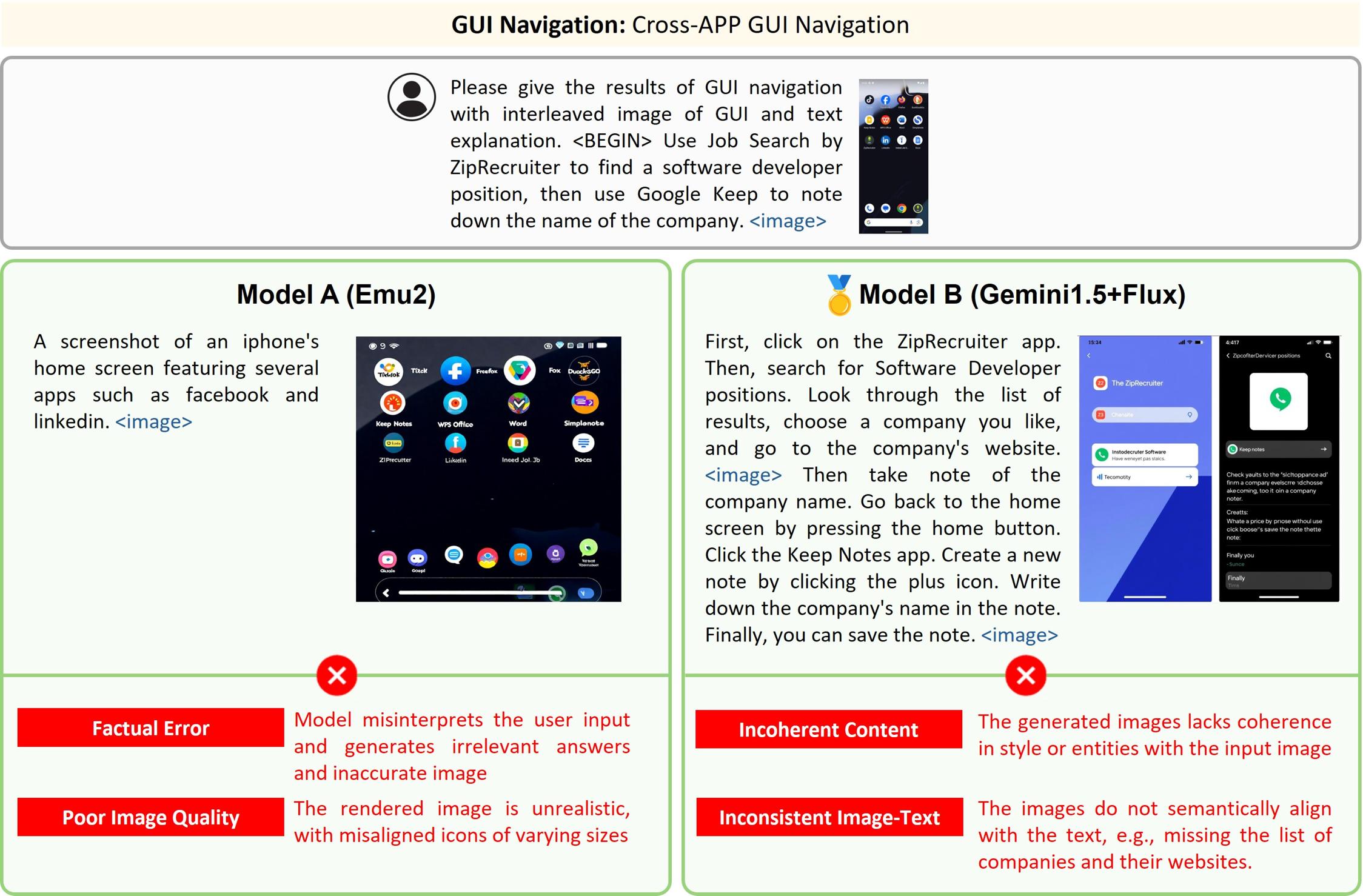} 
    \hspace{3mm}
    \end{subfigure}
    
\end{figure*}

\begin{figure*}[h]\ContinuedFloat
    
    \begin{subfigure}{\textwidth}
    \centering
    \includegraphics[width=0.98\linewidth]{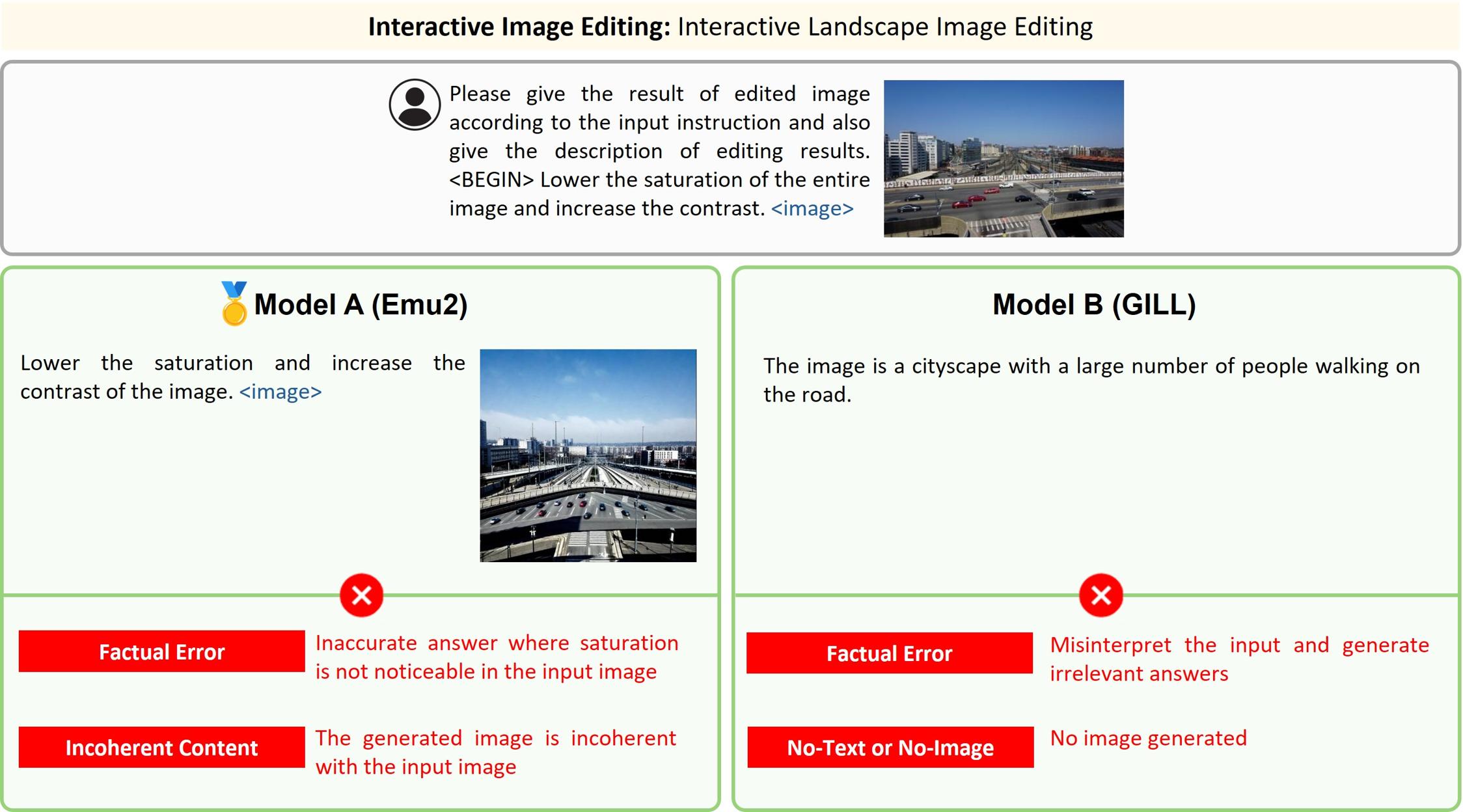} 
    \hspace{3mm}
    \end{subfigure}  

     \begin{subfigure}{\textwidth}
    \centering
    \includegraphics[width=0.98\linewidth]{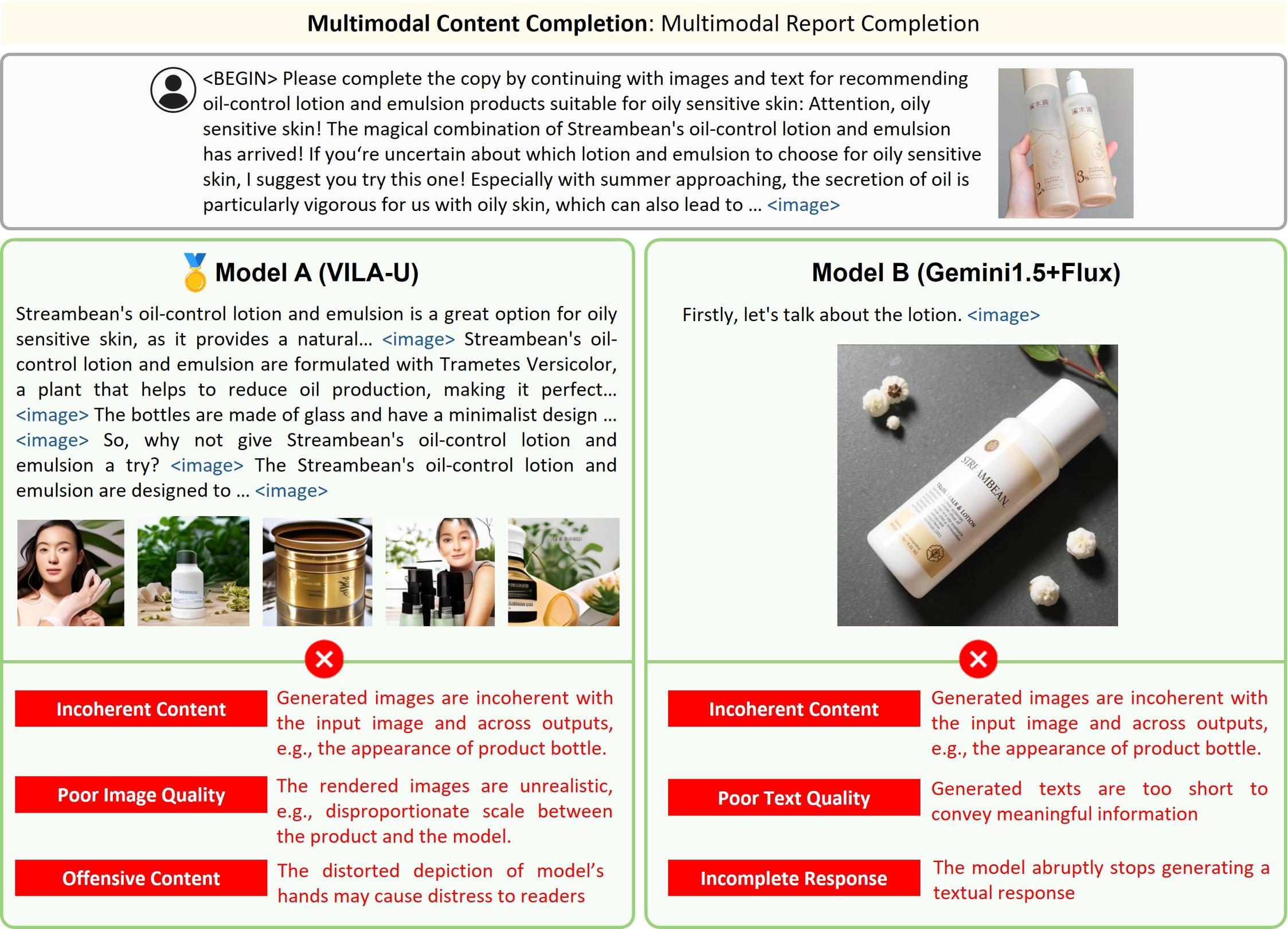} 
    \hspace{3mm}
    \end{subfigure}  
    
\end{figure*}

\begin{figure*}[h]\ContinuedFloat
   
    \begin{subfigure}{\textwidth}
    \centering
    \includegraphics[width=0.98\linewidth]{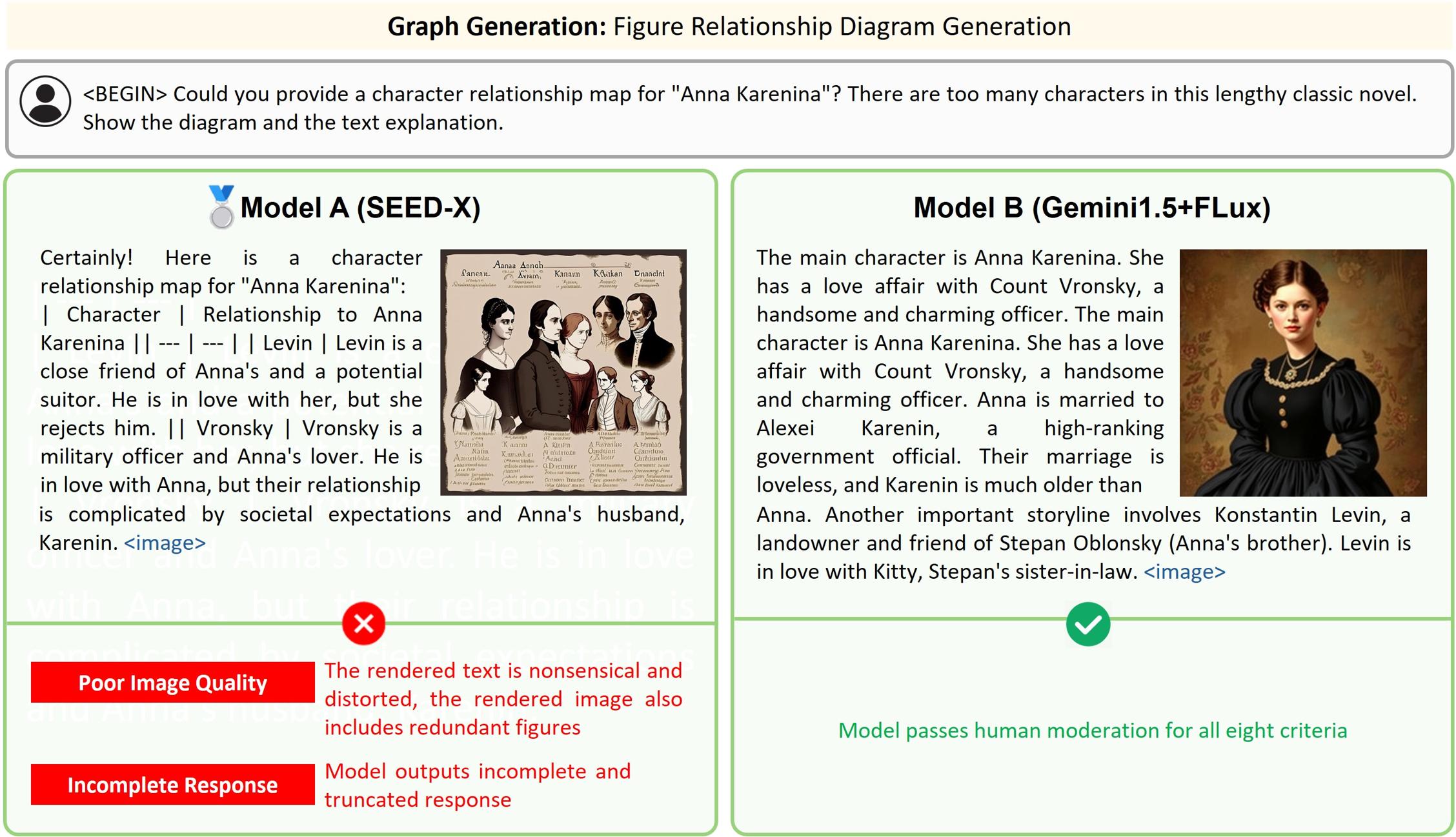} 
    \hspace{3mm}
    \end{subfigure}  

    \begin{subfigure}{\textwidth}
    \centering
    \includegraphics[width=0.98\linewidth]{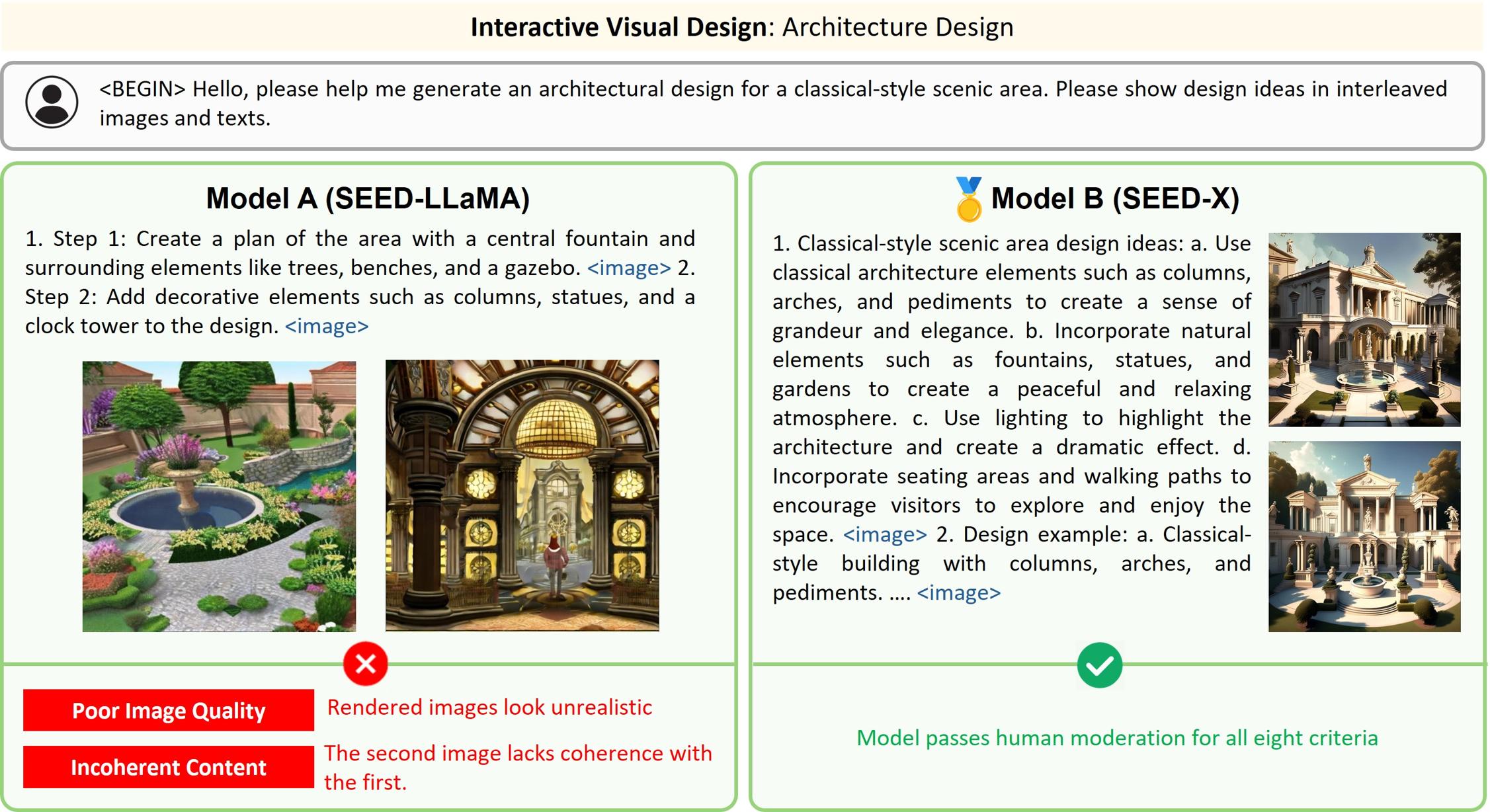} 
    \hspace{3mm}
    \end{subfigure}  
    
\end{figure*}

\begin{figure*}[h]\ContinuedFloat
    
    \begin{subfigure}{\textwidth}
    \centering
    \includegraphics[width=0.98\linewidth]{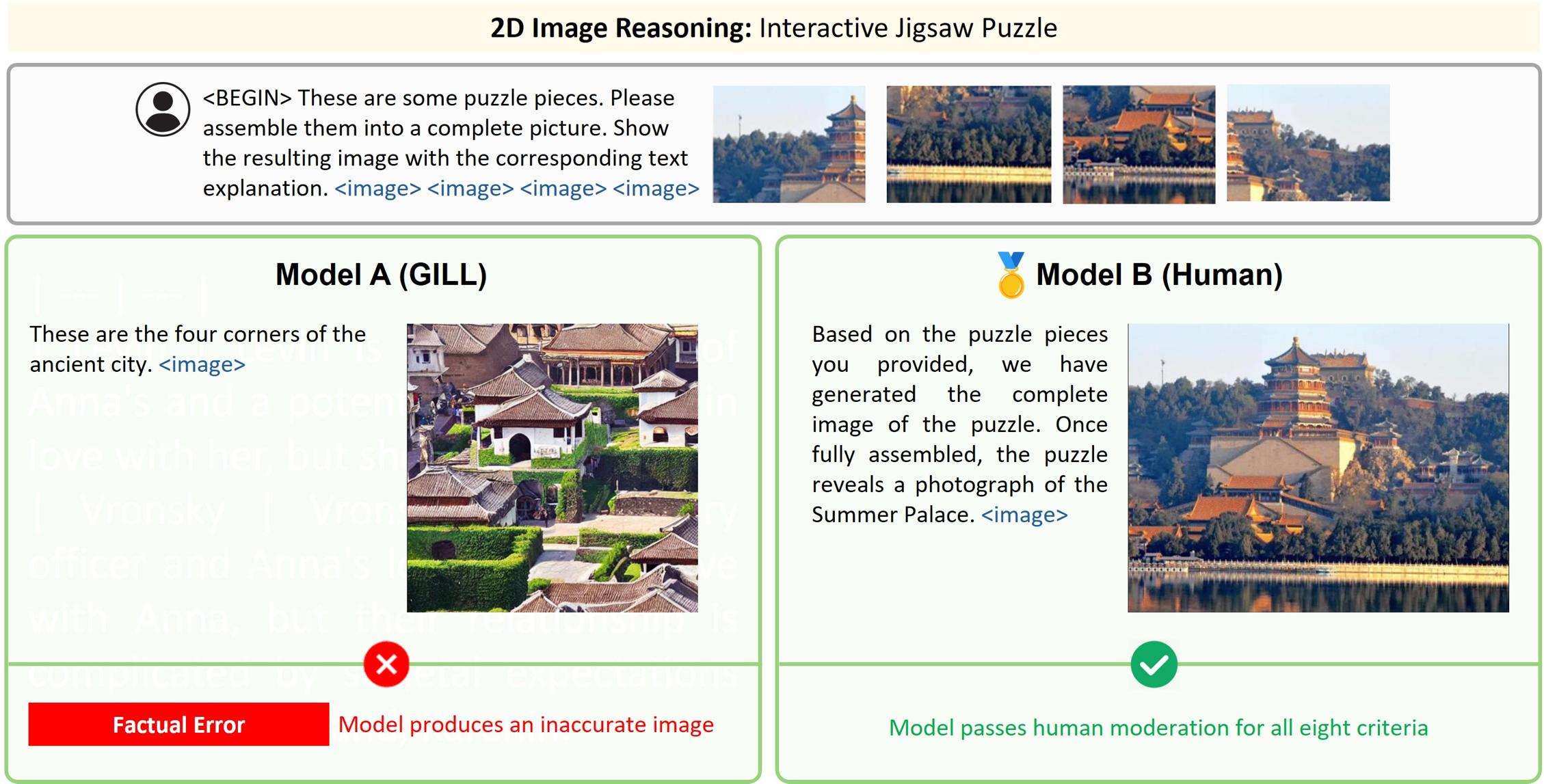} 
    \hspace{3mm}
    \end{subfigure}  

    \begin{subfigure}{\textwidth}
    \centering
    \includegraphics[width=0.98\linewidth]{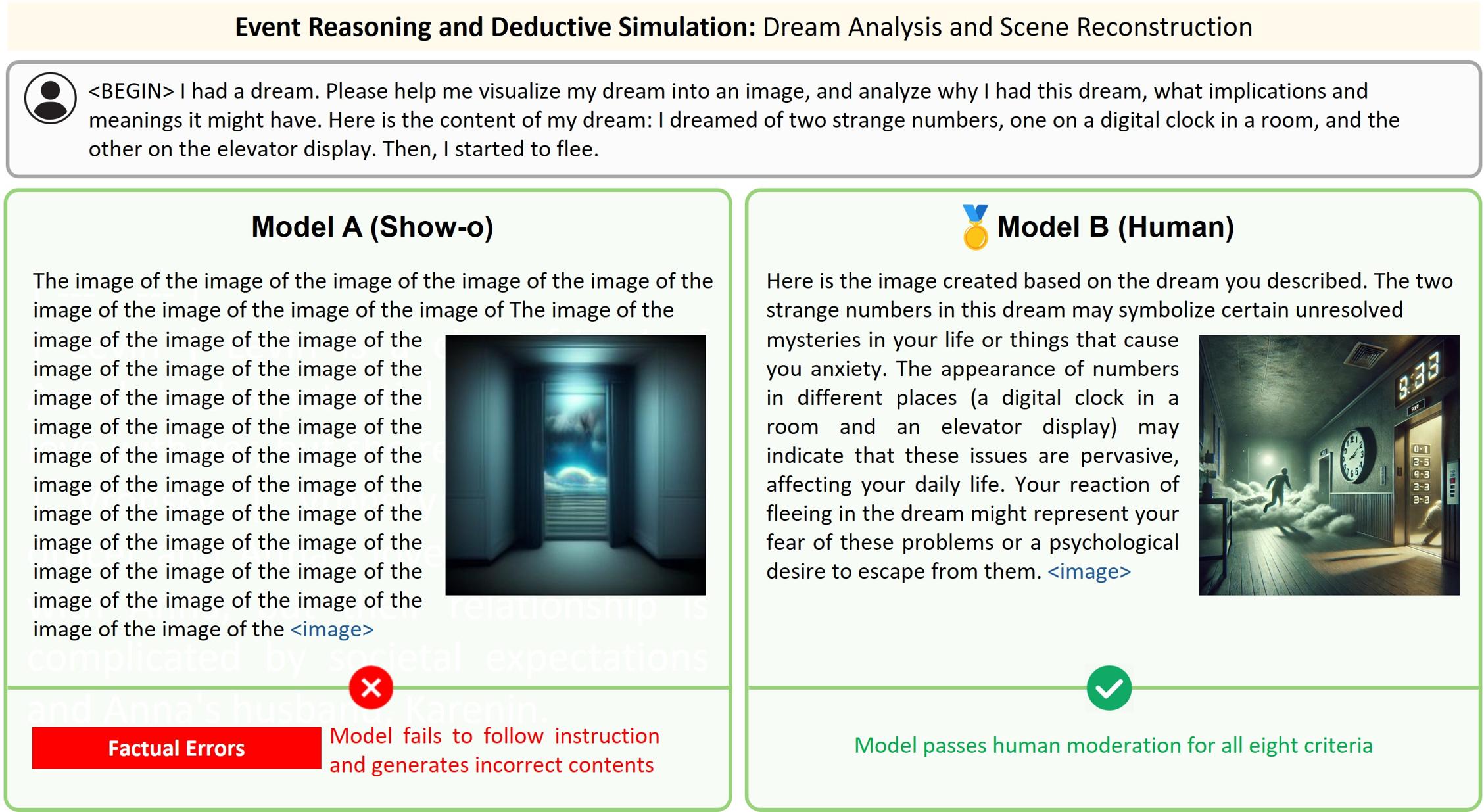} 
    \hspace{3mm}
    \end{subfigure} 
 
\end{figure*}

\begin{figure*}[h]\ContinuedFloat
    
    \begin{subfigure}{\textwidth}
    \centering
    \includegraphics[width=0.98\linewidth]{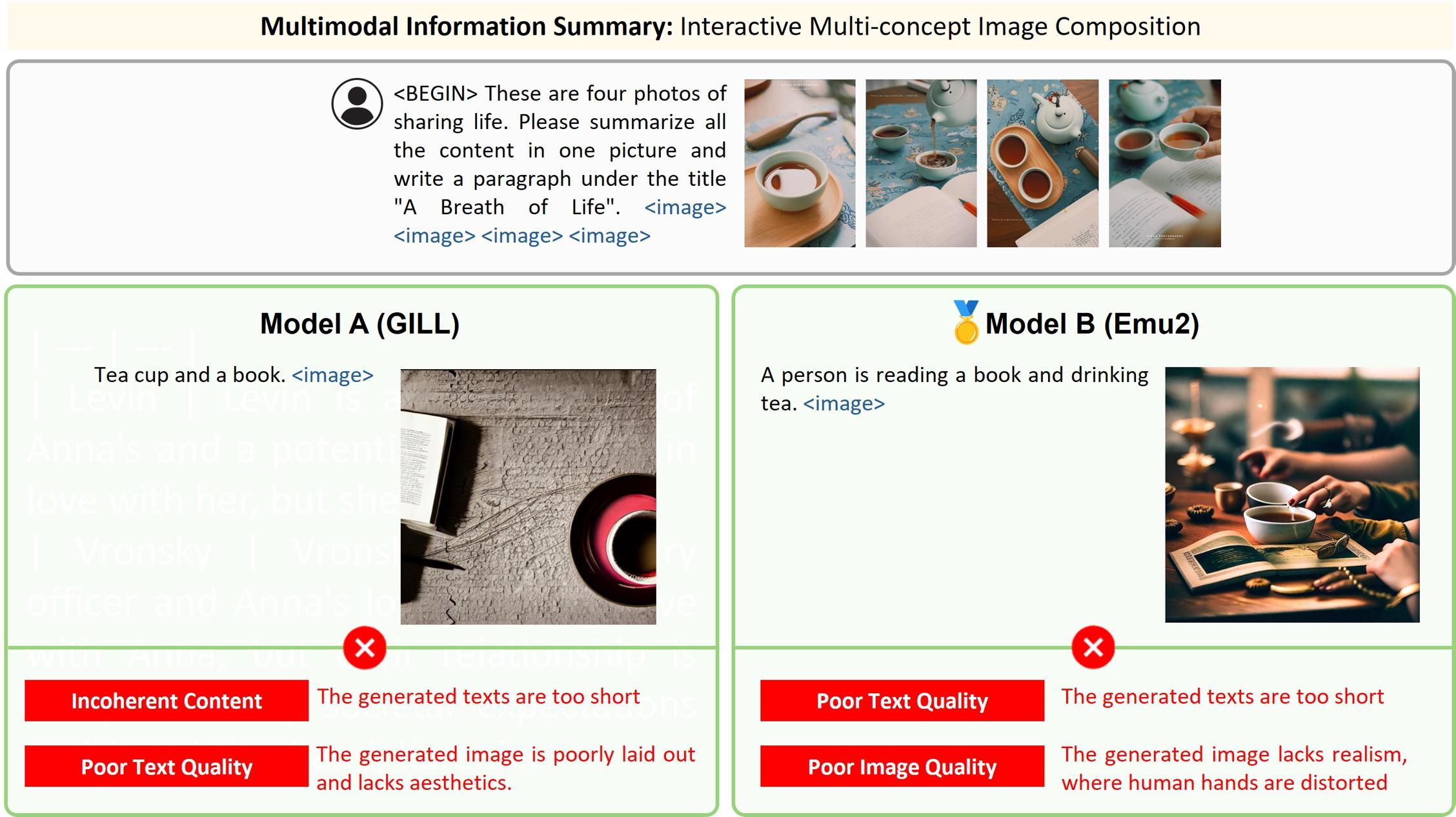} 
    \hspace{3mm}
    \end{subfigure}  

    \begin{subfigure}{\textwidth}
    \centering
    \includegraphics[width=0.98\linewidth]{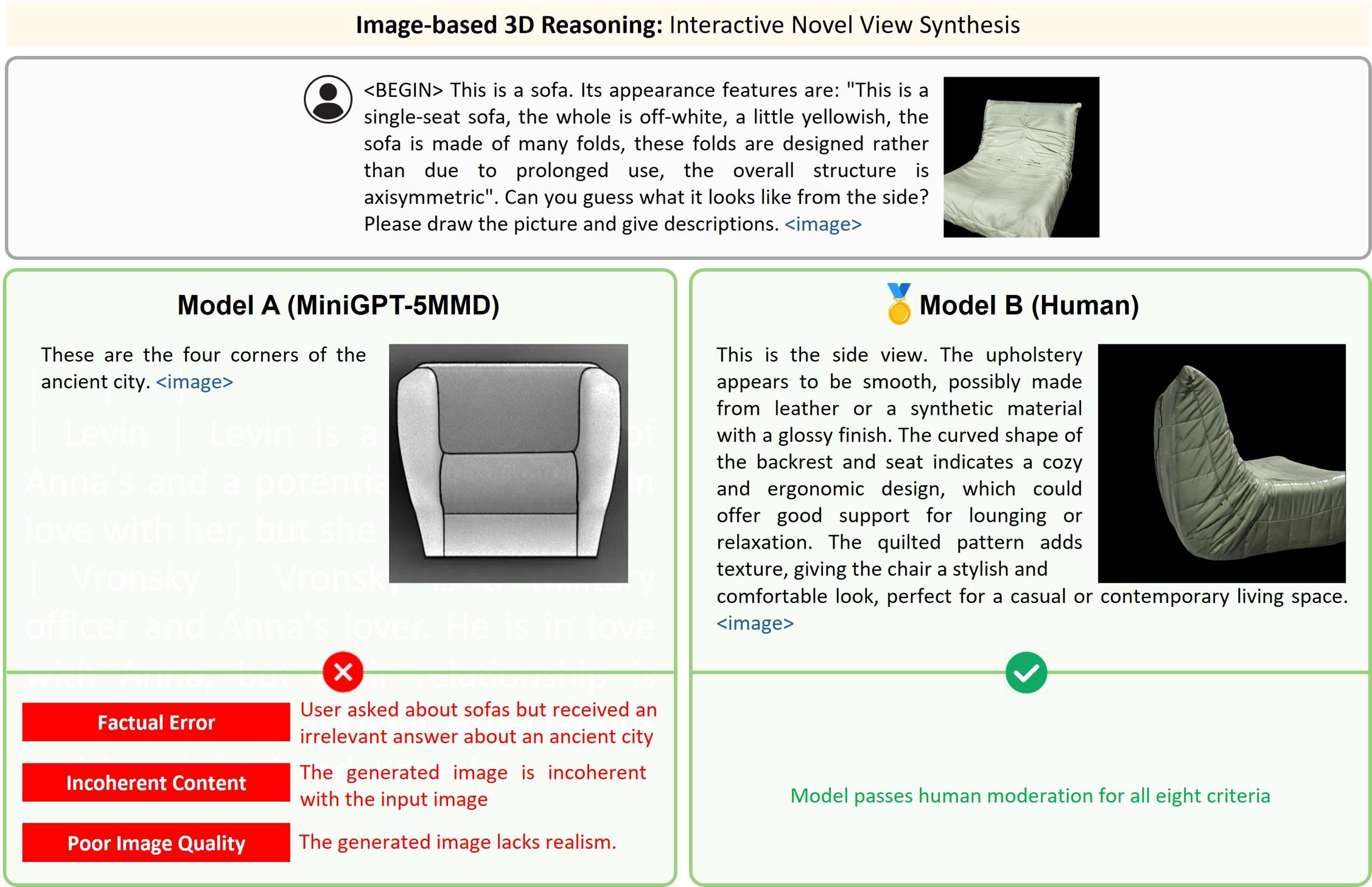} 
    \hspace{3mm}
    \end{subfigure}  
    
\end{figure*}

\begin{figure*}[h]\ContinuedFloat
    
    \begin{subfigure}{\textwidth}
    \centering
    \includegraphics[width=0.98\linewidth]{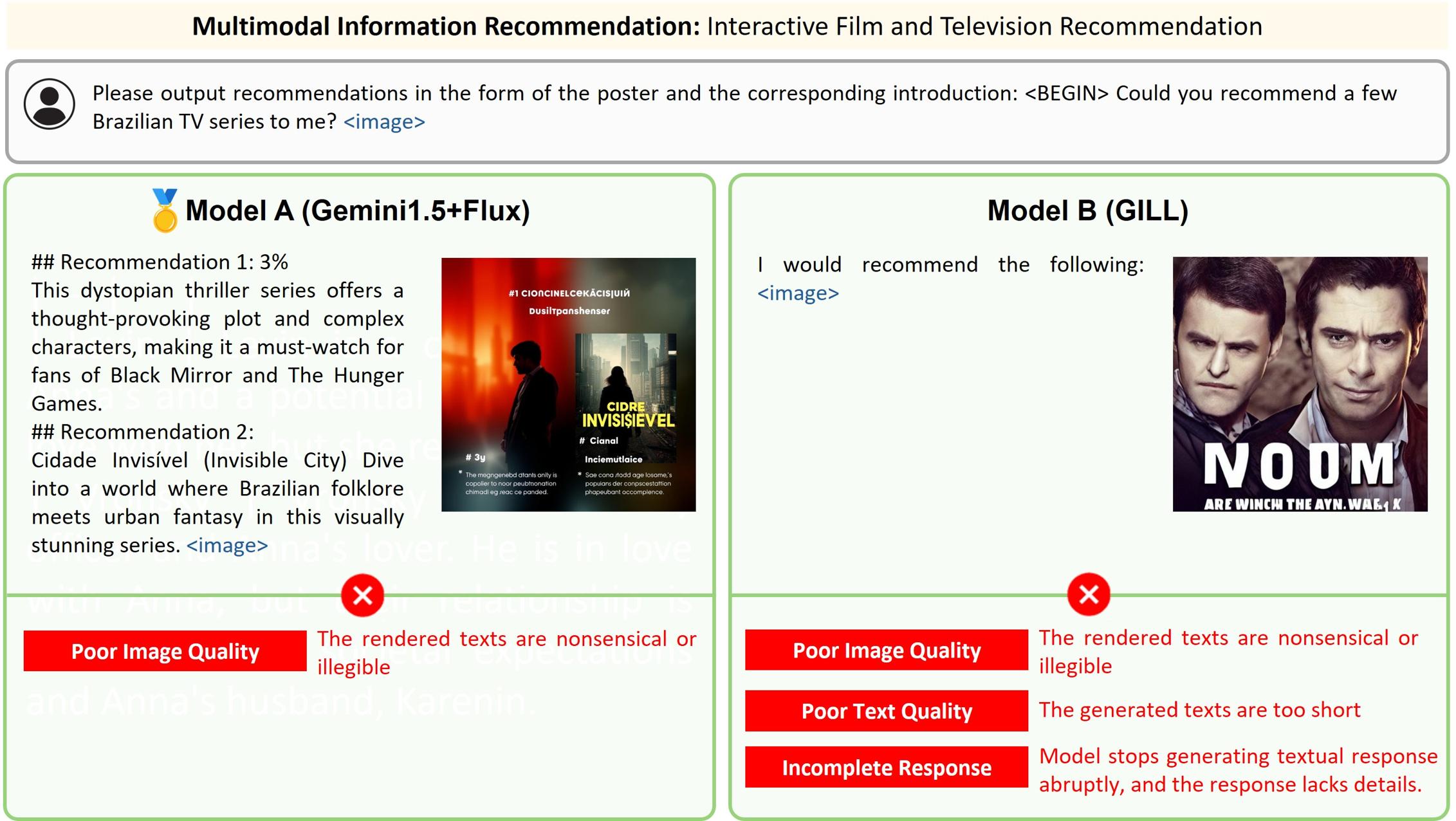} 
    \hspace{3mm}
    \end{subfigure}  

    \begin{subfigure}{\textwidth}
    \centering
    \includegraphics[width=0.98\linewidth]{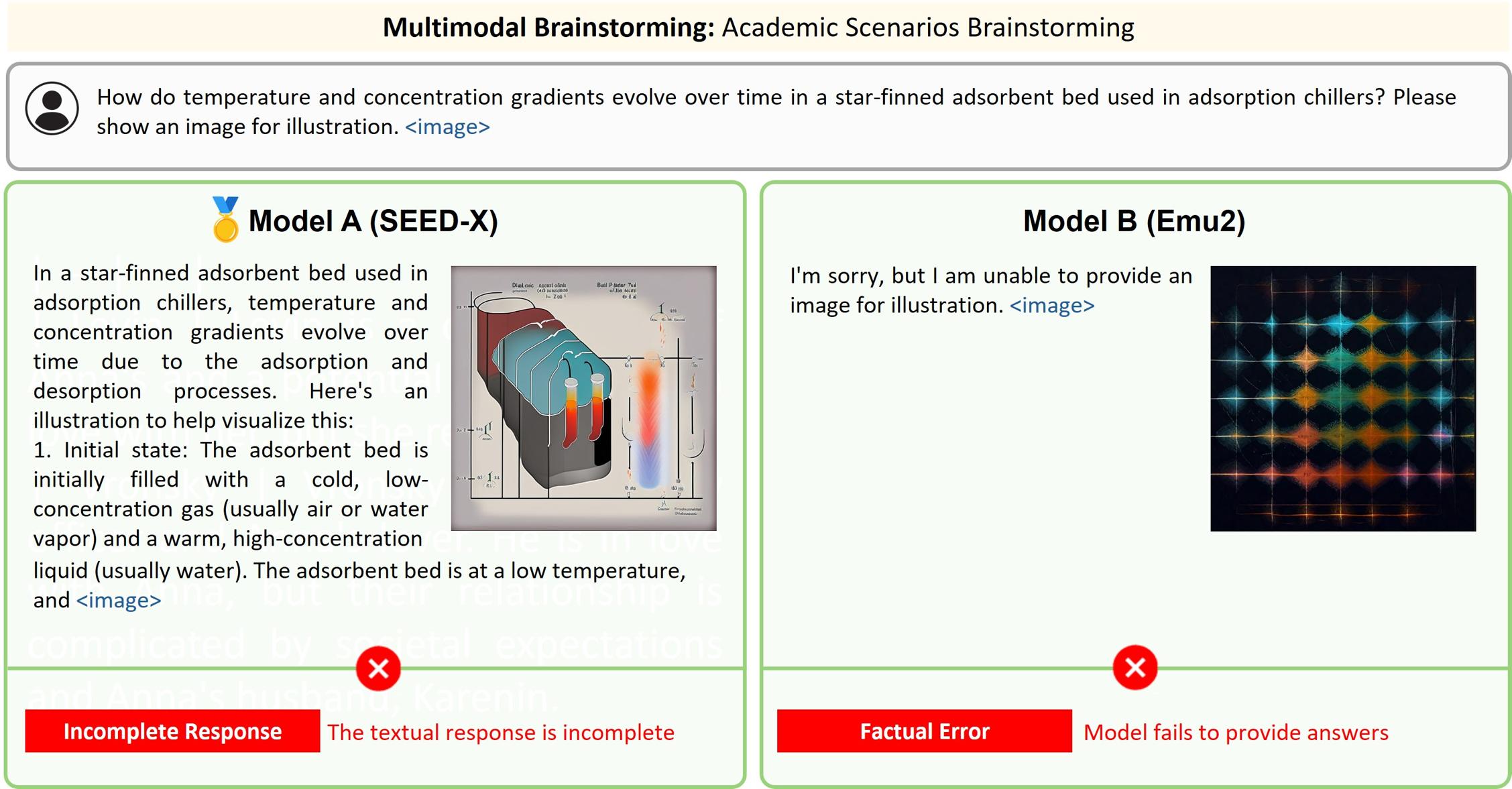} 
    \hspace{3mm}
    \end{subfigure}  
    
\end{figure*}

\begin{figure*}[h]\ContinuedFloat
    
    \begin{subfigure}{\textwidth}
    \centering
    \includegraphics[width=0.98\linewidth]{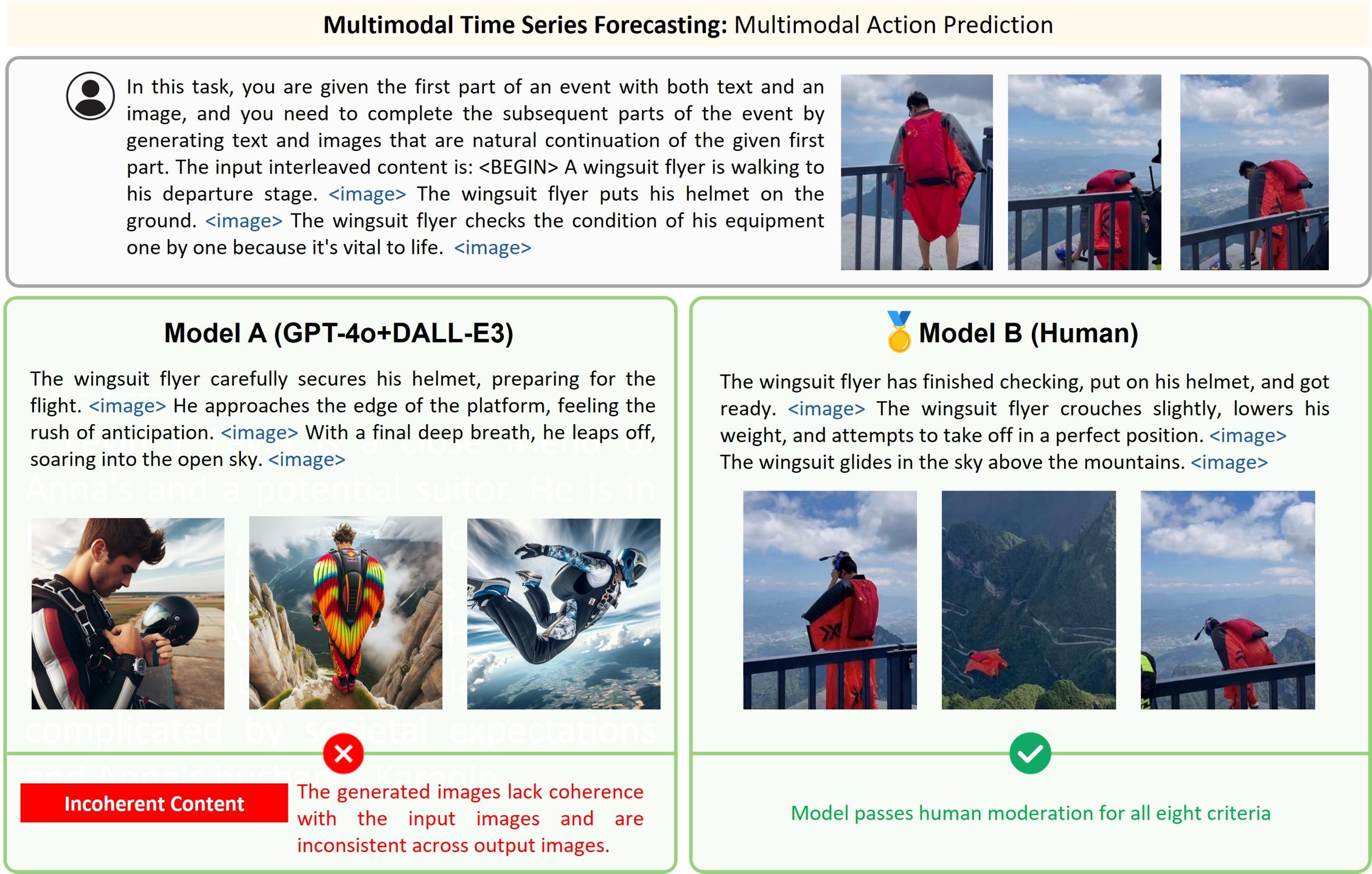} 
    \hspace{3mm}
    \end{subfigure} 

    \begin{subfigure}{\textwidth}
    \centering
    \includegraphics[width=0.98\linewidth]{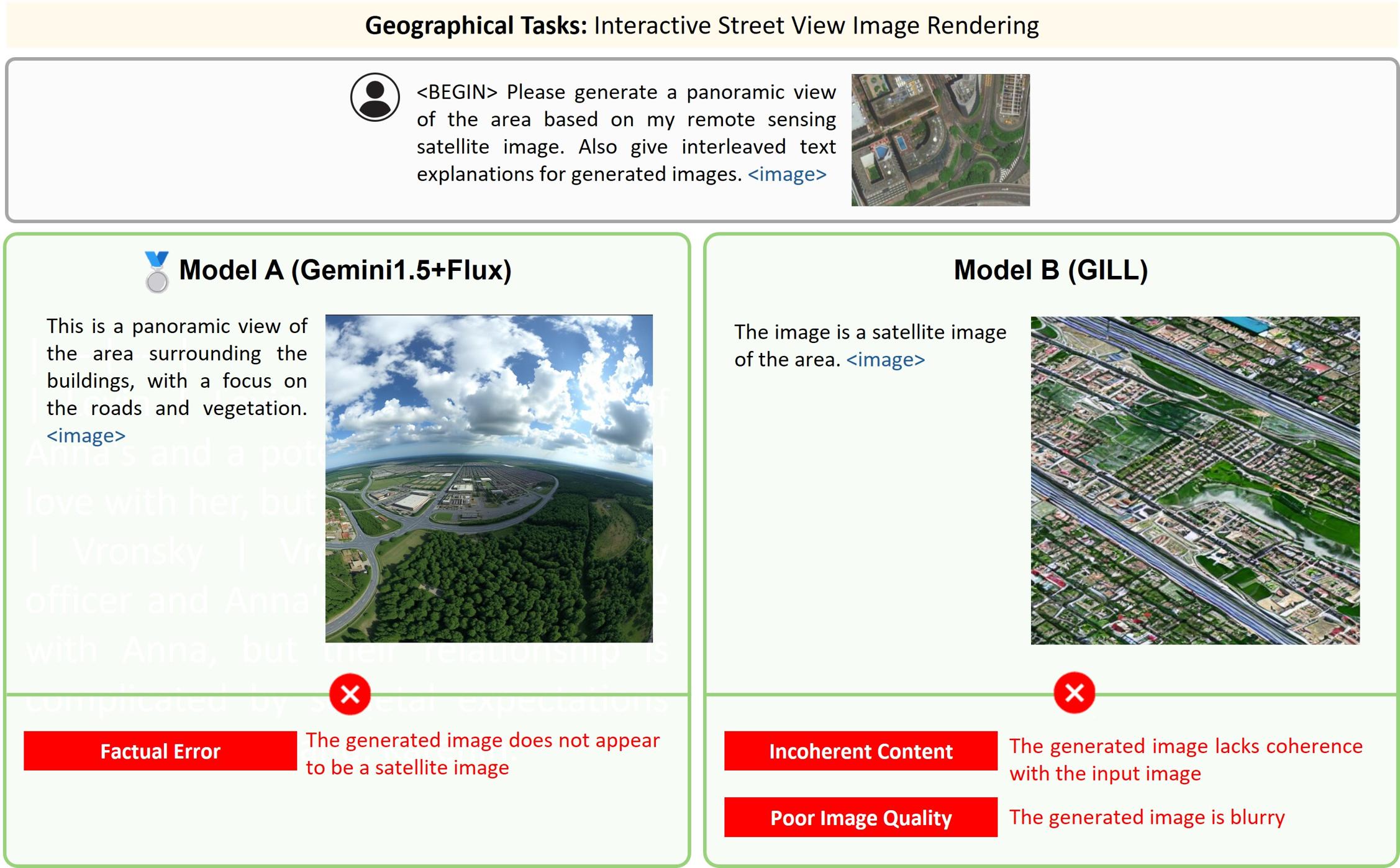} 
    \hspace{3mm}
    \end{subfigure}  
    
\end{figure*}

\begin{figure*}[h]\ContinuedFloat
    
    \begin{subfigure}{\textwidth}
    \centering
    \includegraphics[width=0.98\linewidth]{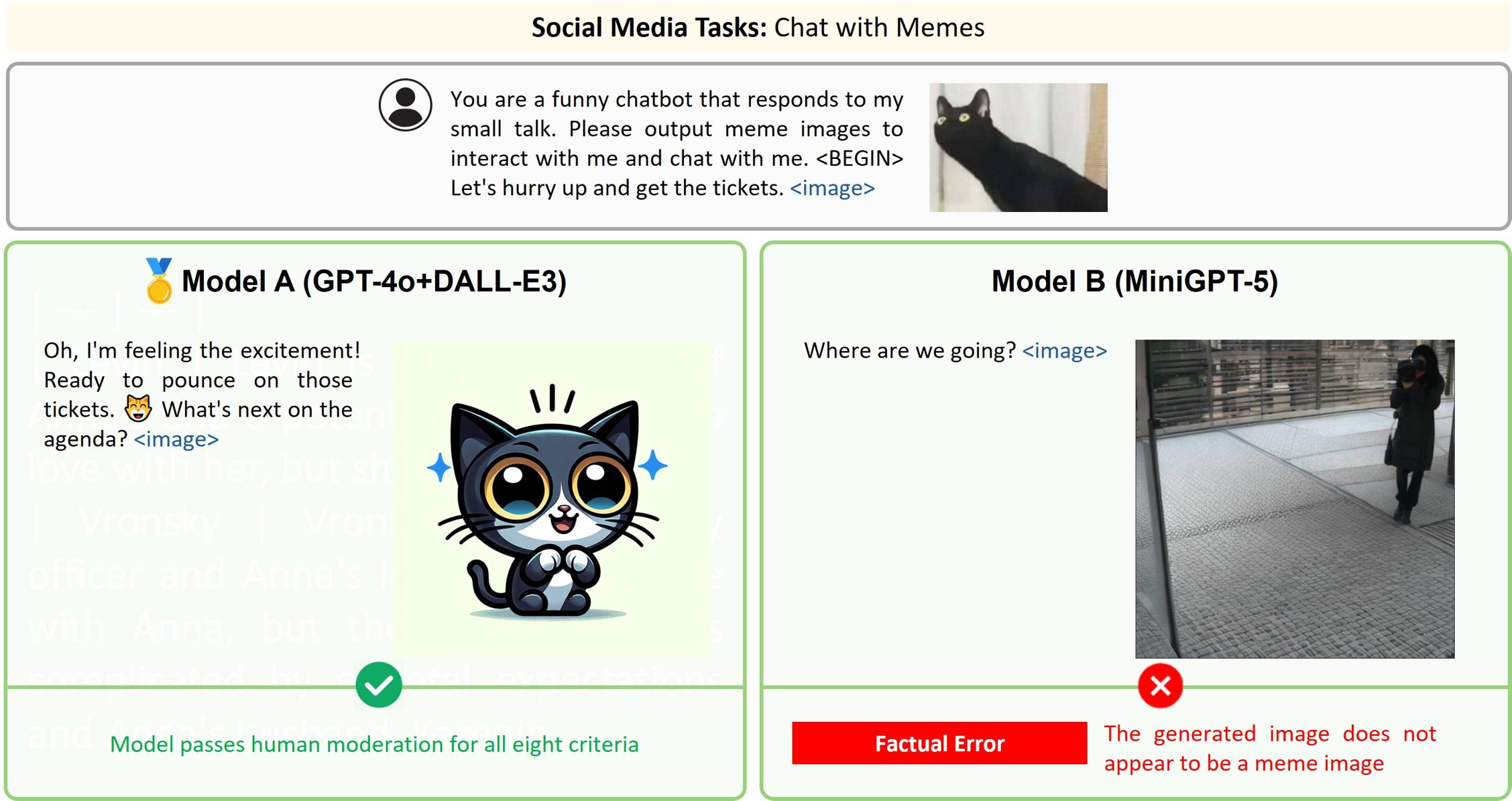} 
    \hspace{3mm}
    \end{subfigure}  

    \begin{subfigure}{\textwidth}
    \centering
    \includegraphics[width=0.98\linewidth]{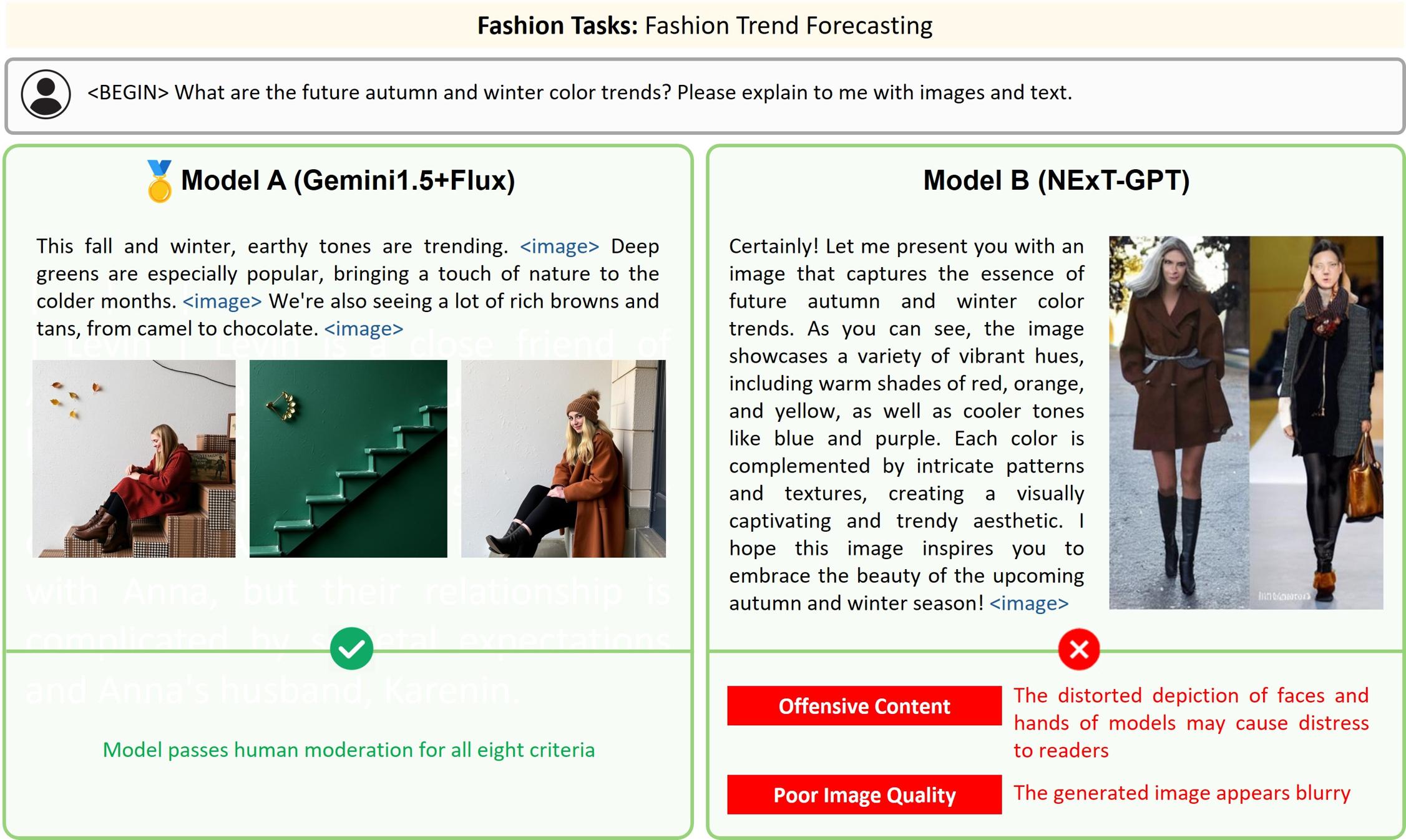} 
    \hspace{3mm}
    \end{subfigure}  
    
\end{figure*}

\begin{figure*}[h]\ContinuedFloat
    
    \begin{subfigure}{\textwidth}
    \centering
    \includegraphics[width=0.98\linewidth]{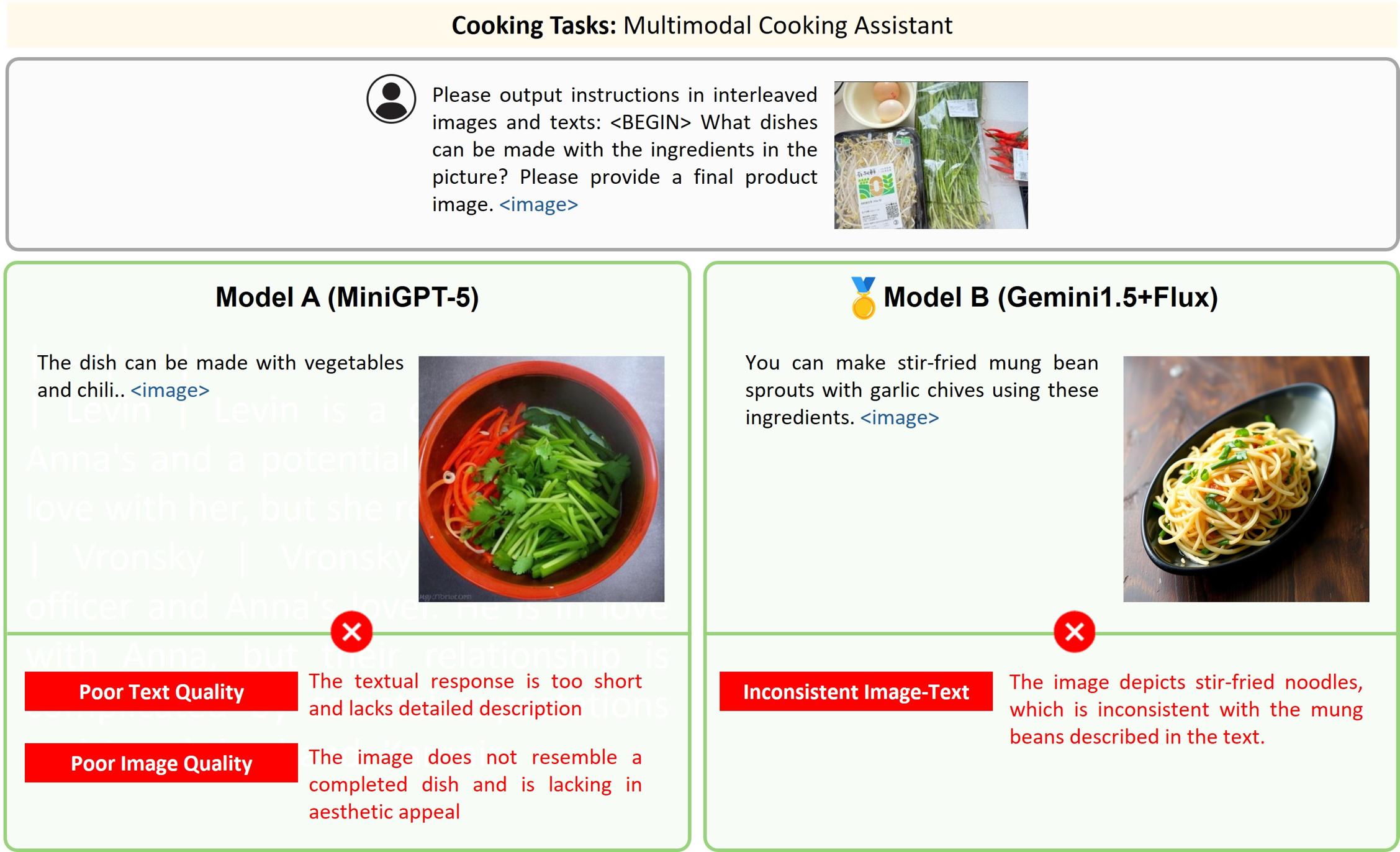} 
    \hspace{3mm}
    \end{subfigure}  

    \begin{subfigure}{\textwidth}
    \centering
    \includegraphics[width=0.98\linewidth]{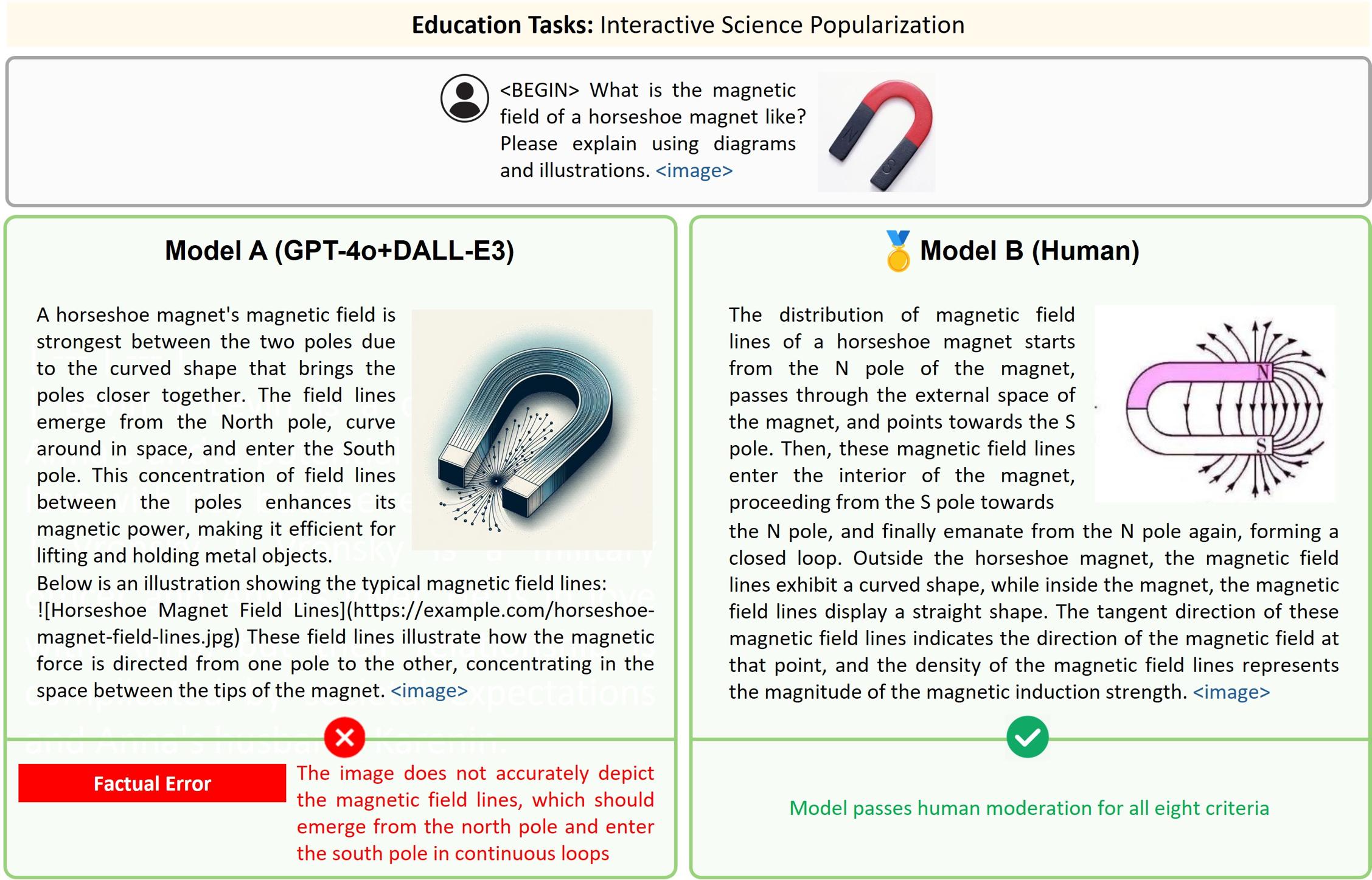} 
    \hspace{3mm}
    \end{subfigure}  
    
\end{figure*}

\begin{figure*}[h]\ContinuedFloat
    
    \begin{subfigure}{\textwidth}
    \centering
    \includegraphics[width=0.98\linewidth]{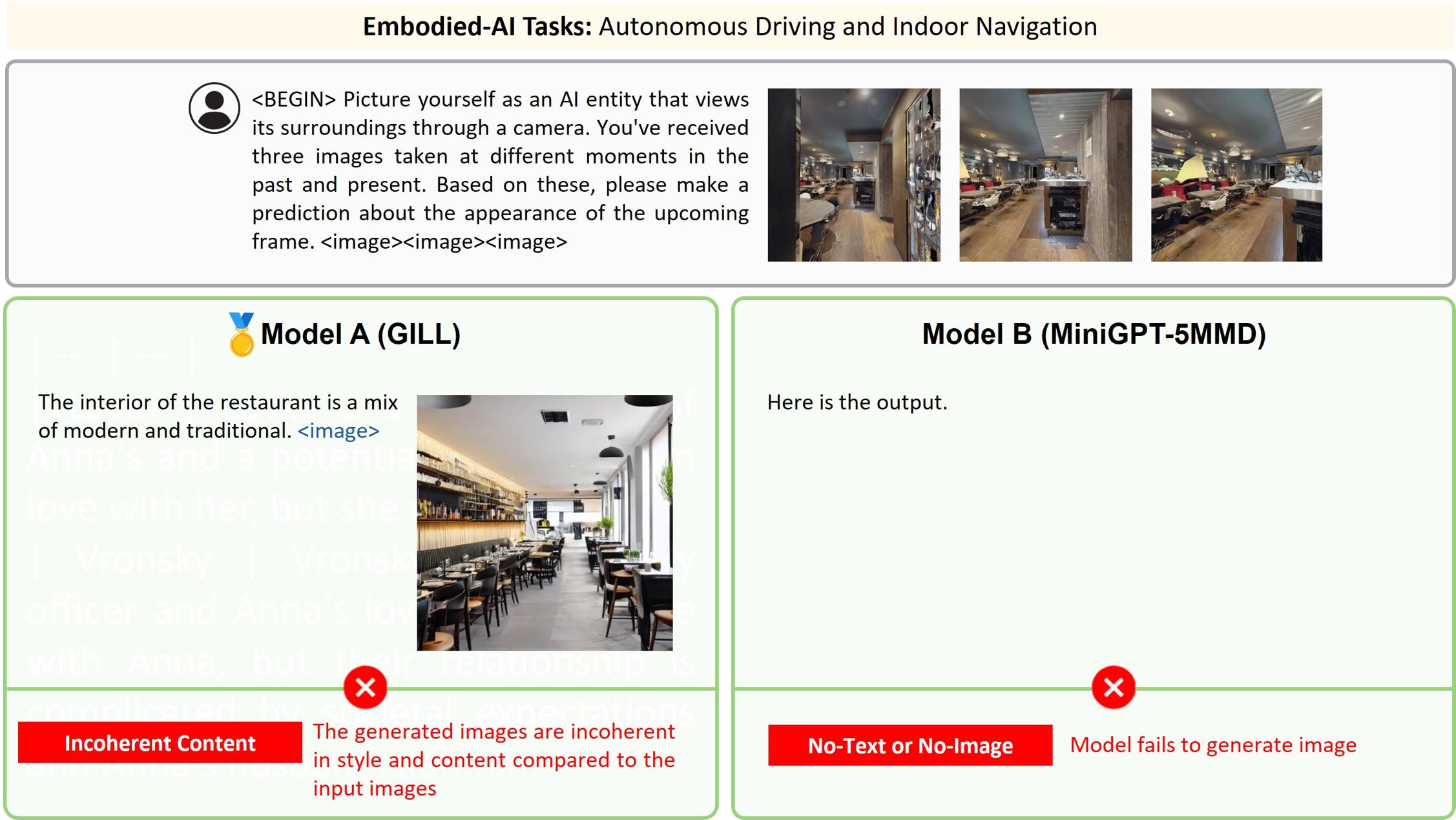} 
    \hspace{3mm}
    \end{subfigure}  
    
    \begin{subfigure}{\textwidth}
    \centering
    \includegraphics[width=0.98\linewidth]{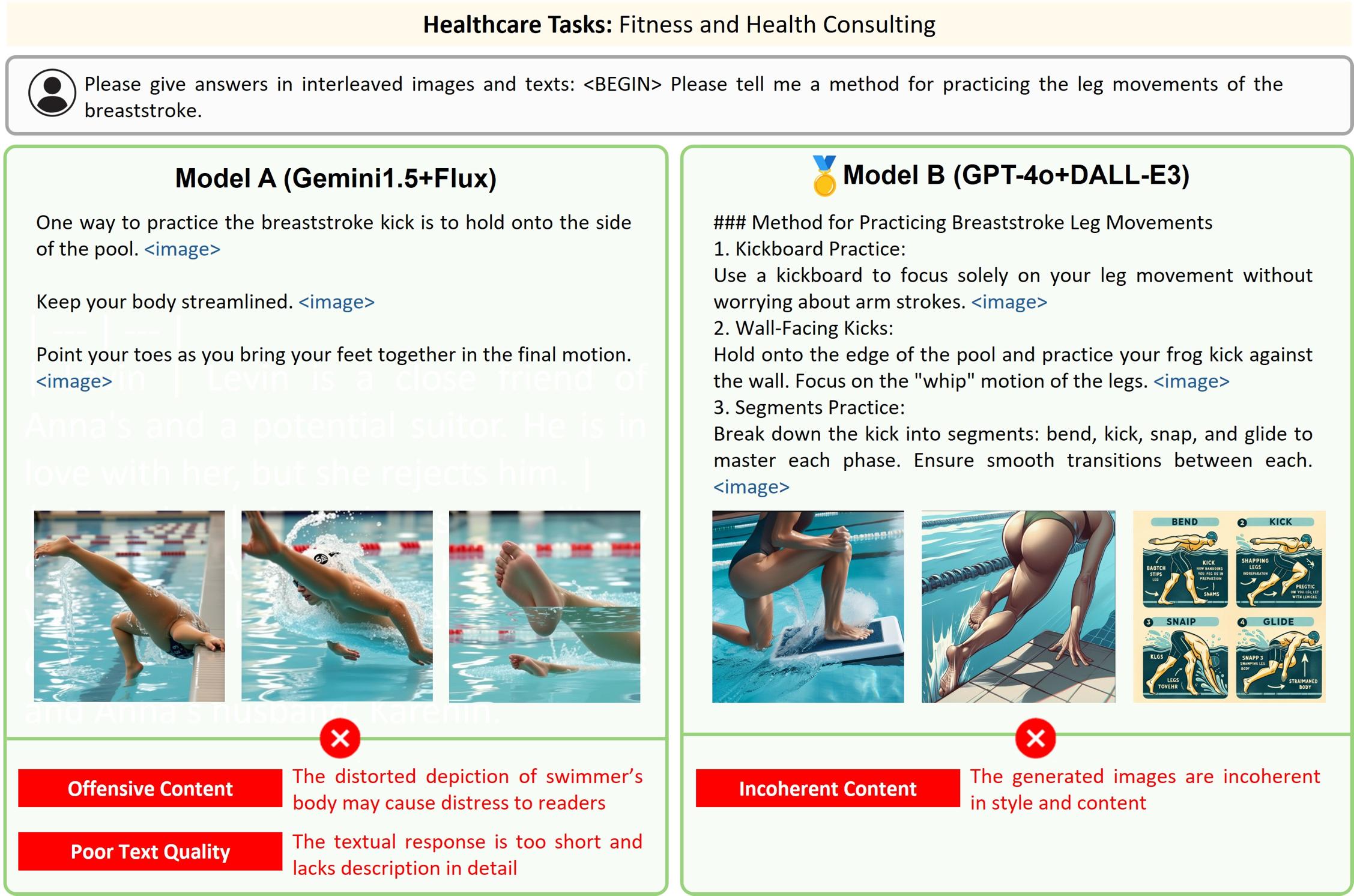} 
    \hspace{3mm}
    \end{subfigure}
    
\end{figure*}



\end{document}